\documentclass{article} % For LaTeX2e
\usepackage[dvipsnames]{xcolor}
\usepackage{iclr2024_conference,times}

\usepackage{amsmath,amsfonts,bm}

\def\eqref#1{equation~\ref{#1}}
\def\1{\bm{1}}

\def\vs{{\bm{s}}}

\DeclareMathAlphabet{\mathsfit}{\encodingdefault}{\sfdefault}{m}{sl}
\SetMathAlphabet{\mathsfit}{bold}{\encodingdefault}{\sfdefault}{bx}{n}

\def\gD{{\mathcal{D}}}

\def\gL{{\mathcal{L}}}

\def\gT{{\mathcal{T}}}
\def\gU{{\mathcal{U}}}

\usepackage{url}
\usepackage{graphicx} % Required for inserting images

\usepackage[pagebackref,breaklinks,colorlinks,citecolor=ForestGreen,linkcolor=BrickRed]{hyperref}

\usepackage{caption}
\usepackage{subcaption}
\usepackage{booktabs}
\usepackage{multirow}
\usepackage{gensymb}
\usepackage{etoc}
\usepackage{pifont}
\usepackage{xspace}
\usepackage{adjustbox}
\usepackage{makecell}
\usepackage[normalem]{ulem}
\usepackage[capitalize]{cleveref}
\usepackage{wrapfig}
\usepackage{caption}

\definecolor{darkgray}{gray}{0.45}
\newcounter{rownumbers}
\newcommand\rownum{{\color{darkgray} \stepcounter{rownumbers}\arabic{rownumbers}.}}

\crefname{section}{Sec.}{Secs.}
\Crefname{section}{Sec.}{Secs.}
\Crefname{table}{Tab.}{Tabs.}
\crefname{table}{Tab.}{Tabs.}
\Crefname{figure}{Fig.}{Figs.}
\crefname{figure}{Fig.}{Figs.}
\Crefname{equation}{Eq.}{Eqs.}
\crefname{equation}{Eq.}{Eqs.}

\newcommand{\ltnorm}[1]{\left\|#1\right\|_2}

\makeatletter
\DeclareRobustCommand\onedot{\futurelet\@let@token\@onedot}
\def\@onedot{\ifx\@let@token.\else.\null\fi\xspace}

\def\eg{\emph{e.g}\onedot} 
\def\ie{\emph{i.e}\onedot} 
 
\def\etc{\emph{etc}\onedot} \def\vs{\emph{vs}\onedot}

\makeatother

\newcommand{\myparagraph}[1]{\noindent\textbf{#1.}}
\newcommand{\nodotparagraph}[1]{\noindent\textbf{#1}}

\newcommand{\retforloc}{Ret4Loc-HOW\xspace}
\newcommand{\howam}{Ret4Loc-HOW\xspace}

\newcommand{\Synth}{\howam-\textbf{Synth}\xspace}
\newcommand{\Synthp}{\howam-\textbf{Synth+}\xspace}
\newcommand{\Synthpp}{\howam-\textbf{Synth++}\xspace}

\newcommand{\cmark}{\ding{52}}

\definecolor{teal}{RGB}{41,120,108}

\title{Weatherproofing \underline{Ret}rieval \underline{for} \underline{Loc}alization with Generative AI \& Geometric Consistency}

\author{Yannis Kalantidis$^*$ \And Mert B\"{u}lent Sar{\i}y{\i}ld{\i}z\thanks{Equal contribution} \And Rafael S. Rezende \AND Philippe Weinzaepfel \And Diane Larlus \And Gabriela Csurka  \AND
{\normalfont \centering \hfill NAVER LABS Europe \hfill } \\
}

\etocdepthtag.toc{mtchapter}
\etocsettagdepth{mtappendix}{none}

\iclrfinalcopy % Uncomment for camera-ready version, but NOT for submission.   
\begin{document}

\maketitle

\begin{abstract}
State-of-the-art visual localization approaches generally rely on a first image retrieval step whose role is crucial.
Yet, retrieval often struggles when facing varying conditions, due to \eg weather or time of day, with dramatic consequences on the visual localization accuracy. 
In this paper, we improve this retrieval step and tailor it to the final localization task. Among the several changes we advocate for, 
we propose to synthesize variants of the training set images, obtained from generative text-to-image models, 
in order to automatically expand the training set towards a number of nameable variations that particularly hurt visual localization. 
These changes result in \textbf{Ret4Loc}, a training approach that %uses 
learns from such synthetic variants together with real images and that leverages geometric consistency for filtering and sampling. Experiments %and experimentally 
show that it leads to large improvements on multiple challenging visual localization and place recognition benchmarks.
\newline
Project page: \url{https://europe.naverlabs.com/ret4loc}
\end{abstract}

\section{Introduction}\label{sec:introduction}

\begin{wrapfigure}[19]{R}{0.46\linewidth}
    \vspace{-56pt}
    \begin{center}
        \includegraphics[width=.99\linewidth]{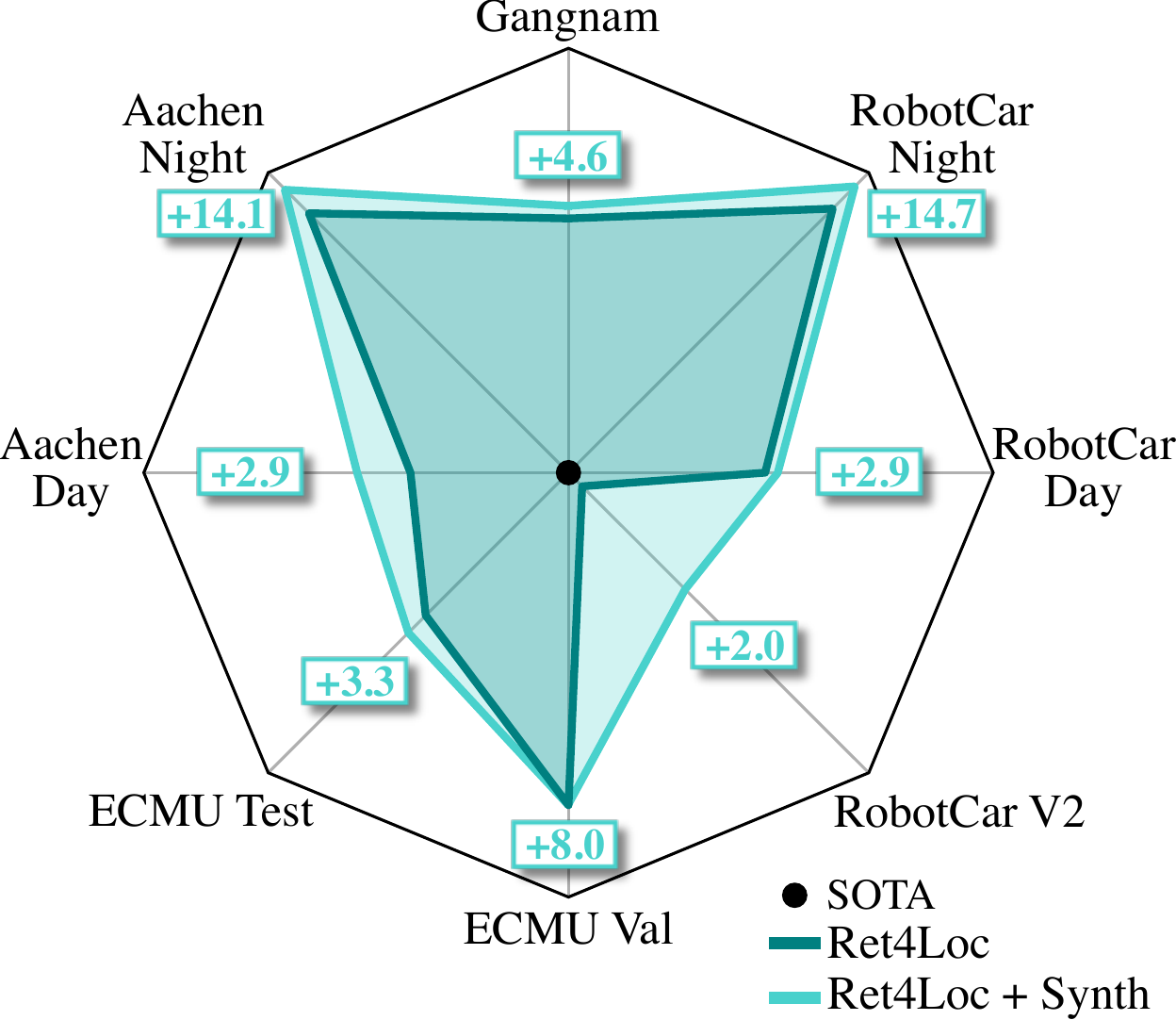}
    \end{center}
    \vspace{-0.5\baselineskip}
    \caption{
        \textbf{Gains in localization accuracy} using our Ret4Loc models compared to the state of the art (\textbf{black dot}). We show results for our best models trained on \textit{only real} ({\color{teal}{\textbf{Ret4Loc}}}) or \textit{real and synthetic images} ({\color{Aquamarine}{\textbf{Ret4Loc + Synth}})}, for 7 outdoor and 1 indoor dataset splits. Axes in log-scale.
    }
    \label{fig:teaser}
\end{wrapfigure}

Visual localization, the task of estimating the camera pose for a novel view of a known scene, is a core component of the perception system of autonomous
platforms. It generally relies on image retrieval techniques that provide an approximate pose estimate that is further refined.
As~\citet{humenberger2022investigating} show, this retrieval step significantly impacts the overall visual localization accuracy,
yet an improved retrieval performance does not necessarily imply a better visual localization.
This is in part explained by the second stage of visual localization;
The pose estimated by the retrieval step is refined \eg using Structure-from-Motion, so it can be,
to some extent, resilient to isolated errors.
Yet, when a challenging visual query leads to retrieved candidates that are wrong in a consistent manner, no recovery is possible. This is why this paper does not study retrieval in isolation, but through the prism of results from the full localization pipeline.

One of the major challenges faced by retrieval models is linked to appearance variations caused by lighting, weather and seasonal changes.
This has been known to the community for a while, as shown by the many methods tailored to one of those variations~(see~\Cref{sec:related}), and is typically referred to as \textit{long-term visual localization}~\citep{sattler2018benchmarking,vonStumbergRAL20GNNetTheGausNewtonLoss4MultiWeatherRelocalization,ToftPAMI22LongTermVisualLocalizationRevisited}.

In this paper, instead of building a custom retrieval model for each single such variation,
we propose to address all of them jointly.
We start from state-of-the-art landmark retrieval and place recognition method HOW~\citep{tolias2013aggregate}
and design retrieval models tailored to the task of visual localization. We will refer to those as \underline{Ret}rieval \underline{for} Visual \underline{Loc}alization or \textit{\textbf{Ret4Loc}} models.
First, we improve the training process and use strong data augmentations, a change proved crucial to the final localization performance. Second, we go beyond generic pixel transformations, and explicitly tackle the domain shifts relevant to the task: the ones related to weather, season, and time of day.

\looseness=-1
To that end, we leverage the fact that challenging conditions for visual localization can be expressed in
\textit{plain text}.
Thanks to impressive progress in generative modeling,
it is now possible to alter
images in a realistic manner using a textual prompt~\citep{brooks2022instructpix2pix}. Here, we directly use the adversarial conditions identified by the community~\citep{ToftPAMI22LongTermVisualLocalizationRevisited} as a set of textual prompts. They correspond to changing illumination (day to dusk/dawn/night), weather (clear to sunny/rainy/snowy), and season (summer to winter).
We use those to generate synthetic variants that extend the
set of images used to train retrieval models, improving their resilience to such changing conditions.

The resulting extended training set is indeed more diverse, and can be used to train models more robust to such conditions.
Yet, off-the-shelf generative models provide no guarantee that these high-level image manipulations
will preserve location-specific information, crucial for visual localization: they might alter the content on top of applying the requested domain shift.
To mitigate this issue, we leverage the fact that
the training relies on image pairs
and propose several geometry-aware strategies
to filter the extended dataset and even to alter the training process itself. 

Ret4Loc is evaluated following two protocols: a faster one that estimates the query's pose using the top retrieved images, and a slower but more accurate one based on Structure-from-Motion that uses the 3D map of the scene.
We show that our retrieval models achieve consistent gains in both cases, for many standard datasets. Fig.~\ref{fig:teaser} summarizes our gains under the first protocol.

\myparagraph{Contributions}
We tailor the training of retrieval models to the challenging task of long-term visual localization. We use language-based data augmentation to improve the robustness of retrieval models to adversarial daytime, weather and season variations.
We further introduce geometry-based strategies for filtering and sampling from the pool of synthetic images and report significant gains over the state of the art on six common benchmarks. To our knowledge, this is the first time language-based generative models are used for generating and validating targeted scene alterations while preserving the crucial visual characteristics needed for an instance-level recognition task.

\section{Ret4Loc: Learning retrieval models for visual localization}\label{sec:method}

\looseness=-1
We start from the state of the art in landmark retrieval and gradually
tailor it to visual localization.
We observe that the training of top models is lacking,
\ie data augmentation and a more adaptive learning framework are missing.
On top of standard pixel-level augmentations, we explore more drastic synthetic variations specifically created to target the domain shifts long identified by the visual localization community.
Next, because the extreme nature of those variations conflicts with the level of precision required in retrieval for localization, we describe a geometric consistency score that can be used for selecting and sampling the synthetic data.

\subsection{Tailoring existing retrieval methods to visual localization}\label{sec:ret4loc}

After benchmarking a broad range of retrieval methods for visual localization, \citet{humenberger2022investigating} exposed the critical role of the retrieval step on the final localization performance.
We extended their study and found that more recent landmark retrieval methods like HOW~\citep{tolias2020learning} and FIRe~\citep{weinzaepfel2022learning} are even more competitive (see Sec.~\ref{sec:exp_main_results}).
We therefore build our retrieval framework on top of HOW which uses a contrastive loss over a training set of matching pairs depicting the same landmark. 
The loss is applied on aggregated global features at training time.
At test time, HOW is typically paired with ASMK~\citep{tolias2013aggregate} matching.

Let us assume access to a training set $\gD$ of images suitable for landmark retrieval with matching pairs, \ie images depicting the same landmark or scene.
Formally, the training set can be seen as a set of training tuples $\gU_{qp}$, each composed of a matching pair $(q, p)$, built from a \textit{query} image and a \textit{positive} image for that query, and a small set of $M$ \textit{negatives} images,
\ie
$\gU_{qp} =  (q, p,  n_1, ..., n_{M} )$.
Let $f(x)$ be the aggregated feature vector for image $x$. It is computed as a weighted average of all local features, the weights being proportional to the local features' $\ell_2$ norm, and then $\ell_2$-normalized. Given a matching pair $(q, p)$ and the corresponding tuple $\gU_{qp}$,
the loss used for training by \citet{tolias2020learning} is given by:
\begin{equation}
\gL_{c}(\gU_{qp}) = 
%\gL_{c}(q, p) = 
\ltnorm{f(q) - f(p)}^2 + \sum_{m=1}^{M} \big[ \mu - \ltnorm{f(q) - f(n_m)}^2 \big]^+,
\label{eq:how_loss}
\end{equation}
where $\mu$ is a margin hyper-parameter, and $[\cdot]^+$ denotes the positive part function $max(0,\cdot)$.

\myparagraph{Towards a better retrieval step for visual localization}
Landmark retrieval methods tend to only use basic data augmentations~\citep{BertonCVPR22DeepVisualGeoLocalizationBenchmark},
while state-of-the-art methods such as HOW~\citep{tolias2013aggregate} do not even use any during training. 
We tailor HOW's training process to long-term visual localization by learning more robust features.
First, we apply domain randomization, that has been shown by~\cite{volpi2021cvpr} to improve resilience to domain shifts. 
For this, we apply a random cropping mechanism and we use AugMix~\citep{hendrycks2019augmix} that mixes multiple augmentations.
Other improvements include using the AdamW~\citep{loshchilov2017decoupled} optimizer and training for longer with a cosine learning rate schedule.\footnote{Further implementation details and hyper-parameters of our training setup can be found in~\Cref{sec:app_implementation_details}.}

We refer to models trained with this setup as \underline{Ret}rieval \underline{for} Visual \underline{Loc}alization or Ret4Loc. 
In this paper we explore Ret4Loc models that follow HOW for training (\textbf{\howam}), but other models, \eg FIRe, could have been used instead.
\howam models boost visual localization performance for both indoor and outdoor benchmarks and hence greatly improve over the state of the art for any type of localization (see results in~\Cref{sec:exp_main_results}). 

Although effective, augmentations described above
are not tailored to visual localization. Next, we leverage the fact that the main challenges faced by long-term visual localization can be \textit{named}.

\subsection{Synthesizing variants for diverse weather \& time conditions}\label{sec:synthetic}

Our goal is to improve the resilience of retrieval models to challenging conditions such as changes related to seasons, weather, and time of day.
Those changes are not only predictable, they can also be clearly and concisely expressed via natural language.
We therefore propose to use generative models to synthesize such challenging scenarios.
DALL-E~\citep{ramesh2021zero} or Stable Diffusion~\citep{rombach2022high} have demonstrated impressive text-to-image generation ability.
Methods like InstructPix2Pix~\citep{brooks2022instructpix2pix} extend them to \textit{altering} images via a textual prompt.

Given a set of textual prompts for scenarios we care about, we employ InstructPix2Pix to generate multiple synthetic \textit{variants} for every image in our training set. % \dlt{$\gD$}.
Formally, let $g(\cdot)$ be a generative model that takes as input an image $x$ and a textual prompt $t$, and produces $\tilde{x}_t$, a synthetic variant of image $x$ with respect to textual prompt $t$, \ie $\tilde{x}_t = g(x; t)$. Given our training set $\gD$ and a set of $T$ {textual prompts} $\gT = \{ t_1,..,t_T \}$,
we generate an extended dataset
which contains the original images as well as their $T$ variants $\tilde{x}_t = g(x; t)$, for every image in $\gD$ and
for all $t \in \{1,..,T\}$.

\begin{figure}
     \centering
     \begin{subfigure}[b]{.51\textwidth}
         \centering
         \includegraphics[width=\textwidth]{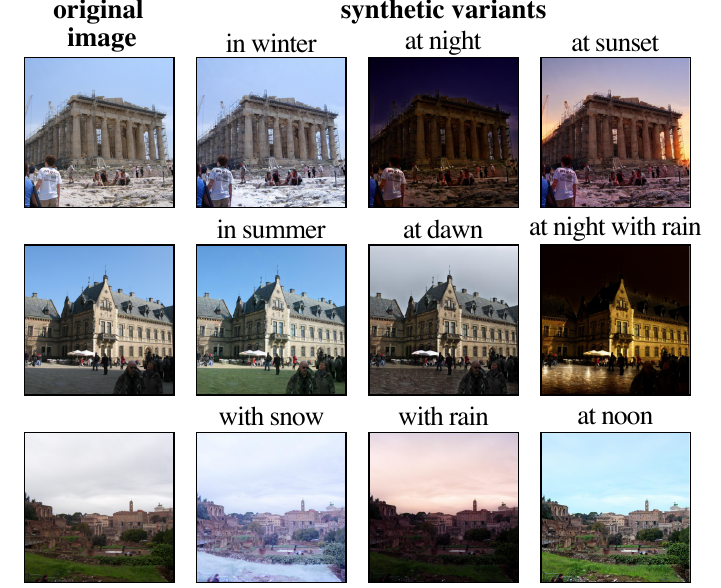}
         \vspace{-8pt}
         \caption{Synthetic variants for the images shown on the left.}
         \label{fig:synthetic_data}
     \end{subfigure}
     % \vspace{12pt}
     \hspace{15pt}
     \begin{subfigure}[b]{.44\textwidth}
         \centering
         \includegraphics[width=.95\textwidth]{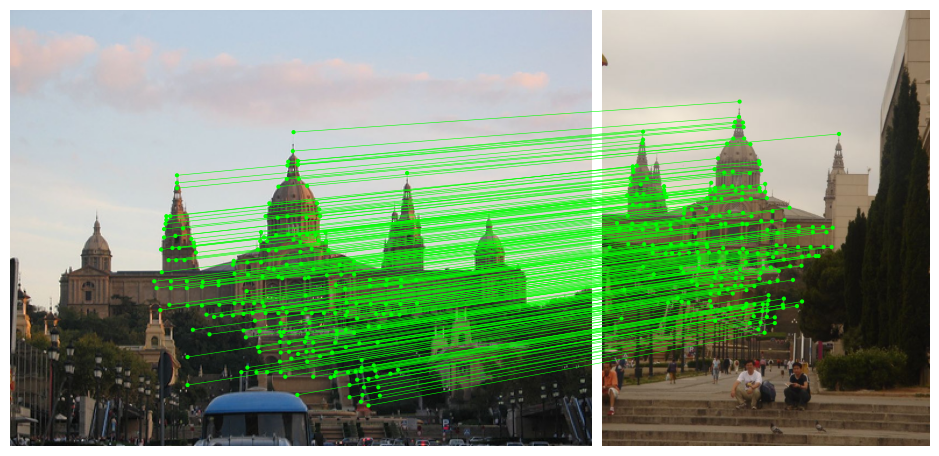}\\
         \includegraphics[width=.95\textwidth]{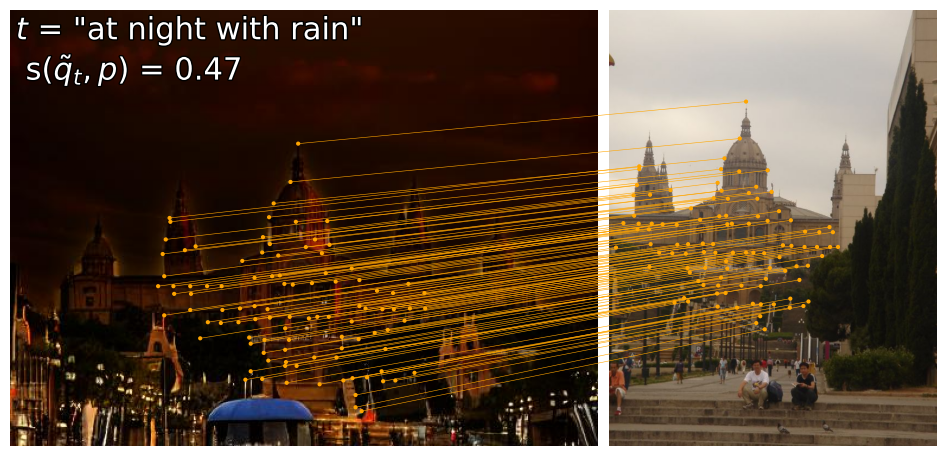}\\
          \vspace{-4pt}
         \caption{Geometric correspondences computed before (top) and after (bottom) synthetic alteration.}
         \label{fig:pair_validation}
     \end{subfigure}
     \vspace{-0.9\baselineskip}
     \caption{(Left) Synthetic variants for several prompts (the full set of variants is shown in Fig.~\ref{fig:all_variants}). (Right) Estimated local correspondences between two matching images before and after alteration.}
    \label{fig:loc_aachen_ecmu}
\end{figure}

The seminal Robotcar Seasons visual localization benchmark~\citep{maddern20171} has identified the most common domain shifts for long-term visual localization.
We use those to define a set of 11 textual prompts to create synthetic variants for our training set:
\texttt{`at dawn'}, \texttt{`at dusk'}, \texttt{`at noon'}, \texttt{`at sunset'}, \texttt{`in winter'}, \texttt{`in summer'}, \texttt{`with rain'}, \texttt{`with snow'}, \texttt{`with sun'}, \texttt{`at night with rain'}, \texttt{`at night'}.
Fig.~\ref{fig:synthetic_data} shows a few generated variants for 3 images from the training set. The complete set of 11 variants can be found in~\Cref{sec:app_synthetic}.

\subsection{Training with synthetic variants}
\label{sec:training}

Once we have generated multiple variants for all training images, we sample them to form \textit{synthetic tuples} that we use during training together with the original ones. 
To obtain a synthetic tuple, we propose to substitute the query image $q$ and all negatives of an original training set tuple with their synthetic variants.
Note that the positive image $p$ is never substituted and therefore the matching pair of each synthetic tuple contains both an altered and an original image.
To make sure negatives in the tuple remain challenging, we choose variants consistently across a tuple, \ie
we use variants from the same textual prompt for the query and the negatives.

Formally, given a tuple $\gU_{qp}$ corresponding to the matching pair $(q,p)$, we produce a synthetic tuple 
$\tilde{\gU}^t_{qp} = ( \tilde{q}^t, p, \tilde{n}^t_1, ...,  \tilde{n}^t_M) $   
for the variant corresponding to textual prompt $t$
by replacing $q$ with $\tilde{q}_t$ and all the negatives $n_m$ with $\tilde{n}_m^t$. 
Intuitively, this means that, when a synthetic tuple is used in the contrastive loss of HOW (\Cref{eq:how_loss}), the first part of the loss, which uses a synthetic $\tilde{q}^t$  with an unaltered $p$, brings the representations of the original positive image $p$ and of the synthetic variant of the query $\tilde{q}^t$ close to each other. At the same time, the loss pushes the synthetic variant of the query $\tilde{q}^t$ away from the representations of all the synthetic negative images $\tilde{n}_m^t$, in the domain corresponding to the selected textual prompt. 
This aligns with our intuition that i) the query feature should be invariant to the different domain shifts described by the prompts and simultaneously ii) given any domain shift, the query feature should be different enough from its associated negatives. 

\looseness=-1
We experimented with several ways of using synthetic tuples during training. 
The most effective one uses both the original tuple and one or more synthetic tuples sampled from the set of possible ones for the different prompts.  
By default, we uniformly sample $K$ textual prompts to select
synthetic tuples among $T$ options (we discuss a geometry-based sampling alternative next in~\Cref{sec:geometric}).
Formally, let us consider a set of $K$ synthetic tuples $\{ \tilde{\gU}^1_{qp}, ..., \tilde{\gU}^K_{qp} \}$ corresponding to  $K \geq  1$ selected variants. 
We can define the set of tuples  $\tilde{\gU} = \{ \gU_{qp}, \tilde{\gU}^1_{qp}, ..., \tilde{\gU}^K_{qp} \}$ that contains the original tuple $\gU_{qp}$ and all synthetic ones. The loss is then computed over this extended set of tuples.

\myparagraph{Loss computation with synthetic tuples (\Synth)}
A simple way to compute the overall loss for the extended tuple set $\tilde{\gU}$ would be to sum the individual losses incurred from the real and synthetic tuples, \ie $\gL_{c}(\tilde{\gU}) =\gL_c(\gU_{qp}) + \sum_k \gL_c(\tilde{\gU}^k_{qp})$.
An alternative that performs better in practice is to first aggregate features from all variants of each image in the original tuple independently, and then compute a single loss on the aggregated features.
More precisely, let $Q,P,N_m$ respectively be sets of features extracted for 
the corresponding  query, positive\footnote{Although we do not substitute $p$ with synthetic variants, we do sample different augmentations of the positive for every synthetic tuple and still end up with a set of features in P that we can aggregate.} 
and each of the negative images 
that appear in the extended tuple set $\tilde{\gU}$. 
Let $\phi$ be an aggregation function, \eg simple averaging.
The function $\phi$ 
produces a single aggregated feature vector for each set of features %$Q/P/N_m$. 
$Q$, $P$, and $N_m$. The loss then can be applied on the aggregated vectors, \ie:
\begin{equation}
\gL_{c}(\tilde{\gU}) = \ltnorm{\phi(Q) - \phi(P)}^2 + \sum_{m=1}^{M} \big[ \mu - \ltnorm{\phi(Q) - \phi(N_m)}^2 \big]^+.
\label{eq:emb_avg}
\end{equation}
In this loss, all original and synthetic query, positive and negative images impact each other, \ie tuples are not treated independently anymore. 
By default we use the loss on the aggregated features, \ie the loss from \Cref{eq:emb_avg}, for Ret4Loc models. We ablate this choice in the experiments.

\subsection{Incorporating geometric constraints for synthetic variants}
\label{sec:geometric}

In this section we describe an automatic way of assessing to which extent important characteristics of the scene are retained
during the image generation process, and propose two strategies that incorporate this information 
during training.
We argue that the geometric consistency of a matching pair should not change significantly when the query image is replaced by a synthetic variant. This means that geometric correspondences estimated between a pair should mostly remain after altering one of the images with a generative model. Using this insight, we define a \textit{geometric consistency score} to assess the degree to which a synthetic pair
preserves location characteristics shared across the matching images. 

Let $(q,p)$ be a matching pair from the training set, and $(\tilde{q}^t, p)$ its corresponding synthetic pair where the query is replaced by a synthetic variant obtained with textual prompt $t$. 
Let $c(\cdot,\cdot)$ denote a 
matching function that returns a set of \textit{correspondences} $C_{qp}$, \ie a set of geometrically consistent local matches between $q$ and $p$. We can also compute correspondences $C_{\tilde{q}^tp}$ for the synthetic pair and then filter them such that only correspondences that existed in the original pair $(q,p)$ are kept: $||C_{\tilde{q}^tp}||^\prime = ||C_{\tilde{q}^tp}|| \cap ||C_{qp}||$. We propose to use the ratio of correspondences remaining after replacing $q$ by $\tilde{q}^t$ in the tuple, $s(\tilde{q}^t,p) = \frac{||C_{\tilde{q}^tp}||^\prime}{||C_{qp}||}$, as a geometric consistency score. This score ranks synthetic pairs according to the level of preservation of their geometry.
In \Cref{fig:pair_validation} we show correspondences between images of a matching pair, before and after replacing the query with one synthetic variant.
Next, we present two ways of exploiting this score during training.

\myparagraph{Synthetic tuple filtering (\Synthp)} 
We can directly use the geometric consistency score above to select the synthetic tuples that are used during training. The score is in $[0,1]$ and essentially measures the percentage of correspondences remaining for the matching pair after the query is altered. 
One can set a threshold $\tau$ under which a synthetic tuple will be discarded; we otherwise refer to it as \textit{valid}. 
Increasing the threshold means that we are being more and more selective.
We can therefore restrict the sampling of tuples during training to only pick among the valid ones, \ie when sampling synthetic tuples for an original training tuple.

\looseness=-1
\myparagraph{Geometry-aware sampling (\Synthpp)}
When choosing synthetic tuples, by default, we sample a textual prompt $t \in \gT$ uniformly.
However, we now have a filtered set of tuples, each with a geometric consistency score $s$ 
which correlates with the level of local appearance preservation.
We propose to use this score to also compute the probability of sampling variants.
Sampling a valid synthetic tuple $\tilde{\gU}^t_{qp}$
proportionally to %$1\frac{1}{s(\tilde{q}^t,p)}$ 
$1/s(\tilde{q}^t,p)$ means that tuples with a larger drop in correspondences are picked more often; This weighing scheme favors valid tuples that are ``harder''. We refer to this as \textit{geometry-aware sampling}. Note that this only works well in conjunction with tuple filtering,
as variants with very low geometric consistency should never be considered.

\section{Experiments}\label{sec:experiments}

In this section, we evaluate our Ret4Loc models
on visual localization and place recognition datasets.
Two of those are indoor datasets while the others are outdoor ones.
Although part of our contributions explicitly targets outdoor localization, we show that Ret4Loc also leads to state-of-the-art results for indoor localization.
We present extended results for place recognition in~\Cref{sec:app_retrieval_results}.

\myparagraph{Datasets}
Following HOW~\citep{tolias2020learning}, we use the SfM-120K dataset~\citep{radenovic2018fine} to train all Ret4Loc models.
We augment SfM-120k by generating 11 synthetic variants for each image with the process described in~\Cref{sec:synthetic} and use them to build synthetic tuples that are used during training.
We evaluate localization accuracy on five datasets from the Visual Localization Benchmark,\footnote{\url{https://www.visuallocalization.net/}} \ie RobotCar Seasons~\citep{maddern20171}, Aachen Day-Night v1.1~\citep{ZhangIJCV21ReferencePoseGenerationVisLoc}, Extended CMU Seasons~\citep{ToftPAMI22LongTermVisualLocalizationRevisited},
Gangnam Station B2~\citep{lee2021large} and Baidu Mall~\citep{sun2017dataset}. For completeness, we also evaluate place recognition on the popular Tokyo 24/7 dataset~\citep{Torii-CVPR2015}.

\myparagraph{Synthetic variant generation and geometric consistency} We use the public InstructPix2Pix model from~\citet{brooks2022instructpix2pix} for generating synthetic variants.\footnote{\url{https://github.com/timothybrooks/instruct-pix2pix}}
We use 20 inference sampling steps for diffusion, and set the text prompt and image guidance scale to 10 and 1.6, respectively.
We ran synthetic image generation as a pre-processing step, for all prompts in parallel. 
This took approximately a day (more details in~\Cref{sec:app_timings}).
We employ LightGlue~\citep{lindenberger2023lightglue}
as our matching function $c$ to estimate the geometric consistency score $s(\cdot,\cdot)$ that is used for filtering or sampling synthetic variants.

\myparagraph{Training details}
We build on the HOW codebase\footnote{\url{https://github.com/gtolias/how}} and, unless otherwise stated, we follow the training protocol from HOW with the modifications presented in~\cref{sec:ret4loc}.
We train ResNet50 models~\citep{he2016resnet}.
We use randomly resized crops of size $768\times512$ during training as additional data-augmentation as well as to provide the network with fixed sized inputs.
After preliminary experiments, we found that using the original plus two synthetic variants ($K$=2) perform best.
Our model trains in less than 8 hours on a single A100 GPU (more details in~\Cref{sec:app_timings}).

\myparagraph{Evaluation protocols}
At test-time, we observe a large variance in performance when running the exact same training setup with different seeds.
To get more robust results, we report results after averaging the weights of 3 models for each setup, \ie we average the weights of 3 models trained
from 3 different seeds~\citep{wortsman2022model}.
We follow HOW and use multi-scale queries together with ASMK matching.
To measure visual localization performance, we use the Kapture localization toolbox.\footnote{\url{https://github.com/naver/kapture-localization}}
In the retrieval step of this toolbox, a retrieval model produces a shortlist of the top-$k$ nearest database images per query. For our experiments, we replace the retrieval model with the public HOW or FIRe models, or one of the different Ret4Loc flavors we designed. %want to evaluate.
The rest of the pipeline, \ie the pose estimation step, is shared across all experiments, and follows two of the protocols presented by~\citet{humenberger2022investigating}: a) pose approximation with equal weighted barycenter (\textit{EWB}) using the poses of the top-$k$ retrieved database images; or b) pose estimation with a global 3D-map (\textit{SfM}) and R2D2~\citep{revaud2019r2d2} local features.
More implementation and protocol details are presented in~\Cref{sec:app_implementation_details}.

{
\setlength{\tabcolsep}{2pt}
\setlength{\tabcolsep}{3pt}
\setcounter{rownumbers}{0}

\def\colgroupspace{{\hskip 7pt}}
\def\colgroupspacelarge{{\hskip 8pt}}

\begin{table}[t]
    \centering
    \adjustbox{max width=\textwidth}{
    \begin{tabular}{rl @{\colgroupspace} ccc @{\colgroupspacelarge} cc @{\colgroupspace} cc @{\colgroupspace} cc @{\colgroupspace} cc @{\colgroupspace} cc}
    \toprule
    & \multirow{3}{*}{Model} &
        \multirow{3}{*}{VMix} &
        \multirow{3}{*}{Filt.} &
        \multirow{3}{*}{GaS} &
        \multicolumn{2}{c}{\textbf{RobotCar-v2}} &
        \multicolumn{2}{c}{\textbf{ECMU-val}} &
        \multicolumn{2}{c}{\textbf{RobotCar-Day}} &
        \multicolumn{2}{c}{\textbf{RobotCar-Night}} &
        \multicolumn{2}{c}{\textbf{Tokyo 24/7}} \\
        &&&&&
         \multicolumn{2}{c}{EWB (top-1)} &
         \multicolumn{2}{c}{EWB, ($\max$)} &
         \multicolumn{2}{c}{SfM, ($\max$)} &
         \multicolumn{2}{c}{SfM, ($\max$)} &
         \multicolumn{2}{c}{Recall@$k$} \\
    & &  & & &
        \small{.5m/5\degree} & \small{5m/10\degree} &
        \small{.5m/5\degree} & \small{5m/10\degree} &
        \small{.25m/2\degree} & \small{.5m/5\degree}  &
        \small{.25m/2\degree} & \small{.5m/5\degree}  &
        R@1 & R@10  \\
    \midrule

    % ----------------------------------------------------------------
    % row 4 - HOW
    % ----------------------------------------------------------------
    \rownum & HOW & -- & -- & --
    & 29.8 & 74.4
    & 25.9  & 81.1
    & 52.6  & 81.1
    & 17.4  & 35.0
    & 89.2 & 96.5 \\
    % ----------------------------------------------------------------
    \cmidrule{2-15}
    % ----------------------------------------------------------------

    \rownum & \howam  & -- & -- & --
    & 30.6 & 75.0
    & 33.4  & 88.6
    & 52.8  & \textbf{81.5}
    & 21.8  & 40.4
    & 89.2 & \textbf{97.1} \\

    % ----------------------------------------------------------------
    \cmidrule{2-15}
    % ----------------------------------------------------------------

    \rownum & \multirow{6}{*}{\shortstack[c]{Ret4Loc-HOW \\ + synthetic}}  & -- & -- & --
    & 31.5 & 	75.9
    & 33.4  & 88.0
    & 52.8  & 81.1
    & \textbf{25.1}  & \textbf{47.1}
    & 88.9 & 96.2 \\

    \rownum && \cmark & -- & --
    & 31.7 & 76.3
    & 34.3 & 88.6
    & 52.7 & 81.2
    & 23.0  & 43.6
    & 90.5& 96.2 \\

    % ----------------------------------------------------------------
    \cmidrule{4-15}
    % ----------------------------------------------------------------
    &&& \multicolumn{12}{l}{\emph{Variations using geometric consistency}} \vspace{2pt} \\

    \rownum &  & \cmark & \cmark & --
    & \textbf{31.8}	 & \textbf{77.1}
    & 34.4  & 88.8
    & 52.8  & \textbf{81.5}
    & 22.1  & 43.7
    & 90.2 & 96.2 \\

    \rownum && \cmark & \cmark & \cmark
    & 31.3 & 	76.4
    & \textbf{34.5}  & \textbf{89.1}
    & \textbf{53.0}  & 81.4
    & 21.0  & 41.7
    & \textbf{91.1} & \textbf{97.1}\\

    \bottomrule
    \end{tabular}
    }
    \caption{
        \textbf{Impact of synthetic data and geometric consistency}. We report results 
        for different flavors of Ret4Loc training with and without the use of synthetic variants and geometric consistency. \textit{VMix} refers to the use of variant mixing with~\cref{eq:emb_avg} instead of summing the different losses.
        \textit{Filt.} refers to synthetic tuple filtering,  \textit{GaS} to geometry-aware sampling.
        Columns denoted as \textit{max} report the top performance achieved across all different values of top-$k$, \ie the best across $k \in \{1,\ldots,50\}$.
        }
    \label{tab:ablations}
\end{table}
}

\subsection{Results}
\label{sec:exp_main_results}

\begin{figure}[t!]
    \centering
    \includegraphics[width=0.99\linewidth]{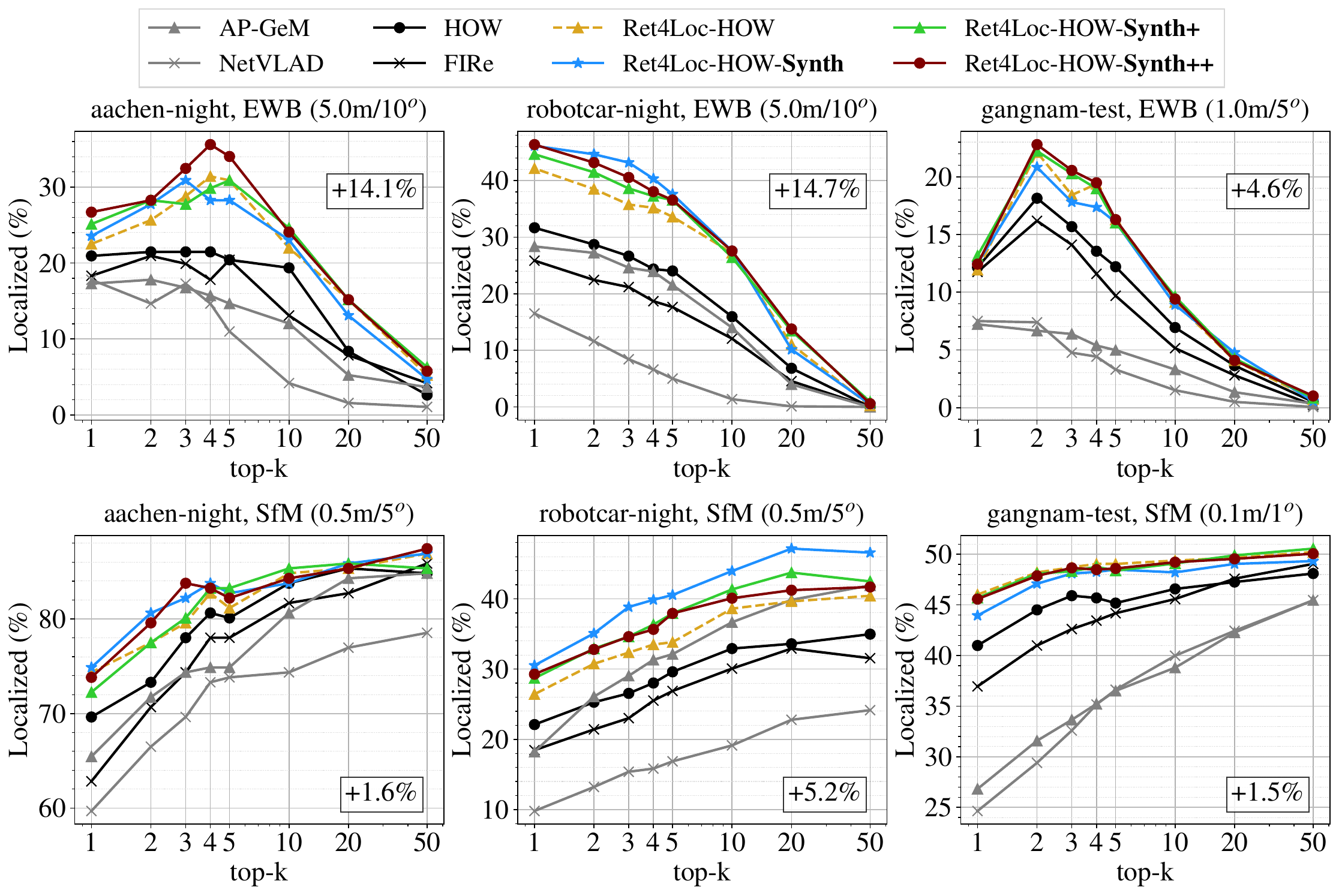}
    \caption{
        \textbf{Localization accuracy as a function of the top-$k$ retrieved images} for
        Ret4Loc models and the state of the art.
        \textit{Top}: Pose approximation (EWB) protocol.
        \textit{Bottom}: Structure-from-Motion (SfM) based protocol.
        \Synth variations using geometric consistency are denoted with a "+".
        In each plot, we further denote the top gains achieved using Ret4Loc models over the best performing competing method.
        See~\cref{sec:app_extended_results} for a complete set of results on more datasets.
    }
    \label{fig:kapture_results}
    \vspace{-13pt}
\end{figure}

In~\Cref{tab:ablations} we present results for several \retforloc models trained under different setups.
We see that the basic \retforloc setup (row 2) already brings some consistent gains over HOW.
Using synthetic variants during training further improves performance even without any filtering (rows 3-4). We also see that mixing the variants generally improves performance (rows 3 \vs 4) in all datasets apart from RobotCar-Night.
Inspecting our complete set of results on 6 datasets, overall we see consistent gains when incorporating synthetic data during training.
From rows 5 and 6, it appears that we can get the top performance in all datasets except RobotCar-Night by incorporating geometric information, either as a way of filtering or also sampling the synthetic variants.

\begin{minipage}[t]{\textwidth}
    \begin{minipage}[b]{0.59\textwidth}
        \centering
        {
\setlength{\tabcolsep}{3pt}
\setcounter{rownumbers}{0}
\def\colgroupspace{{\hskip 5pt}}

    \centering
    \adjustbox{max width=\textwidth}{
    \begin{tabular}{rlccc @{\colgroupspace} ccc}
    \toprule
    & \multirow{3}{*}{Model} & 
        \multicolumn{2}{c}{\textbf{RobotCar-v2}} &
        \multicolumn{2}{c}{\textbf{ECMU-test}} &
        \multicolumn{2}{c}{\textbf{Tokyo 24/7}} \\
    & & 
        \multicolumn{2}{c}{\small{EWB (top-1)}} &
        \multicolumn{2}{c}{\small{EWB (top-1)}} &
        \multicolumn{2}{c}{Recall@$k$} \\
    & & 
        \small{.5m/5\degree} & \small{5m/10\degree} & 
        \small{.5m/5\degree} & \small{5m/10\degree} &
            R@1 & R@10  \\
    \midrule
    \multicolumn{8}{l}{\emph{Single stage retrieval methods}} \\
    
    \rownum & GCL$^*$ 
    & 21.9 & 74.7 
    & 18.2 & 74.9
    & 69.5 & 85.1 \\

    \rownum & FiRE$^\ddagger$ 
    & 26.0 & 74.0  
    & 24.4 & 83.6 
    & 84.8	& 92.7 \\
    
    \rownum & HOW$^\ddagger$ 
    & 29.8 & 74.4 
    & 30.5 & 87.9
    & 89.2 & 96.5 \\ 
    
    \cmidrule{2-8}
    
    \rownum & \howam$^\ddagger$ 
    & 30.6	& 75.0	
    & 32.6	&90.5 
    & 89.2 & \textbf{97.1} \\ 

    \rownum & \Synthp$^\ddagger$ 
    & \textbf{31.8}	 & \textbf{77.1}	 
    & 34.1 & 91.0 
    & 90.2 & 96.2 \\ 

    \rownum & \Synthpp$^\ddagger$ 
    & 31.3 & 76.4 
    & \textbf{34.2}	 & \textbf{91.2}
    & \textbf{91.1} & \textbf{97.1}\\ 
    
    \midrule
    \multicolumn{8}{l}{\emph{Retrieval methods \textbf{with re-ranking}}} \\
    
    \rownum & SP-SuperGlue$^{*}$  
    & \underline{35.4} & 85.4 
    & 30.7 & \underline{96.7}  
    & 88.2 & 90.2 \\

    \rownum & DELG$^{*}$ 
    & 8.4 & 76.8 
    & 21.1 & 93.6
    & \underline{95.9} & \underline{97.1}\\
    
    \rownum & Patch NetVLAD$^{*}$ 
    & 35.3 & \textbf{90.9} 
    & \underline{36.2} & 96.2 
    & 86.0 & 90.5 \\
    
    \cmidrule{2-8}
    
    \rownum & \Synthpp$^{\ddagger}$ 
    & \textbf{38.6} & \underline{84.2}
    & \textbf{42.4}	 & \textbf{98.3}
    & \textbf{97.5} & \textbf{98.4}\\

    \bottomrule
    \end{tabular}
    }
}
        \captionof{table}{\textbf{Comparison to the state of the art.} 
            $^\ddagger$~denotes methods using ASMK. $^*$~denotes results from GCL.}
        \label{tab:sota}
    \end{minipage}
    \hfill
    \begin{minipage}[b]{0.39\textwidth}
        \centering
        
    \centering
    \adjustbox{max width=\textwidth}{
    
        \begin{tabular}{@{}lccc@{}}
            \toprule
            Method              & Purity & ARI & NMI \\ \midrule
            HOW                 & 0.82 & 0.75 & 0.95 \\
            Ret4Loc-HOW         & 0.85 & 0.78 & 0.96 \\
            Ret4Loc-HOW-Synth   & 0.89 & \textbf{0.85} & \textbf{0.97} \\
            Ret4Loc-HOW-Synth+  & \textbf{0.90} & \textbf{0.85} & \textbf{0.97} \\
            Ret4Loc-HOW-Synth++ & {0.89} & \textbf{0.85} & \textbf{0.97} \\ \bottomrule
        \end{tabular}
    }
        \captionof{table}{
            \textbf{Feature space analysis}.
            We measure how well an image and its synthetic variants get clustered in the different Ret4Loc feature spaces. 
            We sample 1000 images from SfM-120k,
            consider each image and its 11 variants a ``class'' and cluster their features into 1000 clusters.
            We report purity, adjusted random index (ARI) and normalized mutual information (NMI).
        }
        \label{tab:clustering}
    \end{minipage}

\end{minipage}

\looseness=-1
To complement \Cref{tab:ablations}, \Cref{fig:kapture_results} presents localization results across multiple top-$k$ results. Here, we only compare different variants of our method
to HOW and FIRe, as they already surpass the older methods evaluated by~\citet{humenberger2022investigating}. We clearly see the superiority of all Ret4Loc models across protocols and top-$k$ values, with gains exceeding $10\%$ in localization accuracy for the case of night queries. The right-most plots show results on Gangnam Station B2~\citep{lee2021large}, a dataset for \textit{indoor} localization. We see that strong augmentations help indoor localization as well. It is also nice to observe that the addition of synthetic data tailored to outdoor domain shifts does not hurt indoor performance, making
Ret4Loc 
\textit{the state-of-the-art retrieval method for both indoor and outdoor localization}.
Complete sets of results are presented in~\Cref{sec:app_extended_results}.

\looseness=-1
\myparagraph{Comparison to the state of the art}
We found two methods designed for landmark image retrieval, HOW~\citep{tolias2020learning} and FIRe~\citep{weinzaepfel2022learning}, to be state-of-the-art for the localization benchmarks we considered, a finding also confirmed by~\citet{aiger2023yes}. We also compare Ret4Loc to recent place recognition methods like GCL~\citep{leyva2023data}.
and TransVPR~\citep{wang2022transvpr}. 
We also report results after a complementary geometric re-ranking step, \ie after re-ranking the top 100 retrieval results from our top Ret4Loc model using LightGlue~\citep{lindenberger2023lightglue}. We compare to SP+SuperGlue~\citep{sarlin20superglue}, DELG~\citep{cao2020unifying} and Patch NetVLAD~\citep{hausler2021patch}. Details are provided in~\Cref{sec:rerank}.

In~\Cref{tab:sota} we compare \retforloc to the most recent published results on visual localization. For this reason, we restrict the comparison to the protocols used in related works, \ie EWB and top-1. We see that our method
outperforms by a large margin the very recent GCL,
a method also using contrastive learning.
This means that,
in theory, training with synthetic variants could also benefit GCL.
We considered methods with reranking in a separate %section as
part of the table as they benefit from an additional reranking step, that could be applied to
all single stage retrieval methods. We also clearly mark methods that use ASMK matching, that seems to help for the task. Concurrent work by~\citet{aiger2023yes} replaces ASMK matching with CANN and shows large gains for HOW and FIRe. We see CANN as orthogonal and complementary to Ret4Loc training.

In~\Cref{fig:teaser} we show a birds-eye view of the significant gains over the state of the art (SOTA) one can get with and without the use of synthetic data, \ie using the \Synthpp
and \retforloc models, respectively. We denote as SOTA
the best performance per dataset across all single stage retrieval methods,
\ie in practice either HOW or FIRe. We report results using the EWB protocol, for the highest accuracy threshold
and show the radial axes in log scale.

\section{Discussion and analysis}
\label{sec:discussion}

\myparagraph{Feature space analysis}
To better understand how the representation space changes from the baseline HOW to \retforloc, and
by using synthetic data and geometry, we performed some feature analysis on a subset of SfM-120k validation images and all their synthetic variants.
We measure alignment and uniformity~\citep{wang2020understanding} for features from different retrieval models (\cref{fig:alignment_uniformity} in the Appendix) and observe that the use of synthetic data leads to a better balance between these
two quantities, as opposed to \howam and to the baseline HOW.
We also performed a clustering analysis (\cref{tab:clustering}) and found that, among other metrics, cluster purity improves by $2.5\%$ in \retforloc over the baseline, and then by 3-4\% more when using synthetic data and geometry.
More details on this analysis can be found in \cref{sec:app_feature_analysis}.

\looseness=-1
\myparagraph{Potential alternatives to the generation of synthetic variants}
The textual prompts we use are concise and
only contain one main alteration. Following recent works on prompt engineering, our method could be extended by creating more intricate prompts~\citep{zhou2022learning};
we find such explorations beyond the scope of this paper.
We use InstructPix2Pix, but any generative model that can alter an image using a textual prompt could be used instead. The set of prompts can be replaced or extended to include other domain shifts as long as they can be expressed in natural language.

\myparagraph{Synthetic variants and indoor localization}
From~\cref{sec:exp_main_results} we see that the synthetic variants created with outdoor localization in mind also benefit indoor localization datasets.
This is consistent with findings that extreme data augmentation helps generalization~\citep{sariyildiz2023improving,volpi2021cvpr}.
It is worth mentioning that the SfM-120k dataset also contains images taken inside famous landmarks
(see examples of synthetic variants in~\Cref{fig:all_variants_fail} in Appendix). Ignoring a few obvious failure cases we discuss below, the resulting images are not completely unrealistic, \ie time and weather changes are a specific case of scene illumination changes.
We found that geometric consistency is often preserved after replacing an image with a synthetic variant (see \Cref{fig:geom_indoor} in Appendix).

\myparagraph{Failure cases for synthesis}
We observed a number of failures in the image generation process, the majority for indoor images, as expected (see some such cases in the middle and bottom examples of~\Cref{fig:all_variants_fail} in the Appendix).
From~\Cref{tab:ablations}, we see that keeping such images during training does not significantly hurt the model's performance.
We believe this is because, even in such cases, part of the instance (\eg the door in the middle example) is still visible, even in the most distorted variants.

\myparagraph{Limitations} As discussed by~\citet{sariyildiz2023fake}, using generic generative models for specific vision tasks can be seen as performing targeted distillation given a set of textual prompts.
This means that success is generally bounded by the expressivity of the generative models and their ability to generate useful alterations for the provided textual prompts.
\section{Related Work}\label{sec:related}

We discuss here the most related works. A broader discussion can be found in Appendix Sec.~\ref{sec:AppRW}.

\myparagraph{Synthetic data for improved visual understanding}
Most computer vision tasks are expected to be resilient to some image appearance variations, \eg day to night, weather, and seasonal changes. This is typically handled with data augmentation~\citep{
% zhang2017mixup, 
hendrycks2019augmix,YunICCV19CutMixRegStrategyToTrainStrongClassifWithLocFeats} and sometimes with more complex physical models. The latter synthetically modify the training images towards specific targeted weather or time of day conditions. They for instance add synthetic fog \citep{KahramanTR16InfluenceOfFogOnComputerVisionAlgorithms,SakaridisIJCV17SemanticFoggySceneUnderstandingWithSyntheticData}, rain \citep{HalderICCV19PhysicsBasedRendering4ImprovingRobustness2Rain,TremblayIJCV21RainRendering4EvaluatingAndImprovingRobustness2BadWeather,HuCVPR19DepthAttentionalFeaturesSingleImageRainRemoval} or darken images \citep{LengyelICCV21ZeroShotDayNightDAWithPhysicsPrior}.
Another strategy is to remove weather-related artifacts such as snow \citep{RenCVPR17VideoDesnowingAndDerainingBasedOnMatrixDecomposition}, raindrop~\citep{QianCVPR18AttentiveGAN4RaindropRemoval,YouPAMI16AdherentRaindropModelingDetectionAndRemovalInVideo}, haze~\citep{CaiTIP17DehazeNetAnEnd2EndSystem4SingleHazeRemovaly,LiICCV17AODNetAllInOneDehazingNetwork,LiCVPR18SingleImageDehazingViaConditionalGenerativeAdversarialNetwork,RenCVPR18GatedFusionNetwork4SingleImageDehazing,ZhangCVPR18DenselyConnectedPyramidDehazingNetwork}, or darkness~\citep{JenicekICCV19NoFearOfTheDark,WuCVPR21DANNetOneStageDANetworkUnsupervisedNighttime,mohwald2023dark}.
Closer to our work, \cite{FahesICCV23PODAPromptDrivenZeroShotDA} uses a textual description of the target domain to perform domain adaptation for segmentation.

With the recent success of generative models, producing \citep{rombach2022high,ZhangICCV2AddingConditionalControl2T2IDiffusionModels,SahariaX22PhotorealisticText2ImageDiffusionModelsWithDeepLanguageUnderstanding} 
and modifying 
\citep{brooks2022instructpix2pix,NicholICML22GLIDETowardsPhotorealisticImageGenerationEditingDiffusionModels,BarronCVPR23NULLTextInversion4EditingRealImagesUsingGuidedDiffusionModels,ParmarSIGGRAPH23ZeroShotImage2ImageTranslation,dunlap2023alia} images became within reach of even inexperienced users who can simply describe the desired content or changes in natural language.
Such tools have been used to extend \citep{dunlap2023alia,TrabuccoX23EffectiveDataAugmentationWithDiffusionModels} or even replace~\citep{he2023synthetic,azizi2023synthetic,sariyildiz2023fake} training image collections. 
We position our work on the continuity of this line of research, and extend our dataset using textual prompts relevant to localization.

\myparagraph{Visual localization under challenging conditions}
Image retrieval is often used as an initial step  for structure-based visual localization~\citep{humenberger2022investigating} as it limits the search range in the 3D maps~\citep{HumenbergerX20RobustImageRetrievalBasedVisLocKapture,TairaPAMI21InLocIndoorVisualLocalization,SarlinCVPR19FromCoarsetoFineHierarchicalLocalization} and it even provides an approximate localization that is sufficient for place recognition~\citep{ZamirB16LargeScaleVisualGeoLocalization,LowryTROB16VisualPlaceRecognitionSurvey,BertonCVPR22DeepVisualGeoLocalizationBenchmark}.
However, visual localization  methods, including their retrieval component, 
need to be resilient to image appearance changes due to day to night changes  as well as weather and seasonal variations. 
This problem is called \textit{long-term visual localization}~\citep{sattler2018benchmarking,vonStumbergRAL20GNNetTheGausNewtonLoss4MultiWeatherRelocalization,ToftPAMI22LongTermVisualLocalizationRevisited}. 
On top of standard data augmentation  or photometric normalization~\citep{JenicekICCV19NoFearOfTheDark} applied to retrieval models~\citep{revaud2019learning}, domain adaptation is sometimes used. %applied. 
It considers a single domain shift, \eg day-\vs-night~\citep{AnooshehICRA19NightToDayImageTranslationLocalization,ZhengECCV20ForkGANSeeingIntoRainyNight},
or several ones~\citep{HuTIP21DASGILDA4SemanticAndGeometricAwareImageBasedLocalization}. Tailored approaches have also been proposed. 
\citet{PoravITSC19DontWorryAboutWeather} learn weather-specific adapters while \citet{XinICRA19LocalizingDiscriminativeVisualLandmarksPlaceRecogn, LarssonICCV19FineGrainedSISSelfSupISVL} build from aligned image pairs composed of the exact same scenes observed under different conditions. 
Such requirement is rarely possible in practice.
In contrast, we show that a tailored augmentation of the training set using both simple image processing~\citep{hendrycks2019augmix} as well as generative models with simple textual prompts can be automated and greatly improves performance on long-term visual localization benchmarks~\citep{sattler2018benchmarking}. 
Another way to improve robustness to such variations is to leverage semantics during pose approximation, \eg, define semantic-aware detectors which favor reliable regions~\citep{XueCVPR23SFD2SemanticGuidedFeatureDetectionDescription}. These methods are orthogonal to any improvement to the retrieval step, such as the ones we propose in this paper.

\section{Conclusions}\label{sec:conclusions}

In this paper we propose to harness the power of recent text-to-image generative models and direct it towards specific domain shifts that affect visual localization. 
Generative models are extremely well suited for this task.
First, unlike many other vision tasks, most of the challenges impacting the model can be expressed in natural language.
Second, the fact that we are provided with matching pairs for training 
can be {used} for automatic quality control. This is all the more important
as generative models are known to fail in unpredictable manner.
To our knowledge, this is the first time such generative models are used for producing and validating targeted scene alterations.
Finally, it is worth noting that our contributions are not handcrafted towards localization, and hence we can envision applications to other vision tasks.

\newpage
\myparagraph{Acknowledgments} The authors would like to sincerely thank Martin Humenberger, Jerome Revaud and Giorgos Tolias for many helpful discussions, as well as Yohann Cabon for providing us with Kapture evaluation scripts and datasets.

\myparagraph{Ethics Statement} 
Our method automatically generates and uses altered versions of the training set images for training retrieval models.
Those altered variants are automatically produced by generative models, using textual prompts. Our prompts were explicitly targeting seasonal, weather and time of day changes and our extended dataset was tested only on academic benchmarks for this study. Yet, those generative models are not fully understood and may generate biased or even problematic images. They have been released with little information about their training data and process, and we have used them as is. Consequently, a deeper study is required before using any retrieval models of such kind in production. Moreover, we can already foresee undesirable consequences in a number of other scenarios, \eg when prompts of a sensitive nature are used. In general, when using the output of a model in any application involving humans, a rigorous validation needs to be performed.

\myparagraph{Reproducibility Statement}
We have built our approach as much as possible on top of public codebases, data and models. We have provided the training hyperparameters (\Cref{sec:app_implementation_details}) and as much experimental details as possible. We believe that all the information available allows to fully reproduce our experimental validation.
Moreover, we plan to publicly release models that will enable reproducing all results presented in the paper.

\bibliography{bib}

\begin{thebibliography}{134}
\providecommand{\natexlab}[1]{#1}
\providecommand{\url}[1]{\texttt{#1}}
\expandafter\ifx\csname urlstyle\endcsname\relax
  \providecommand{\doi}[1]{doi: #1}\else
  \providecommand{\doi}{doi: \begingroup \urlstyle{rm}\Url}\fi

\bibitem[Aiger et~al.(2023)Aiger, Araujo, and Lynen]{aiger2023yes}
Dror Aiger, Andr{\'e} Araujo, and Simon Lynen.
\newblock {Yes, we CANN: Constrained Approximate Nearest Neighbors for local
  feature-based visual localization}.
\newblock In \emph{Proc. {ICCV}}, 2023.

\bibitem[Anoosheh et~al.(2019)Anoosheh, Sattler, Timofte, Pollefeys, and
  Van~Gool]{AnooshehICRA19NightToDayImageTranslationLocalization}
Asha Anoosheh, Torsten Sattler, Radu Timofte, Marc Pollefeys, and Luc Van~Gool.
\newblock {Night-to-day Image Translation for Retrieval-based Localization}.
\newblock In \emph{Proc. {ICRA}}, 2019.

\bibitem[Arandjelovi\'{c} et~al.(2016)Arandjelovi\'{c}, Gron\'{a}t, Torii,
  Pajdla, and Sivic]{ArandjelovicCVPR16NetVLADPlaceRecognition}
Relja Arandjelovi\'{c}, Petr Gron\'{a}t, Akihiko Torii, Tom\'{a}\v{s} Pajdla,
  and Josef Sivic.
\newblock {{NetVLAD}: {CNN} Architecture for Weakly Supervised Place
  Recognition}.
\newblock In \emph{Proc. {CVPR}}, 2016.

\bibitem[Azizi et~al.(2023)Azizi, Kornblith, Saharia, Norouzi, and
  Fleet]{azizi2023synthetic}
Shekoofeh Azizi, Simon Kornblith, Chitwan Saharia, Mohammad Norouzi, and
  David~J. Fleet.
\newblock {Synthetic Data from Diffusion Models Improves {ImageNet}
  Classification}.
\newblock arXiv:2304.08466, 2023.

\bibitem[Babenko \& Lempitsky(2015)Babenko and
  Lempitsky]{babenko2015aggregating}
Artem Babenko and Victor~S. Lempitsky.
\newblock {Aggregating Deep Convolutional Features for Image Retrieval}.
\newblock In \emph{Proc. {ICCV}}, 2015.

\bibitem[Badino et~al.(2011)Badino, Huber, and Kanade]{badino2011cmu}
Hernan Badino, Daniel Huber, and Takeo Kanade.
\newblock The cmu visual localization data set.
\newblock \emph{Computer Vision Group}, 2011.

\bibitem[Benbihi et~al.(2020)Benbihi, Arravechia, Geist, and
  Pradalier]{BenbihiICRA20ImageBasedPlaceRecBucolicEnv}
Assia Benbihi, St{\'e}phanie Arravechia, Matthieu Geist, and C{\'e}dric
  Pradalier.
\newblock {Image-Based Place Recognition on Bucolic Environment Across Seasons
  From Semantic Edge Description }.
\newblock In \emph{Proc. {ICRA}}, 2020.

\bibitem[Berton et~al.(2022)Berton, Mereu, Trivigno, Masone, Csurka, Sattler,
  and Caputo]{BertonCVPR22DeepVisualGeoLocalizationBenchmark}
Gabriele Berton, Riccardo Mereu, Gabriele Trivigno, Carlo Masone, Gabriela
  Csurka, Torsten Sattler, and Barbara Caputo.
\newblock {Deep Visual Geo-localization Benchmark}.
\newblock In \emph{Proc. {CVPR}}, 2022.

\bibitem[Brachmann \& Rother(2019)Brachmann and
  Rother]{BrachmannICCV19ExpertSampleConsensusReLocalization}
Eric Brachmann and Carsten Rother.
\newblock {Expert Sample Consensus Applied to Camera Re-Localization}.
\newblock In \emph{Proc. {ICCV}}, 2019.

\bibitem[Brachmann \& Rother(2022)Brachmann and
  Rother]{BrachmannPAMI22VisualCameraReLocalizationFromRGBAndRGBDImagesUsingDSAC}
Eric Brachmann and Carsten Rother.
\newblock {Visual Camera Re-Localization from RGB and RGB-D Images Using DSAC}.
\newblock \emph{{IEEE. Trans. PAMI}}, 44\penalty0 (9), 2022.

\bibitem[Brooks et~al.(2023)Brooks, Holynski, and
  Efros]{brooks2022instructpix2pix}
Tim Brooks, Aleksander Holynski, and Alexei~A. Efros.
\newblock {InstructPix2Pix: Learning to Follow Image Editing Instructions}.
\newblock In \emph{Proc. {CVPR}}, 2023.

\bibitem[Cabon et~al.(2020)Cabon, Murray, and
  Humenberger]{CabonX20VirtualKITTI2}
Yohann Cabon, Naila Murray, and Martin Humenberger.
\newblock {Virtual KITTI 2}.
\newblock arXiv:2001.10773, 2020.

\bibitem[Cai et~al.(2017)Cai, Xu, Jia, Qing, and
  Tao]{CaiTIP17DehazeNetAnEnd2EndSystem4SingleHazeRemovaly}
Bolun Cai, Xiangmin Xu, Kui Jia, Chunmei Qing, and Dacheng Tao.
\newblock {DehazeNet: An End-to-End System for Single Image Haze Removal}.
\newblock \emph{{IEEE} Trans. on Image Processing}, 25\penalty0 (11), 2017.

\bibitem[Cao et~al.(2020)Cao, Araujo, and Sim]{cao2020unifying}
Bingyi Cao, Andre Araujo, and Jack Sim.
\newblock Unifying deep local and global features for image search.
\newblock In \emph{Proc. {ECCV}}, 2020.

\bibitem[Cavallari et~al.(2017)Cavallari, Golodetz, Lord, Valentin, Stefano,
  and Torr]{CavallariCVPR17OntheFlyCameraRelocalisation}
Tommaso Cavallari, Stuart Golodetz, {Nicholas A.} Lord, Julien Valentin, {Luigi
  Di} Stefano, and {Philip H.S.} Torr.
\newblock {On-The-Fly Adaptation of Regression Forests for Online Camera
  Relocalisation}.
\newblock In \emph{Proc. {CVPR}}, 2017.

\bibitem[Cheng et~al.(2021)Cheng, Wei, Bao, Chen, Wen, and
  Zhang]{ChengICCV21DualPathLearning4DASS}
Yiting Cheng, Fangyun Wei, Jianmin Bao, Dong Chen, Fang Wen, and Wenqiang
  Zhang.
\newblock {Dual Path Learning for Domain Adaptation of Semantic Segmentation}.
\newblock In \emph{Proc. {ICCV}}, 2021.

\bibitem[Choi et~al.(2019)Choi, Kim, and
  Kim]{ChoiICCV19SelfEnsemblingGANBasedDataAugmentationDASemSegm}
Jaehoon Choi, Taekyung Kim, and Changick Kim.
\newblock {Self-Ensembling with {GAN}-based Data Augmentation for Domain
  Adaptation in Semantic Segmentation}.
\newblock In \emph{Proc. {ICCV}}, 2019.

\bibitem[Csurka et~al.(2017)Csurka, Baradel, Chidlovskii, and
  Clinchant]{CsurkaTASKCV17DiscrepancyBasedNetworksUDA}
Gabriela Csurka, Fabien Baradel, Boris Chidlovskii, and St\'{e}phane Clinchant.
\newblock {Discrepancy-Based Networks for Unsupervised Domain Adaptation: A
  Comparative Study}.
\newblock In \emph{{ICCV Workshops }}, 2017.

\bibitem[Csurka et~al.(2022)Csurka, Volpi, and
  Chidlovskii]{CsurkaFTCGV22SemanticImageSegmentationTwoDecadesOfResearch}
Gabriela Csurka, Riccardo Volpi, and Boris Chidlovskii.
\newblock {Semantic Image Segmentation: Two Decades of Research}.
\newblock \emph{{Foundations and Trends {\textregistered} in Computer Graphics
  and Vision}}, 1--2\penalty0 (14), 2022.

\bibitem[DeTone et~al.(2018)DeTone, Malisiewicz, and
  Rabinovich]{detone2018superpoint}
Daniel DeTone, Tomasz Malisiewicz, and Andrew Rabinovich.
\newblock Superpoint: Self-supervised interest point detection and description.
\newblock In \emph{{CVPR Workshops }}, pp.\  224--236, 2018.

\bibitem[Dunlap et~al.(2023)Dunlap, Umino, Zhang, Yang, Gonzalez, and
  Darrell]{dunlap2023alia}
Lisa Dunlap, Alyssa Umino, Han Zhang, Jiezhi Yang, Joseph Gonzalez, and Trevor
  Darrell.
\newblock {Diversify Your Vision Datasets with Automatic Diffusion-based
  Augmentation}.
\newblock arXiv:2305.16289, 2023.

\bibitem[Fahes et~al.(2023)Fahes, Vu, Bursuc, P{\'e}rez, and
  de~Charette]{FahesICCV23PODAPromptDrivenZeroShotDA}
Mohammad Fahes, Tuan-Hung Vu, Andrei Bursuc, Patrick P{\'e}rez, and Raoul
  de~Charette.
\newblock {P{\O}DA: Prompt-driven Zero-shot Domain Adaptation}.
\newblock In \emph{Proc. {ICCV}}, 2023.

\bibitem[Gaidon et~al.(2016)Gaidon, Wang, Cabon, and
  Vig]{GaidonCVPR16VirtualWorldsAsProxyTracking}
Adrien Gaidon, Qiao Wang, Yohann Cabon, and Eleonora Vig.
\newblock {Virtual Worlds as Proxy for Multi-Object Tracking Analysis}.
\newblock In \emph{Proc. {CVPR}}, 2016.

\bibitem[Garg et~al.(2018)Garg, Suenderhauf, and
  Milford]{GargRSSC18LoSTAppearanceInvariantPlaceRecognVisualSemantics}
Sourav Garg, Niko Suenderhauf, and Michael Milford.
\newblock {LoST? Appearance-Invariant Place Recognition for Opposite Viewpoints
  using Visual Semantics}.
\newblock In \emph{{Robotics: Science and Systems Conference}}, 2018.

\bibitem[Gatys et~al.(2015)Gatys, Ecker, and
  Bethge]{GatysNIPS15TextureSynthesisCNN}
{Leon A.} Gatys, {Alexander S.} Ecker, and Matthias Bethge.
\newblock {Texture Synthesis Using Convolutional Neural Networks}.
\newblock In \emph{Proc. {NeurIPS}}, 2015.

\bibitem[Germain et~al.(2019)Germain, Bourmaud, and
  Lepetit]{Germain3DV19SparseToDenseHypercolumnMatchingVisLoc}
Hugo Germain, Guillaume Bourmaud, and Vincent Lepetit.
\newblock {Sparse-to-Dense Hypercolumn Matching for Long-Term Visual
  Localization}.
\newblock In \emph{{Proc. 3DV}}, 2019.

\bibitem[Halder et~al.(2019)Halder, Lalonde, and {de
  Charette}]{HalderICCV19PhysicsBasedRendering4ImprovingRobustness2Rain}
Shirsendu~Sukanta Halder, Jean-François Lalonde, and Raoul {de Charette}.
\newblock {Physics-based Rendering for Improving Robustness to Rain}.
\newblock In \emph{Proc. {ICCV}}, 2019.

\bibitem[Hausler et~al.(2021)Hausler, Garg, Xu, Milford, and
  Fischer]{hausler2021patch}
Stephen Hausler, Sourav Garg, Ming Xu, Michael Milford, and Tobias Fischer.
\newblock Patch-netvlad: Multi-scale fusion of locally-global descriptors for
  place recognition.
\newblock In \emph{Proc. {CVPR}}, 2021.

\bibitem[He et~al.(2016)He, Zhang, Ren, and Sun]{he2016resnet}
Kaiming He, Xiangyu Zhang, Shaoqing Ren, and Jian Sun.
\newblock {Deep Residual Learning for Image Recognition}.
\newblock In \emph{Proc. {CVPR}}, 2016.

\bibitem[He et~al.(2023)He, Sun, Yu, Xue, Zhang, Torr, Bai, and
  Qi]{he2023synthetic}
Ruifei He, Shuyang Sun, Xin Yu, Chuhui Xue, Wenqing Zhang, Philip Torr, Song
  Bai, and Xiaojuan Qi.
\newblock {Is Synthetic Data from Generative Models Ready for Image
  Recognition?}
\newblock In \emph{Proc. {ICLR}}, 2023.

\bibitem[Hendrycks et~al.(2020)Hendrycks, Mu, Cubuk, Zoph, Gilmer, and
  Lakshminarayanan]{hendrycks2019augmix}
Dan Hendrycks, Norman Mu, Ekin~D. Cubuk, Barret Zoph, Justin Gilmer, and Balaji
  Lakshminarayanan.
\newblock {AugMix: A Simple Data Processing Method to Improve Robustness and
  Uncertainty}.
\newblock \emph{ICLR}, 2020.

\bibitem[Hoffman et~al.(2018)Hoffman, Tzeng, Park, Zhu, Isola, Saenko, Efros,
  and Darrel]{HoffmanICML18CyCADACycleConsistentAdversarialDA}
Judy Hoffman, Eric Tzeng, Taesung Park, Jun-Yan Zhu, Phillip Isola, Kate
  Saenko, Alexei~A. Efros, and Trevor Darrel.
\newblock {CyCADA: Cycle-Consistent Adversarial Domain Adaptation}.
\newblock In \emph{Proc. {ICML}}, 2018.

\bibitem[Hu et~al.(2021)Hu, Qiao, Cheng, Liu, and
  Wang]{HuTIP21DASGILDA4SemanticAndGeometricAwareImageBasedLocalization}
Hanjiang Hu, Zhijian Qiao, Ming Cheng, Zhe Liu, and Hesheng Wang.
\newblock {DASGIL: Domain Adaptation for Semantic and Geometric-aware
  Image-based Localization}.
\newblock \emph{{IEEE} Trans. on Image Processing}, 30, 2021.

\bibitem[Hu et~al.(2019)Hu, Fu, Zhu, and
  Heng]{HuCVPR19DepthAttentionalFeaturesSingleImageRainRemoval}
Xiaowei Hu, Chi-Wing Fu, Lei Zhu, and Pheng-Ann Heng.
\newblock {Depth-Attentional Features for Single-Image Rain Removal}.
\newblock In \emph{Proc. {CVPR}}, 2019.

\bibitem[Huang \& Belongie(2017)Huang and
  Belongie]{HuangICCV17ArbitraryStyleTransfer}
Sun Huang and Serge Belongie.
\newblock {Arbitrary Style Transfer in Real-time with Adaptive Instance
  Normalization}.
\newblock In \emph{Proc. {ICCV}}, 2017.

\bibitem[Humenberger et~al.(2020)Humenberger, Cabon, Guerin, Morat, Revaud,
  Rerole, Pion, de~Souza, Leroy, and
  Csurka]{HumenbergerX20RobustImageRetrievalBasedVisLocKapture}
Martin Humenberger, Yohann Cabon, Nicolas Guerin, Julien Morat, J\'{e}r\^{o}me
  Revaud, Philippe Rerole, No\'{e} Pion, Cesar de~Souza, Vincent Leroy, and
  Gabriela Csurka.
\newblock {Robust Image Retrieval-based Visual Localization using Kapture}.
\newblock arXiv:2007.13867, 2020.

\bibitem[Humenberger et~al.(2022)Humenberger, Cabon, Pion, Weinzaepfel, Lee,
  Gu{\'e}rin, Sattler, and Csurka]{humenberger2022investigating}
Martin Humenberger, Yohann Cabon, No{\'e} Pion, Philippe Weinzaepfel, Donghwan
  Lee, Nicolas Gu{\'e}rin, Torsten Sattler, and Gabriela Csurka.
\newblock {Investigating the Role of Image Retrieval for Visual Localization:
  An Exhaustive Benchmark}.
\newblock \emph{{IJCV}}, 130\penalty0 (7), 2022.

\bibitem[Irschara et~al.(2009)Irschara, Zach, Frahm, and
  Bischof]{IrscharaCVPR09FromSFMLocationRecognition}
Arnold Irschara, Christopher Zach, Jan-Michael Frahm, and Horst Bischof.
\newblock {From Structure-from-Motion Point Clouds to Fast Location
  Recognition}.
\newblock In \emph{Proc. {CVPR}}, 2009.

\bibitem[Jenicek \& Chum(2019)Jenicek and Chum]{JenicekICCV19NoFearOfTheDark}
Tomas Jenicek and Ond\v{r}ej Chum.
\newblock {No Fear of the Dark: Image Retrieval under Varying Illumination
  Conditions}.
\newblock In \emph{Proc. {ICCV}}, 2019.

\bibitem[Johnson et~al.(2019)Johnson, Douze, and J{\'e}gou]{johnson2019billion}
Jeff Johnson, Matthijs Douze, and Herv{\'e} J{\'e}gou.
\newblock Billion-scale similarity search with {GPUs}.
\newblock \emph{IEEE Transactions on Big Data}, 7\penalty0 (3), 2019.

\bibitem[Kahraman \& {de Charette}(2017)Kahraman and {de
  Charette}]{KahramanTR16InfluenceOfFogOnComputerVisionAlgorithms}
Sule Kahraman and Raoul {de Charette}.
\newblock {Influence of Fog on Computer Vision Algorithms}.
\newblock Technical Report hal-01620602, {INRIA, Paris}, 2017.

\bibitem[Kendall et~al.(2015)Kendall, Grimes, and
  Cipolla]{KendallICCV15PoseNetCameraRelocalization}
Alex Kendall, Matthew Grimes, and Roberto Cipolla.
\newblock {PoseNet: a Convolutional Network for Real-Time {6-DOF} Camera
  Relocalization}.
\newblock In \emph{Proc. {ICCV}}, 2015.

\bibitem[Kim et~al.(2017)Kim, Dunn, and
  Frahm]{KimCVPR17LearnedContextualFeatureReweightingGeoLoc}
{Hyo Jin} Kim, Enrique Dunn, and Jan-Michael Frahm.
\newblock {Learned Contextual Feature Reweighting for Image Geo-Localization}.
\newblock In \emph{Proc. {CVPR}}, 2017.

\bibitem[Kobyshev et~al.(2014)Kobyshev, Riemenschneider, and {Van
  Gool}]{Kobyshev3DV14MatchingFeaturesCorrectlyThroughSemanticUnderstanding}
Nikolay Kobyshev, Hayko Riemenschneider, and Luc {Van Gool}.
\newblock {Matching Features Correctly through Semantic Understanding}.
\newblock In \emph{{Proc. 3DV}}, 2014.

\bibitem[Larsson et~al.(2019)Larsson, Stenborg, Toft, Hammarstrand, Sattler,
  and Kahl]{LarssonICCV19FineGrainedSISSelfSupISVL}
Mans Larsson, Erik Stenborg, Carl Toft, Lars Hammarstrand, Torsten Sattler, and
  Fredrik Kahl.
\newblock {Fine-Grained Segmentation Networks: Self-Supervised Segmentation for
  Improved Long-Term Visual Localization}.
\newblock In \emph{Proc. {ICCV}}, 2019.

\bibitem[Lee et~al.(2021)Lee, Ryu, Yeon, Lee, Kim, Han, Cabon, Weinzaepfel,
  Gu{\'e}rin, Csurka, et~al.]{lee2021large}
Donghwan Lee, Soohyun Ryu, Suyong Yeon, Yonghan Lee, Deokhwa Kim, Cheolho Han,
  Yohann Cabon, Philippe Weinzaepfel, Nicolas Gu{\'e}rin, Gabriela Csurka,
  et~al.
\newblock Large-scale localization datasets in crowded indoor spaces.
\newblock In \emph{Proc. {CVPR}}, 2021.

\bibitem[Lengyel et~al.(2021)Lengyel, Garg, Milford, and {van
  Gemert}]{LengyelICCV21ZeroShotDayNightDAWithPhysicsPrior}
Attila Lengyel, Sourav Garg, Michael Milford, and Jan~C. {van Gemert}.
\newblock {Zero-Shot Day-Night Domain Adaptation with a Physics Prior}.
\newblock In \emph{Proc. {ICCV}}, 2021.

\bibitem[Leyva-Vallina et~al.(2023)Leyva-Vallina, Strisciuglio, and
  Petkov]{leyva2023data}
Mar{\'\i}a Leyva-Vallina, Nicola Strisciuglio, and Nicolai Petkov.
\newblock {Data-Efficient Large Scale Place Recognition With Graded Similarity
  Supervision}.
\newblock In \emph{Proc. {CVPR}}, 2023.

\bibitem[Li et~al.(2017)Li, Peng, Wang, Xu, and
  Feng]{LiICCV17AODNetAllInOneDehazingNetwork}
Boyi Li, Xiulian Peng, Zhangyang Wang, Jizheng Xu, and Dan Feng.
\newblock {AOD-Net: All-in-One Dehazing Network}.
\newblock In \emph{Proc. {ICCV}}, 2017.

\bibitem[Li et~al.(2018{\natexlab{a}})Li, Pan, Li, and
  Tan]{LiCVPR18SingleImageDehazingViaConditionalGenerativeAdversarialNetwork}
Runde Li, Jinshan Pan, Zechao Li, and Jinhui Tan.
\newblock {Single Image Dehazing via Conditional Generative Adversarial
  Network}.
\newblock In \emph{Proc. {CVPR}}, 2018{\natexlab{a}}.

\bibitem[Li et~al.(2018{\natexlab{b}})Li, Liu, Li, Yang, and
  Kautz]{LiECCV18ClosedFormImageStylization}
Yijun Li, Ming-Yu Liu, Xueting Li, Ming-Hsuan Yang, and Jan Kautz.
\newblock {A Closed-form Solution to Photorealistic Image Stylization}.
\newblock In \emph{Proc. {ECCV}}, 2018{\natexlab{b}}.

\bibitem[Li et~al.(2010)Li, Snavely, and
  Huttenlocher]{LiECCV10LocationRecPriorFeatureMatching}
Yunpeng Li, Noah Snavely, and Dan Huttenlocher.
\newblock {Location Recognition Using Prioritized Feature Matching}.
\newblock In \emph{Proc. {ECCV}}, 2010.

\bibitem[Li et~al.(2019)Li, Yuan, and
  Vasconcelos]{LiCVPR19BidirectionalLearningDASemSegm}
Yunsheng Li, Lu~Yuan, and Nuno Vasconcelos.
\newblock {Bidirectional Learning for Domain Adaptation of Semantic
  Segmentation}.
\newblock In \emph{Proc. {CVPR}}, 2019.

\bibitem[Lindenberger et~al.(2023)Lindenberger, Sarlin, and
  Pollefeys]{lindenberger2023lightglue}
Philipp Lindenberger, Paul-Edouard Sarlin, and Marc Pollefeys.
\newblock {LightGlue: Local Feature Matching at Light Speed}.
\newblock In \emph{Proc. {ICCV}}, 2023.

\bibitem[Liu et~al.(2019)Liu, Li, and
  Dai]{LiuICCV19StochasticAttractionRepulsionEmbedding}
Liu Liu, Hongdong Li, and Yuchao Dai.
\newblock {Stochastic Attraction-Repulsion Embedding for Large Scale Image
  Localization}.
\newblock In \emph{Proc. {ICCV}}, 2019.

\bibitem[Liu et~al.(2017)Liu, Breuel, and
  Kautz]{LiuNIPS17UnsupervisedI2ITranslationNetworks}
Ming-Yu Liu, Thomas Breuel, and Jan Kautz.
\newblock {Unsupervised Image-to-Image Translation Networks}.
\newblock In \emph{Proc. {NeurIPS}}, 2017.

\bibitem[Loshchilov \& Hutter(2019)Loshchilov and
  Hutter]{loshchilov2017decoupled}
Ilya Loshchilov and Frank Hutter.
\newblock {Decoupled Weight Decay Regularization}.
\newblock In \emph{Proc. {ICLR}}, 2019.

\bibitem[Lowry et~al.(2016)Lowry, S\"{u}nderhauf, Newman, Leonard, Cox, Corke,
  and Milford]{LowryTROB16VisualPlaceRecognitionSurvey}
Stephanie Lowry, Niko S\"{u}nderhauf, Paul Newman, {John J.} Leonard, David
  Cox, Peter Corke, and {Michael J.} Milford.
\newblock {Visual Place Recognition: A Survey}.
\newblock \emph{{IEEE Transactions on Robotics}}, 32\penalty0 (1), 2016.

\bibitem[Maddern et~al.(2017)Maddern, Pascoe, Linegar, and
  Newman]{maddern20171}
Will Maddern, Geoff Pascoe, Chris Linegar, and Paul Newman.
\newblock {1 Year, 1000 Km: The Oxford RobotCar Dataset}.
\newblock \emph{Intl. J. of {Robotics Research}}, 36\penalty0 (1), 2017.

\bibitem[Mohwald et~al.(2023)Mohwald, Jenicek, and Chum]{mohwald2023dark}
Albert Mohwald, Tomas Jenicek, and Ond{\v{r}}ej Chum.
\newblock Dark side augmentation: Generating diverse night examples for metric
  learning.
\newblock In \emph{Proc. {ICCV}}, 2023.

\bibitem[Mokady et~al.(2023)Mokady, Hertz, Aberman, Pritch, and
  Cohen-Or]{BarronCVPR23NULLTextInversion4EditingRealImagesUsingGuidedDiffusionModels}
Ron Mokady, Amir Hertz, Kfir Aberman, Yael Pritch, and Daniel Cohen-Or.
\newblock {NULL-Text Inversion for Editing Real Images Using Guided Diffusion
  Models}.
\newblock In \emph{Proc. {CVPR}}, 2023.

\bibitem[Nichol et~al.(2022)Nichol, Dhariwal, Ramesh, Shyam, Mishkin, McGrew,
  Sutskever, and
  Chen]{NicholICML22GLIDETowardsPhotorealisticImageGenerationEditingDiffusionModels}
Alex Nichol, Prafulla Dhariwal, Aditya Ramesh, Pranav Shyam, Pamela Mishkin,
  Bob McGrew, Ilya Sutskever, and Mark Chen.
\newblock {GLIDE: Towards Photorealistic Image Generation and Editing with
  Text-Guided Diffusion Models}.
\newblock In \emph{Proc. {ICML}}, 2022.

\bibitem[Noh et~al.(2017)Noh, Araujo, Sim, Weyand, and Han]{noh2017large}
Hyeonwoo Noh, Andre Araujo, Jack Sim, Tobias Weyand, and Bohyung Han.
\newblock Large-scale image retrieval with attentive deep local features.
\newblock In \emph{Proc. {ICCV}}, 2017.

\bibitem[Paolicelli et~al.(2022)Paolicelli, Tavera, Berton, Masone, and
  Caputo]{PaolicelliX22LearningSemantics4VisualPlaceRecognitionThroughMultiScaleAttention}
Valerio Paolicelli, Antonio Tavera, Gabriele Berton, Carlo Masone, and Barbara
  Caputo.
\newblock {Learning Semantics for Visual Place Recognition through Multi-Scale
  Attention}.
\newblock arXiv:2201.09701, 2022.

\bibitem[Parmar et~al.(2023)Parmar, Singh, Zhang, Li, Lu, and
  Zhu]{ParmarSIGGRAPH23ZeroShotImage2ImageTranslation}
Gaurav Parmar, Krishna~Kumar Singh, Richard Zhang, Yijun Li, Jingwan Lu, and
  Jun-Yan Zhu.
\newblock {Zero-shot Image-to-Image Translation}.
\newblock In \emph{Proc. {SIGGRAPH}}, 2023.

\bibitem[Paszke et~al.(2019)Paszke, Gross, Massa, Lerer, Bradbury, Chanan,
  Killeen, Lin, Gimelshein, Antiga, et~al.]{paszke2019pytorch}
Adam Paszke, Sam Gross, Francisco Massa, Adam Lerer, James Bradbury, Gregory
  Chanan, Trevor Killeen, Zeming Lin, Natalia Gimelshein, Luca Antiga, et~al.
\newblock Pytorch: An imperative style, high-performance deep learning library.
\newblock \emph{Proc. {NeurIPS}}, 2019.

\bibitem[Porav et~al.(2019)Porav, Bruls, and
  Newman]{PoravITSC19DontWorryAboutWeather}
Horia Porav, Tom Bruls, and Paul Newman.
\newblock {Don't Worry About the Weather: Unsupervised Condition-Dependent
  Domain Adaptation}.
\newblock In \emph{Proc. {ITSC}}, 2019.

\bibitem[Qian et~al.(2018)Qian, Tan, Yang, Su, and
  Liu]{QianCVPR18AttentiveGAN4RaindropRemoval}
Rui Qian, Robby~T. Tan, Wenhan Yang, Jiajun Su, and Jiaying Liu.
\newblock {Attentive Generative Adversarial Network for Raindrop Removal from a
  Single Images}.
\newblock In \emph{Proc. {CVPR}}, 2018.

\bibitem[Qin et~al.(2023)Qin, Zhang, Yu, Feng, Yang, Zhou, Wang, Niebles,
  Xiong, Savarese, Ermon, Fu, and
  Xu]{QinX23UniControlAUnifiedDiffusionModel4ControllableVisualGeneratioInTheWild}
Can Qin, Shu Zhang, Ning Yu, Yihao Feng, Xinyi Yang, Yingbo Zhou, Huan Wang,
  Juan~Carlos Niebles, Caiming Xiong, Silvio Savarese, Stefano Ermon, Yun Fu,
  and Ran Xu.
\newblock {UniControl: A Unified Diffusion Model for Controllable Visual
  Generation In the Wild}.
\newblock arXiv:2305.11147, 2023.

\bibitem[Radenovi{\'c} et~al.(2019)Radenovi{\'c}, Tolias, and
  Chum]{radenovic2018fine}
Filip Radenovi{\'c}, Giorgos Tolias, and Ond\v{r}ej Chum.
\newblock {Fine-Tuning {CNN} Image Retrieval with no Human Annotation}.
\newblock \emph{{IEEE. Trans. PAMI}}, 41\penalty0 (7), 2019.

\bibitem[Ramesh et~al.(2021)Ramesh, Pavlov, Goh, Gray, Voss, Radford, Chen, and
  Sutskever]{ramesh2021zero}
Aditya Ramesh, Mikhail Pavlov, Gabriel Goh, Scott Gray, Chelsea Voss, Alec
  Radford, Mark Chen, and Ilya Sutskever.
\newblock Zero-shot text-to-image generation.
\newblock In \emph{Proc. {ICML}}, 2021.

\bibitem[Ren et~al.(2017)Ren, Tian, Han, Chan, and
  Tang]{RenCVPR17VideoDesnowingAndDerainingBasedOnMatrixDecomposition}
Weihong Ren, Jiandong Tian, Zhi Han, Antoni Chan, and Yandong Tang.
\newblock {Video Desnowing and Deraining Based on Matrix Decomposition}.
\newblock In \emph{Proc. {CVPR}}, 2017.

\bibitem[Ren et~al.(2018)Ren, Ma, Zhang, Pan, Cao, Liu, and
  Yang]{RenCVPR18GatedFusionNetwork4SingleImageDehazing}
Wenqi Ren, Lin Ma, Jiawei Zhang, Jinshan Pan, Xiaochun Cao, Wei Liu, and
  Ming-Hsuan Yang.
\newblock {Gated Fusion Network for Single Image Dehazing}.
\newblock In \emph{Proc. {CVPR}}, 2018.

\bibitem[Revaud et~al.(2019{\natexlab{a}})Revaud, Almazan, {de Rezende}, and
  {de Souza}]{revaud2019learning}
Jerome Revaud, Jon Almazan, Rafael~Sampaio {de Rezende}, and C\'{e}sar~Roberto
  {de Souza}.
\newblock {Learning with Average Precision: Training Image Retrieval with a
  Listwise Loss}.
\newblock In \emph{Proc. {ICCV}}, 2019{\natexlab{a}}.

\bibitem[Revaud et~al.(2019{\natexlab{b}})Revaud, De~Souza, Humenberger, and
  Weinzaepfel]{revaud2019r2d2}
Jerome Revaud, Cesar De~Souza, Martin Humenberger, and Philippe Weinzaepfel.
\newblock R2d2: Reliable and repeatable detector and descriptor.
\newblock In \emph{Proc. {NeurIPS}}, 2019{\natexlab{b}}.

\bibitem[Richter et~al.(2016)Richter, Vineet, Roth, and
  Vladlen]{RichterECCV16PlayingForData}
Stephan~R. Richter, Vibhav Vineet, Stefan Roth, and Koltun Vladlen.
\newblock {Playing for Data: Ground Truth from Computer Games}.
\newblock In \emph{Proc. {ECCV}}, 2016.

\bibitem[Rombach et~al.(2022)Rombach, Blattmann, Lorenz, Esser, and
  Ommer]{rombach2022high}
Robin Rombach, Andreas Blattmann, Dominik Lorenz, Patrick Esser, and Bj{\"o}rn
  Ommer.
\newblock {High-Resolution Image Synthesis with Latent Diffusion Models}.
\newblock In \emph{Proc. {CVPR}}, 2022.

\bibitem[Romera et~al.(2019)Romera, Bergasa1, Yang, Alvarez, and
  Barea]{RomeraIVS19BridgingDayNightDomainGapsSIS}
Eduardo Romera, Luis~M. Bergasa1, Kailun Yang, Jose~M. Alvarez, and Rafael
  Barea.
\newblock {Bridging the Day and Night Domain Gap for Semantic Segmentation}.
\newblock In \emph{{IEEE Intelligent Vehicles Symposium}}, 2019.

\bibitem[Ros et~al.(2016)Ros, Sellart, Materzy\'{n}ska, V\'{a}zquez, and
  L\'{o}pez]{RosCVPR16SYNTHIADataset}
German Ros, Laura Sellart, Joanna Materzy\'{n}ska, David V\'{a}zquez, and
  Antonio~M. L\'{o}pez.
\newblock {The {SYNTHIA} Dataset: a Large Collection of Synthetic Images for
  Semantic Segmentation of Urban Scenes}.
\newblock In \emph{Proc. {CVPR}}, 2016.

\bibitem[Russakovsky et~al.(2015)Russakovsky, Deng, Su, Krause, Satheesh, Ma,
  Huang, Karpathy, Khosla, Bernstein, Berg, and Fei-Fei]{ilsvrc}
Olga Russakovsky, Jia Deng, Hao Su, Jonathan Krause, Sanjeev Satheesh, Sean Ma,
  Zhiheng Huang, Andrej Karpathy, Aditya Khosla, Michael Bernstein,
  Alexander~C. Berg, and Li~Fei-Fei.
\newblock {ImageNet Large Scale Visual Recognition Challenge}.
\newblock \emph{{IJCV}}, 115\penalty0 (3), 2015.

\bibitem[Saharia et~al.(2022)Saharia, Chan, Saxena, Li, Whang, Denton,
  Ghasemipour, Ayan, Mahdavi, Lopes, Salimans, Ho, Fleet, and
  Norouz]{SahariaX22PhotorealisticText2ImageDiffusionModelsWithDeepLanguageUnderstanding}
Chitwan Saharia, William Chan, Saurabh Saxena, Lala Li, Jay Whang, Emily
  Denton, Seyed K.~S. Ghasemipour, Burcu~Karagol Ayan, S.~Sara Mahdavi,
  Rapha~Gontijo Lopes, Tim Salimans, Jonathan Ho, David~J. Fleet, and Mohammad
  Norouz.
\newblock {Photorealistic Text-to-Image Diffusion Models with Deep Language
  Understanding}.
\newblock arXiv:2205.11487, 2022.

\bibitem[Sakaridis et~al.(2018)Sakaridis, Dai, and
  Van~Gool]{SakaridisIJCV17SemanticFoggySceneUnderstandingWithSyntheticData}
Christos Sakaridis, Dengxin Dai, and Luc Van~Gool.
\newblock {Semantic Foggy Scene Understanding with Synthetic Data}.
\newblock \emph{{IJCV}}, 126, 2018.

\bibitem[Sakaridis et~al.(2019)Sakaridis, Dai, and {Van
  Gool}]{SakaridisICCV19GuidedCurriculumModelAdaptationSemSegm}
Christos Sakaridis, Dengxin Dai, and Luc {Van Gool}.
\newblock {Guided Curriculum Model Adaptation and Uncertainty-Aware Evaluation
  for Semantic Nighttime Image Segmentation}.
\newblock In \emph{Proc. {ICCV}}, 2019.

\bibitem[Sar{\i}y{\i}ld{\i}z et~al.(2023)Sar{\i}y{\i}ld{\i}z, Alahari, Larlus,
  and Kalantidis]{sariyildiz2023fake}
Mert~B{\"u}lent Sar{\i}y{\i}ld{\i}z, Karteek Alahari, Diane Larlus, and Yannis
  Kalantidis.
\newblock {Fake it till you make it: Learning transferable representations from
  synthetic {ImageNet} clones}.
\newblock In \emph{Proc. {CVPR}}, 2023.

\bibitem[Sariyildiz et~al.(2023)Sariyildiz, Kalantidis, Alahari, and
  Larlus]{sariyildiz2023improving}
Mert~Bulent Sariyildiz, Yannis Kalantidis, Karteek Alahari, and Diane Larlus.
\newblock {No Reason for No Supervision: Improved Generalization in Supervised
  Models}.
\newblock In \emph{Proc. {ICLR}}, 2023.

\bibitem[Sarlin et~al.(2019)Sarlin, Cadena, Siegwart, and
  Dymczyk]{SarlinCVPR19FromCoarsetoFineHierarchicalLocalization}
Paul-Edouard Sarlin, Cesar Cadena, Roland Siegwart, and Marcin Dymczyk.
\newblock {From Coarse to Fine: Robust Hierarchical Localization at Large
  Scale}.
\newblock In \emph{Proc. {CVPR}}, 2019.

\bibitem[Sarlin et~al.(2020)Sarlin, DeTone, Malisiewicz, and
  Rabinovich]{sarlin20superglue}
Paul-Edouard Sarlin, Daniel DeTone, Tomasz Malisiewicz, and Andrew Rabinovich.
\newblock {SuperGlue}: Learning feature matching with graph neural networks.
\newblock In \emph{Proc. {CVPR}}, 2020.

\bibitem[Sattler et~al.(2012)Sattler, Weyand, Leibe, and
  Kobbelt]{sattler2012image}
Torsten Sattler, Tobias Weyand, Bastian Leibe, and Leif Kobbelt.
\newblock Image retrieval for image-based localization revisited.
\newblock In \emph{Proc. {BMVC}}, 2012.

\bibitem[Sattler et~al.(2017)Sattler, Leibe, and
  Kobbelt]{SattlerPAMI17EfficientPrioritizedMatching}
Torsten Sattler, Bastian Leibe, and Leif Kobbelt.
\newblock {Efficient \& Effective Prioritized Matching for Large-Scale
  Image-Based Localization}.
\newblock \emph{{IEEE. Trans. PAMI}}, 39\penalty0 (9), 2017.

\bibitem[Sattler et~al.(2018)Sattler, Maddern, Toft, Torii, Hammarstrand,
  Stenborg, Safari, Okutomi, Pollefeys, Sivic, et~al.]{sattler2018benchmarking}
Torsten Sattler, Will Maddern, Carl Toft, Akihiko Torii, Lars Hammarstrand,
  Erik Stenborg, Daniel Safari, Masatoshi Okutomi, Marc Pollefeys, Josef Sivic,
  et~al.
\newblock Benchmarking 6dof outdoor visual localization in changing conditions.
\newblock In \emph{Proc. {CVPR}}, 2018.

\bibitem[Sattler et~al.(2019)Sattler, Zhou, Pollefeys, and
  Leal-Taix\'{e}]{SattlerCVPR19UnderstandingLimitationsPoseRegression}
Torsten Sattler, Qunjie Zhou, Marc Pollefeys, and Laura Leal-Taix\'{e}.
\newblock {Understanding the Limitations of CNN-based Absolute Camera Pose
  Regression}.
\newblock In \emph{Proc. {CVPR}}, 2019.

\bibitem[Sch\"{o}nberger et~al.(2018)Sch\"{o}nberger, Pollefeys, Geiger, and
  Sattler]{SchonbergerCVPR18SemanticVisualLocalization}
Johannes~L. Sch\"{o}nberger, Marc Pollefeys, Andreas Geiger, and Torsten
  Sattler.
\newblock {Semantic Visual Localization}.
\newblock In \emph{Proc. {CVPR}}, 2018.

\bibitem[Sch\"{o}nberger \& Frahm(2016)Sch\"{o}nberger and
  Frahm]{schoenberger2016sfm}
Johannes~Lutz Sch\"{o}nberger and Jan-Michael Frahm.
\newblock {Structure-from-Motion Revisited}.
\newblock In \emph{Proc. {CVPR}}, 2016.

\bibitem[Se et~al.(2002)Se, Lowe, and
  Little]{SeIROS02GlobalLocalizationDistinctiveVisualFeatures}
Stephen Se, {David G.} Lowe, and {J. J.} Little.
\newblock {Global Localization using Distinctive Visual Features}.
\newblock In \emph{Proc. {IROS}}, 2002.

\bibitem[Shi et~al.(2019)Shi, Shen, Gao, and
  Zhu]{ShiICIP19VisLocSparseSemantic3DMap}
Tianxin Shi, Shuhan Shen, Xiang Gao, and Lingjie Zhu.
\newblock {Visual Localization using Sparse Semantic 3D Map}.
\newblock In \emph{Proc. {ICIP}}, 2019.

\bibitem[Shotton et~al.(2013)Shotton, Glocker, Zach, Izadi, Criminisi, and
  Fitzgibbon]{ShottonCVPR13SceneCoordinateRegression}
Jamie Shotton, Ben Glocker, Christopher Zach, Shahram Izadi, Antonio Criminisi,
  and Andrew Fitzgibbon.
\newblock {Scene Coordinate Regression Forests for Camera Relocalization in
  {RGB-D} Images}.
\newblock In \emph{Proc. {CVPR}}, 2013.

\bibitem[Simonyan \& Zisserman(2015)Simonyan and Zisserman]{simonyan2014very}
Karen Simonyan and Andrew Zisserman.
\newblock Very deep convolutional networks for large-scale image recognition.
\newblock \emph{Proc. {ICLR}}, 2015.

\bibitem[Sun et~al.(2017)Sun, Xie, Luo, and Wang]{sun2017dataset}
Xun Sun, Yuanfan Xie, Pei Luo, and Liang Wang.
\newblock A dataset for benchmarking image-based localization.
\newblock In \emph{Proc. {CVPR}}, 2017.

\bibitem[Taira et~al.(2021)Taira, Okutomi, Sattler, Cimpoi, Pollefeys, Sivic,
  Pajdla, and Akihiko]{TairaPAMI21InLocIndoorVisualLocalization}
Hajime Taira, Masatoshi Okutomi, Torsten Sattler, Mircea Cimpoi, Marc
  Pollefeys, Josef Sivic, Tom\'{a}\v{s} Pajdla, and Torii Akihiko.
\newblock {InLoc: Indoor Visual Localization with Dense Matching and View
  Synthesis}.
\newblock \emph{{IEEE. Trans. PAMI}}, 43\penalty0 (4), 2021.

\bibitem[Thomas \& Kovashka(2019)Thomas and
  Kovashka]{ThomasACCV19ArtisticObjectRecognitionUnsupervisedStyleAdaptation}
Christopher Thomas and Adriana Kovashka.
\newblock {Artistic Object Recognition by Unsupervised Style Adaptation}.
\newblock In \emph{Proc. {ACCV}}, 2019.

\bibitem[Toft et~al.(2017)Toft, Olsson, and
  Kahl]{ToftICCVWS17LongTerm3DLocalizationPoseFromSemanticLabellings}
Carl Toft, Carl Olsson, and Fredrik Kahl.
\newblock {Long-term 3D Localization and Pose from Semantic Labellings}.
\newblock In \emph{{ICCV Workshops }}, 2017.

\bibitem[Toft et~al.(2018)Toft, Stenborg, Hammarstrand, Brynte, Pollefeys,
  Sattler, and Kahl]{ToftECCV18SemanticMatchConsistencyLTVL}
Carl Toft, Erik Stenborg, Lars Hammarstrand, Lucas Brynte, Marc Pollefeys,
  Torsten Sattler, and Fredrik Kahl.
\newblock {Semantic Match Consistency for Long-term Visual Localization}.
\newblock In \emph{Proc. {ECCV}}, 2018.

\bibitem[Toft et~al.(2022)Toft, Maddern, Torii, Hammarstrand, Stenborg, Safari,
  Okutomi, Pollefeys, Sivic, Pajdla, Kahl, and
  Sattler]{ToftPAMI22LongTermVisualLocalizationRevisited}
Carl Toft, Will Maddern, Akihiko Torii, Lars Hammarstrand, Erik Stenborg,
  Daniel Safari, Masatoshi Okutomi, Marc Pollefeys, Josef Sivic, Tomas Pajdla,
  Fredrik Kahl, and Torsten Sattler.
\newblock {Long-Term Visual Localization Revisited}.
\newblock \emph{{IEEE. Trans. PAMI}}, 44\penalty0 (4), 2022.

\bibitem[Toldo et~al.(2020)Toldo, Michieli, Agresti, and
  Zanuttigh]{ToldoIVC20UDAMobileSemSegmCycleConsistencyFeatureAlignment}
Marco Toldo, Umberto Michieli, Gianluca Agresti, and Pietro Zanuttigh.
\newblock {Unsupervised Domain Adaptation for Mobile Semantic Segmentation
  based on Cycle Consistency and Feature Alignment}.
\newblock \emph{Image and Vision Computing}, 95\penalty0 (103889), 2020.

\bibitem[Tolias et~al.(2013)Tolias, Avrithis, and
  J{\'e}gou]{tolias2013aggregate}
Giorgos Tolias, Yannis Avrithis, and Herv{\'e} J{\'e}gou.
\newblock {To Aggregate or not to Aggregate: Selective Match Kernels for Image
  Search}.
\newblock In \emph{Proc. {ICCV}}, 2013.

\bibitem[Tolias et~al.(2020)Tolias, Jenicek, and Chum]{tolias2020learning}
Giorgos Tolias, Tomas Jenicek, and Ond{\v{r}}ej Chum.
\newblock {Learning and Aggregating Deep Local Descriptors for Instance-level
  Recognition}.
\newblock In \emph{Proc. {ECCV}}, 2020.

\bibitem[Torii et~al.(2015{\natexlab{a}})Torii, Arandjelovi\'c, Sivic, Okutomi,
  and Pajdla]{Torii-CVPR2015}
A.~Torii, R.~Arandjelovi\'c, J.~Sivic, M.~Okutomi, and T.~Pajdla.
\newblock 24/7 place recognition by view synthesis.
\newblock In \emph{Proc. {CVPR}}, 2015{\natexlab{a}}.

\bibitem[Torii et~al.(2013)Torii, Sivic, Pajdla, and Okutomi]{torii2013visual}
Akihiko Torii, Josef Sivic, Tomas Pajdla, and Masatoshi Okutomi.
\newblock Visual place recognition with repetitive structures.
\newblock In \emph{Proc. {CVPR}}, 2013.

\bibitem[Torii et~al.(2015{\natexlab{b}})Torii, Sivic, Okutomi, and
  Pajdla]{ToriiPAMI15VisualPlaceRecognRepetitiveStructures}
Akihiko Torii, Josef Sivic, Masatoshi Okutomi, and Tomas Pajdla.
\newblock {Visual Place Recognition with Repetitive Structures}.
\newblock \emph{{IEEE. Trans. PAMI}}, 37\penalty0 (11), 2015{\natexlab{b}}.

\bibitem[Torii et~al.(2018)Torii, Arandjelovi\'{c}, Sivic, Okutomi, and
  Pajdla]{ToriiPAMI18247PlaceRecognitionViewSynthesis}
Akihiko Torii, Relja Arandjelovi\'{c}, Josef Sivic, Masatoshi Okutomi, and
  Tom\'{a}\v{s} Pajdla.
\newblock {24/7 Place Recognition by View Synthesis}.
\newblock \emph{{IEEE. Trans. PAMI}}, 40\penalty0 (2), 2018.

\bibitem[Trabucco et~al.(2023)Trabucco, Doherty, Gurinas, and
  Salakhutdinov]{TrabuccoX23EffectiveDataAugmentationWithDiffusionModels}
Brandon Trabucco, Kyle Doherty, Max Gurinas, and Ruslan Salakhutdinov.
\newblock {Effective Data Augmentation with Diffusion Models}.
\newblock arXiv:2302.07944, 2023.

\bibitem[Tremblay et~al.(2021)Tremblay, Halder, {de Charette}, and
  Lalonde]{TremblayIJCV21RainRendering4EvaluatingAndImprovingRobustness2BadWeather}
Maxime Tremblay, Shirsendu~Sukanta Halder, Raoul {de Charette}, and
  Jean-François Lalonde.
\newblock {Rain Rendering for Evaluating and Improving Robustness to Bad
  Weather}.
\newblock \emph{{IJCV}}, 129, 2021.

\bibitem[Volpi et~al.(2021)Volpi, Larlus, and Rogez]{volpi2021cvpr}
Riccardo Volpi, Diane Larlus, and Gregory Rogez.
\newblock Continual adaptation of visual representations via domain
  randomization and meta-learning.
\newblock In \emph{Proc. {CVPR}}, 2021.

\bibitem[{von Stumberg} et~al.(2020){von Stumberg}, Wenzel, Khan, and
  Cremers]{vonStumbergRAL20GNNetTheGausNewtonLoss4MultiWeatherRelocalization}
Lukas {von Stumberg}, Patrick Wenzel, Qadeer Khan, and Daniel Cremers.
\newblock {GN-Net}: The {Gauss-Newton} loss for multi-weather relocalization.
\newblock \emph{{IEEE} {Robotics and Automation} Letters}, 5\penalty0 (2),
  2020.

\bibitem[Wang et~al.(2021)Wang, Yang, and
  Betke]{WangAAAI21ConsistencyRegularizationDASIS}
Kaihong Wang, Chenhongyi Yang, and Margrit Betke.
\newblock {Consistency Regularization with High-dimensional Non-adversarial
  Source-guided Perturbation for Unsupervised Domain Adaptation in
  Segmentation}.
\newblock In \emph{Proc. {AAAI}}, 2021.

\bibitem[Wang et~al.(2022)Wang, Shen, Zuo, Zhou, and Zheng]{wang2022transvpr}
Ruotong Wang, Yanqing Shen, Weiliang Zuo, Sanping Zhou, and Nanning Zheng.
\newblock {Trans{VPR}: Transformer-based place recognition with multi-level
  attention aggregation}.
\newblock In \emph{Proc. {CVPR}}, 2022.

\bibitem[Wang \& Isola(2020)Wang and Isola]{wang2020understanding}
Tongzhou Wang and Phillip Isola.
\newblock Understanding contrastive representation learning through alignment
  and uniformity on the hypersphere.
\newblock In \emph{Proc. {ICML}}, 2020.

\bibitem[Weinzaepfel et~al.(2022)Weinzaepfel, Lucas, Larlus, and
  Kalantidis]{weinzaepfel2022learning}
Philippe Weinzaepfel, Thomas Lucas, Diane Larlus, and Yannis Kalantidis.
\newblock {Learning Super-Features for Image Retrieval}.
\newblock In \emph{Proc. {ICLR}}, 2022.

\bibitem[Wortsman et~al.(2022)Wortsman, Ilharco, Gadre, Roelofs, Gontijo-Lopes,
  Morcos, Namkoong, Farhadi, Carmon, Kornblith, et~al.]{wortsman2022model}
Mitchell Wortsman, Gabriel Ilharco, Samir~Ya Gadre, Rebecca Roelofs, Raphael
  Gontijo-Lopes, Ari~S Morcos, Hongseok Namkoong, Ali Farhadi, Yair Carmon,
  Simon Kornblith, et~al.
\newblock Model soups: averaging weights of multiple fine-tuned models improves
  accuracy without increasing inference time.
\newblock In \emph{Proc. {ICML}}, 2022.

\bibitem[Wu et~al.(2021)Wu, Wu, Guo, Ju, and
  Wang]{WuCVPR21DANNetOneStageDANetworkUnsupervisedNighttime}
Xinyi Wu, Zhenyao Wu, Hao Guo, Lili Ju, and Song Wang.
\newblock {DANNet: A One-Stage Domain Adaptation Network for Unsupervised
  Nighttime Semantic Segmentation}.
\newblock In \emph{Proc. {CVPR}}, 2021.

\bibitem[Wu et~al.(2018)Wu, Han, Lin, Gokhan~Uzunbas, Goldstein, Nam~Lim, and
  Davis]{WuECCV18DCANDualChannelWiseAlignmentNetworksUDA}
Zuxuan Wu, Xintong Han, Yen-Liang Lin, Mustafa Gokhan~Uzunbas, Tom Goldstein,
  Ser Nam~Lim, and Larry~S. Davis.
\newblock {DCAN: Dual Channel-wise Alignment Networks for Unsupervised Scene
  Adaptation}.
\newblock In \emph{Proc. {ECCV}}, 2018.

\bibitem[Xin et~al.(2019)Xin, Cai, Lu, Xing, Cai, Zhang, Yang, and
  Wang]{XinICRA19LocalizingDiscriminativeVisualLandmarksPlaceRecogn}
Zhe Xin, Yinghao Cai, Tao Lu, Xiaoxia Xing, Shaojun Cai, Jixiang Zhang, Yiping
  Yang, and Yanqing Wang.
\newblock {Localizing Discriminative Visual Landmarks for Place Recognition}.
\newblock In \emph{Proc. {ICRA}}, 2019.

\bibitem[Xu et~al.(2021)Xu, Ma, Wu, Long, and
  Huang]{XuICCV21CDAdaACurriculumDA4NighttimeSiS}
Qi~Xu, Yinan Ma, Jing Wu, Chengnian Long, and Xiaolin Huang.
\newblock {CDAda: A Curriculum Domain Adaptation for Nighttime Semantic
  Segmentation}.
\newblock In \emph{Proc. {ICCV}}, 2021.

\bibitem[Xue et~al.(2023)Xue, Budvytis, and
  Cipolla]{XueCVPR23SFD2SemanticGuidedFeatureDetectionDescription}
Fei Xue, Ignas Budvytis, and Roberto Cipolla.
\newblock {SFD2: Semantic-Guided Feature Detection and Description}.
\newblock In \emph{Proc. {CVPR}}, 2023.

\bibitem[Yang et~al.(2021)Yang, Zhong, Tang, Ding, Sebe, and
  Ricci]{Yang21BiMixBidirectionalMixing4DANighttimeSiS}
Guanglei Yang, Zhun Zhong, Hao Tang, Mingli Ding, Nicu Sebe, and Elisa Ricci.
\newblock {Bi-Mix: Bidirectional Mixing for Domain Adaptive Nighttime Semantic
  Segmentation}.
\newblock arXiv:2111.10339, 2021.

\bibitem[You et~al.(2016)You, Tan, Kawakami, Mukaigawa, and
  Ikeuch]{YouPAMI16AdherentRaindropModelingDetectionAndRemovalInVideo}
Shaodi You, Robby~T. Tan, Rei Kawakami, Yasuhiro Mukaigawa, and Katsushi
  Ikeuch.
\newblock {Adherent Raindrop Detection and Removal in Video}.
\newblock \emph{{IEEE. Trans. PAMI}}, 38\penalty0 (9), 2016.

\bibitem[Yu et~al.(2018)Yu, Chaturvedi, Feng, Taguchi, Lee, Fernandes, and
  Ramalingam]{YuIROS18VLASEVehicleLocalizationAggregatingSemanticEdges}
Xin Yu, Sagar Chaturvedi, Chen Feng, Yuichi Taguchi, Teng-Yok Lee, Clinton
  Fernandes, and Srikumar Ramalingam.
\newblock {VLASE: Vehicle Localization by Aggregating Semantic Edges}.
\newblock In \emph{Proc. {IROS}}, 2018.

\bibitem[Yun et~al.(2019)Yun, Han, Oh, Chun, Choe, and
  Yoo]{YunICCV19CutMixRegStrategyToTrainStrongClassifWithLocFeats}
Sangdoo Yun, Dongyoon Han, Seong~Joon Oh, Sanghyuk Chun, Junsuk Choe, and
  Youngjoon Yoo.
\newblock {CutMix: Regularization Strategy to Train Strong Classifiers with
  Localizable Feature}.
\newblock In \emph{Proc. {ICCV}}, 2019.

\bibitem[Zamir et~al.(2016)Zamir, Hakeem, Gool, Shah, and
  Richard]{ZamirB16LargeScaleVisualGeoLocalization}
{Amir R.} Zamir, Asaad Hakeem, {Luc Van} Gool, Mubarak Shah, and Szeliski
  Richard.
\newblock \emph{{Large-Scale Visual Geo-localization}}.
\newblock Advances in Computer Vision and Pattern Recognition. 2016.

\bibitem[Zhang \& Patel(2018)Zhang and
  Patel]{ZhangCVPR18DenselyConnectedPyramidDehazingNetwork}
He~Zhang and Vishal~M. Patel.
\newblock {Densely Connected Pyramid Dehazing Network}.
\newblock In \emph{Proc. {CVPR}}, 2018.

\bibitem[Zhang et~al.(2023)Zhang, Rao, and
  Agrawala]{ZhangICCV2AddingConditionalControl2T2IDiffusionModels}
Lvmin Zhang, Anyi Rao, and Maneesh Agrawala.
\newblock {Adding Conditional Control to Text-to-Image Diffusion Models}.
\newblock In \emph{Proc. {ICCV}}, 2023.

\bibitem[Zhang et~al.(2021)Zhang, Sattler, and
  Scaramuzza]{ZhangIJCV21ReferencePoseGenerationVisLoc}
Zichao Zhang, Torsten Sattler, and Davide Scaramuzza.
\newblock {Reference Pose Generation for Long-term Visual Localization via
  Learned Features and View Synthesis}.
\newblock \emph{{IJCV}}, 129, 2021.

\bibitem[Zheng et~al.(2020)Zheng, Wu, Han, and
  Shi]{ZhengECCV20ForkGANSeeingIntoRainyNight}
Ziqiang Zheng, Yang Wu, Xinran Han, and Jianbo Shi.
\newblock {ForkGAN: Seeing into the Rainy Night}.
\newblock In \emph{Proc. {ECCV}}, 2020.

\bibitem[Zhou et~al.(2022)Zhou, Yang, Loy, and Liu]{zhou2022learning}
Kaiyang Zhou, Jingkang Yang, Chen~Change Loy, and Ziwei Liu.
\newblock Learning to prompt for vision-language models.
\newblock \emph{{IJCV}}, 130\penalty0 (9), 2022.

\end{thebibliography}
\bibliographystyle{iclr2024_conference}

\newpage

\etocdepthtag.toc{mtappendix}
\etocsettagdepth{mtappendix}{subsection}
\etocsettagdepth{mtchapter}{none}

{
  \hypersetup{linkcolor=black}
  \tableofcontents
}
\appendix
\section{Details on implementation and experimental protocol}
\label{sec:app_implementation_details}

\subsection{Datasets}
In the main paper, we evaluate \retforloc models on the following datasets:

\nodotparagraph{Aachen Day-Night v1.1}~\citep{sattler2012image,ZhangIJCV21ReferencePoseGenerationVisLoc} is a small dataset of high-resolution images.
It contains 6,697 mapping images taken from handheld cameras under daytime in the historical city center of Aachen, Germany.
Meanwhile, its 1,015 test queries were taken with different mobile phones at day and at night.
This dataset encapsulates the scenario of city-wide outdoor localization of users thanks to easily accessible phone photos.
Evaluation is presented separately to day and night images, as the second scenario is more challenging due to day-night domain shift. 

\nodotparagraph{Robotcar Seasons v1}~\citep{maddern20171} consists of an outdoor dataset extracted in Oxford, UK, thanks to three cameras (left, right and rear views) rigged on top of a vehicle roaming around the city. Its 20,864 mapping images were captured during day time, under overcast weather, while its 11,934 queries were captured during both day and night, and under a variety of weather and seasonal conditions. Following~\cite{sattler2018benchmarking}, we report results separately on day and night queries, and only on rear view queries. These splits represent the same day-night shift seen on Aachen Day-Night on top of the domain shift 
related to the seasonal domain shift.
    We also present results on the more recent split named \textbf{Robotcar Seasons v2}~\citep{ToftPAMI22LongTermVisualLocalizationRevisited}, that mixes day and night queries and has a larger (26,121) mapping set. Since kapture does not provide off-the-shelf 3D map for this split, we 
    evaluate our models on \textbf{v2} only for EWB pose approximation.  On \textbf{v1} we report results for both EWB and accurate camera localization using the 
    provided Global-SfM map.

\nodotparagraph{Extended CMU Seasons}~\citep{badino2011cmu,ToftPAMI22LongTermVisualLocalizationRevisited} is based on the CMU Visual Localization Dataset~\citep{badino2011cmu}, and contains over 100,000 images collected over a period of 12 months in Pittsburgh, USA, with two cameras rigged on top of an SUV.
The version of the dataset proposed by \cite{ToftPAMI22LongTermVisualLocalizationRevisited} contains 24 separate sequences used to build submaps, as the full dataset is too large to build a single 3D model. These 24 sequences are split into a validation set, composed of 10 sequences for which the query poses are made available, and a test set, composed of 14 sequences for which no query poses are available and performance can only be tested on the \textit{Long Term Visual Localization Benchmark} server.\footnote{\url{https://www.visuallocalization.net/}}
We present results on both splits, referred to as ECMU-val and ECMU-test respectively. Additionally, we present results grouped according to the queried environment (urban, suburban or park) as well as weather and seasonal conditions (sunny, foliage, snow, etc.) in Table~\ref{tab:ecmu-test}.

\nodotparagraph{Gangnam Station B2}~\citep{lee2021large} is an indoor dataset collected in a floor of the Gangnam metro station in Seoul, Korea. It is composed of 4,518 mapping and 1,786 test queries, collected with both industrial and phone cameras rigged on a dedicated mapping device. Images were collected at a crowed closed space, so scene occlusion is one of its main challenges. In addition, the dataset contains many digital signage and platform screen doors that change appearance over time. 
As this dataset spans a smaller physical space than other localization datasets, its localization accuracy thresholds are lower: 1m/5\degree, 0.25m/2\degree, 0.1m/1\degree, for low, mid, and high-level accuracy, respectively. More details about the localization metrics are presented in~\Cref{sec:metrics}.

\nodotparagraph{Baidu Mall dataset}~\citep{sun2017dataset} was collected in a shopping mall of over 5,000 square meters in China. The dataset is composed of 689 mapping images captured with high-resolution cameras on an empty mall and 2,292 phone camera images taken when the mall was open. This indoor dataset contains repetitive structures, reflective and transparent surfaces, and scene occlusion, which represent a source of challenges complementary to what we observe on outdoor localization. 

\nodotparagraph{Tokyo-24/7}~\citep{Torii-CVPR2015} is a place recognition dataset that
contains 6,332 geo-tagged street-view panoramas from Tokyo, Japan, which are then split into the 75,984 views we use as retrieval gallery. As queries, we use 315 images collected with phone cameras at 125 different locations and times.

\nodotparagraph{Pittsburgh-30k}~\citep{ArandjelovicCVPR16NetVLADPlaceRecognition} is a place recognition dataset, subset of the larger Pittsburgh-250k dataset~\citep{torii2013visual}. Its test set contains 10,000
geo-tagged images downloaded from Google Street View, and 6,816 query images from the Google Pittsburgh Research Dataset. These queries are extracted from the same locations as the gallery images, but at different times and across several months.

\subsection{Training details} \label{sec:app_training_details}

We build on top of the HOW codebase\footnote{\url{https://github.com/gtolias/how}} and mostly follow the protocol for training models from HOW, including hyper-parameters.
More precisely, we finetune a ResNet50 model~\citep{he2016resnet} pretrained on ImageNet-1K~\citep{ilsvrc}. 
We remove the global average pooling and the last fully connected layers from the ResNet50 model, and attach $\ell_2$-norm based spatial pooling and dimensionality reduction (from 2048 dimensions to 128, obtained by PCA) layers to obtain the final representation of an image.
We use ``randomly resized crops'' of size $768\times768$ with the default parameters from torchvision~\citep{paszke2019pytorch} during training as additional data-augmentation, as well as to provide the network with fixed sized inputs.
This improves the computational aspect of our approach especially when we use multiple synthetic variants of an image in the same batch during training.

On top of that, we apply AugMix~\citep{hendrycks2019augmix} data augmentations to random crops with 
its default hyper-parameters suggested by the authors:\footnote{\url{https://github.com/google-research/augmix}} 
severity of augmentation operators: 3, width of augmentation chain: 3, depth of augmentation chain: -1, probability coefficient for Beta and Dirichlet distributions: 1. 
The exception is the list of augmentations, which is appended by color, contrast, brightness and sharpness transformations. We also removed translation and shear, which do not preserve the geometry of an image's local features.

We perform ``episodic'' training on SfM-120K in the sense that in each episode, we sample 2000 random query-positive pairs and a pool of 20000 images to be used for hard negative mining.
We train each model for 50 such episodes.
Given that the size of the training set is close to 100K, our models barely see each image %only 
once during training.
Thanks to using fixed-sized images as network input, we can adjust batch size to be proportional to the number of tuples during training.
To normalize the impact of the number of tuples into gradients, we divide~\Cref{eq:emb_avg} by the number of tuples used in the batch.
Within the limits of our compute infrastructure, using 5 tuples, where each tuple consists of 1 query, 1 positive and 5 negative images, gives the best performance. We used $M=5$  negatives in the tuple as this is the default setting used by HOW. 
We set the learning rate and weight decay arguments for the AdamW optimizer for each model configuration separately (\eg for \retforloc, \Synth, \etc), however, learning rate 1e-5 and weight decay 3e-2 work the best in many cases.

\myparagraph{Issues with variance} We observe large variance in performance when training  with the exact same setup but different seeds. This is also observed up to some degree when training the baseline HOW. We therefore choose to report results for each setup after averaging the weights~\citep{wortsman2022model} of 3 models, \ie obtained with the exact same setup starting from 3 different seeds.

\myparagraph{Retrieval at test-time}
We use the ASMK implementation\footnote{\url{https://github.com/jenicek/asmk}} that further uses the FAISS library~\citep{johnson2019billion}\footnote{\url{https://github.com/facebookresearch/faiss}} for sub-linear indexing. We use 3000 local features for matching and learn the ASMK codebook on SfM-120k. We follow the protocol of HOW/FIRe and perform multi-scale testing over 7 scales.

\subsection{Metrics and evaluation protocols} \label{sec:metrics}

In this section, we further describe the evaluation protocol we briefly mentioned in Section~\ref{sec:experiments} of the main paper. 
As mentioned, our retrieval models output, 
for each query image, a list of the $k$ nearest neighbors among the mapping images, 
for which we know the camera pose. 
For EWB, the approximated pose is simply obtained by an arithmetic mean of the poses of these mapping images. 
This protocol is very light-weight, and ideal for real-time localization scenarios. Its drawback is the lack of any geometric verification for the proposed pose. The quality of the approximated pose also depends  
if images taken from   similar poses  exist amongst 
the mapping images.
Indeed, EWB for $k=1$ is equivalent of approximating the query pose by the pose of its most similar mapping image.  As we increase the value of $k$, performance tends to fall as we account for 
more distant similar images or with different viewpoints, yielding in general a less accurate barycenter.

For Global-SfM, we extract local features for all retrieved $k$ mapping images 
and feed together   with the query image to the kapture-localization pipeline. As these mapping images are from the ones used to compute the SfM model of the scene (in our case with COLMAP~\citep{schoenberger2016sfm}), the SfM  model contains 3D correspondences  to  2D local features 
for these mapping images. Hence, once we establish 2D-2D matching of local features between the query image and the top ranked mapping images,  we obtain a set of 2D-3D matches between the 3D scene representation  and the query image which are used for the query pose estimation via PNP and RANSAC.

In our experiments, both the local features and the COLMAP are obtained  off-the-shelf from kapture\footnote{\url{https://github.com/naver/kapture}}.They both rely on the  `featheR2D2\_32\_20k' local features, --- a variant of R2D2~\citep{revaud2019r2d2} ---
which consists of  20K  32 dimensional features extracted per image. 
In contrast to EWB, the Global-SfM protocol obtains better accuracy as we increase the value of k, since more relative poses can be taken into account and used for triangulation to estimate the  query pose.
On the other hand, larger values of $k$ also increase the computational cost of the protocol, as local feature matching is often the main computational bottleneck. Therefore, for practical applications, the value of $k$ should be selected such that it provides good pose estimation within a time budget.

For all protocols, we measure localization performance by the percentage of correctly estimated camera poses withing a given error threshold. Indeed, given the groundtruth and the estimated positions, we calculate the translation and rotation differences between the two positions~\citep{humenberger2022investigating}, 
and we consider an image as successfully localized if the translation error is below X meters and the rotation error is below Y\degree. We report the percentage of localized queries at three levels of accuracy: low-level (X=5m, Y=10\degree), mid-level (X=0.5m, Y=5\degree) and high-level (X=0.25m, Y=2\degree) accuracy\footnote{As mentioned above, for Gangnam Station  the thresholds are 1m/5\degree, 0.25m/2\degree, 0.1m/1\degree , for low, mid, and high-level accuracy}. 
We prioritize showing different levels of accuracy for different protocols: for EWB, as a first-level approximation, we show low-level accuracy, (in many cases, depending on the mapping images, the high accuracy level is not achievable even with an oracle),
whereas for Global-SfM, the target is precise pose %more precise 
estimation, therefore we prioritize mid and high-level accuracy. 

For place recognition, we follow \cite{Torii-CVPR2015} and consider that two images represent the same place if their camera positions are at most 25 meters apart. We compare performance by Top-k recall (R@k), which measures the percentage of queries for which at least one of the k nearest neighbors represents the same place as the query.

\subsection{Baseline retrieval models evaluated on localization} \label{sec:baselines}

For fairness, we evaluated ourselves some of the top existing retrieval models used for visual localization under the exact same setup we use for the proposed Ret4Loc models. We evaluated the following models under the Kapture framework~\citep{HumenbergerX20RobustImageRetrievalBasedVisLocKapture}:

\begin{itemize}
    \item \textbf{NetVLAD}\footnote{We use the re-implementation from \url{https://github.com/QVPR/Patch-NetVLAD}}~\citep{ArandjelovicCVPR16NetVLADPlaceRecognition} is a model based on a generalized VLAD layer, that aggregates local convolutional features extracted with a VGG-16~\citep{simonyan2014very} backbone, and learned end-to-end with a triplet loss. The model is trained on geo-tagged data collected at different camera angles, lighting conditions and seasons. NetVLAD is a milestone work in visual place recognition and it is used as baseline for different methods in the task~\citep{hausler2021patch,leyva2023data}, and it has been used in state-of-the-art visual localization pipelines~\citep{Germain3DV19SparseToDenseHypercolumnMatchingVisLoc, SarlinCVPR19FromCoarsetoFineHierarchicalLocalization}. We report results with a model trained on Pittsburgh-250k~\citep{torii2013visual}, which outputs a global representation of $4096$ dimensions, from the PyTorch re-implementation of NetVLAD from \cite{hausler2021patch}.
    \item \textbf{AP-GeM}\footnote{\url{https://github.com/naver/deep-image-retrieval}}~\citep{revaud2019learning} is a model trained for image retrieval with a differentiable approximation of the average precision (AP) metric, which is the main evaluation metric used in landmark retrieval, which learns a global image representation obtained from the generalized-mean pooling (GeM pooling)~\citep{radenovic2018fine} of local convolutional features of a ResNet~\citep{he2016resnet}. We follow \cite{HumenbergerX20RobustImageRetrievalBasedVisLocKapture}, who shows that AP-GeM is a strong retrieval model for visual localization, and use the model trained on Google Landmark dataset v1 (GLD)~\citep{noh2017large}.
    \item \textbf{HOW}\footnote{\url{https://github.com/gtolias/how}}~\citep{tolias2020learning} is the retrieval model we use as bases for our Ret4Loc models. HOW is trained on global image representations, obtained from aggregating intermediary convolutional features with SPoC pooling~\citep{babenko2015aggregating}, with a contrastive margin loss over matching image pairs. At inference time, HOW local features are matched with ASMK~\citep{tolias2013aggregate} for improved performance. We use the model released by the authors as baseline.
    \item \textbf{FIRe}\footnote{\url{https://github.com/naver/fire}}~\citep{weinzaepfel2022learning} is a retrieval model that builds upon HOW's formula of training with a global representation while performing inference with matching lower-level representations. However, instead of aggregating local convolutional features, it uses an attention module to propose mid-level \textit{super-features} for matching.
    \cite{weinzaepfel2022learning} also first benchmarked  HOW and FIRe for visual localization on Aachen, hinting to their potential as state-of-the-art methods for the task. 
    For both HOW/FIRe, we use the ResNet50-based model released by the authors, trained on SfM-120k~\citep{radenovic2018fine}, and follow the same test-time protocol as for our models, detailed in~\Cref{sec:app_training_details}.
\end{itemize}

\subsection{Re-ranking experiments} \label{sec:rerank}

In \Cref{tab:sota} and \Cref{tab:vpr} we also report results after geometric re-ranking. 
For each query, we first obtain ranked lists using our \Synthpp model, and match the query image to each of the top 100 closest images using LightGlue~\citep{lindenberger2023lightglue}. We then re-rank the top images according to the number of inliers.

We use the public LightGlue implementation\footnote{\url{https://github.com/cvg/LightGl ue}} with SuperPoint~\citep{detone2018superpoint} features. We extract at most 3072 keypoint features per image, with all images of a dataset resized to a fixed resolution for consistency (1024 pixels as the largest dimension for Tokyo 24/7, RobotCar, and Extended CMU, and 750 pixels for Pittsburgh30k). In order to maximize accuracy, we disable early stopping and iterative point pruning.

% -------------------------------------------------------------------------
% Retrieval Results
% -------------------------------------------------------------------------
\section{Further experimental analyses and results}
\label{sec:app_experiments}

% -------------------------------------------------------------------------
% Extended Results
% -------------------------------------------------------------------------
\subsection{Statistics for the filtering threshold $\tau$}

Here, we study the effect of the selection threshold $\tau$.

\begin{figure}[h]
     \centering
     \begin{subfigure}[b]{.45\textwidth}
         \centering
        \includegraphics[width=\linewidth]{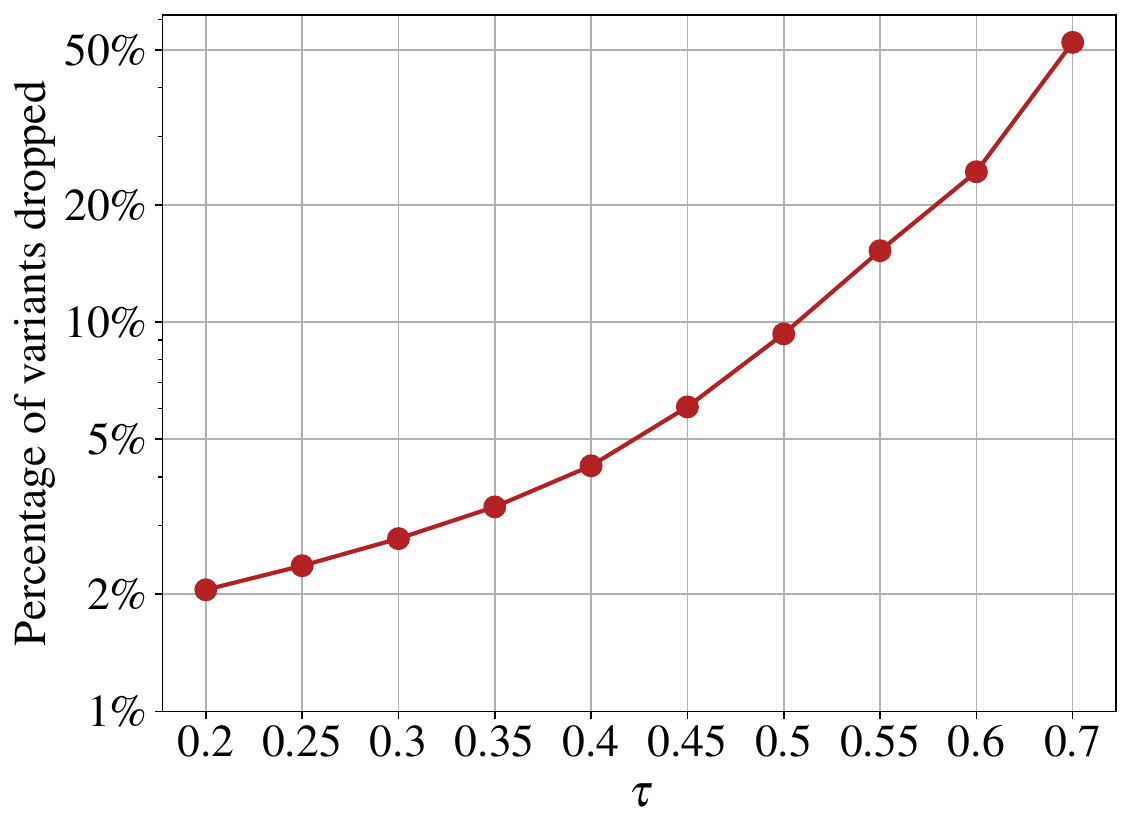}
         \vspace{-4pt}
         \caption{Synthetic pairs dropped for $0.2 \ge \tau \ge 0.7$.}
         \label{fig:perc_variant_droped_per_threshod}
     \end{subfigure}
     \begin{subfigure}[b]{.49\textwidth}
         \centering
         \includegraphics[width=\textwidth]{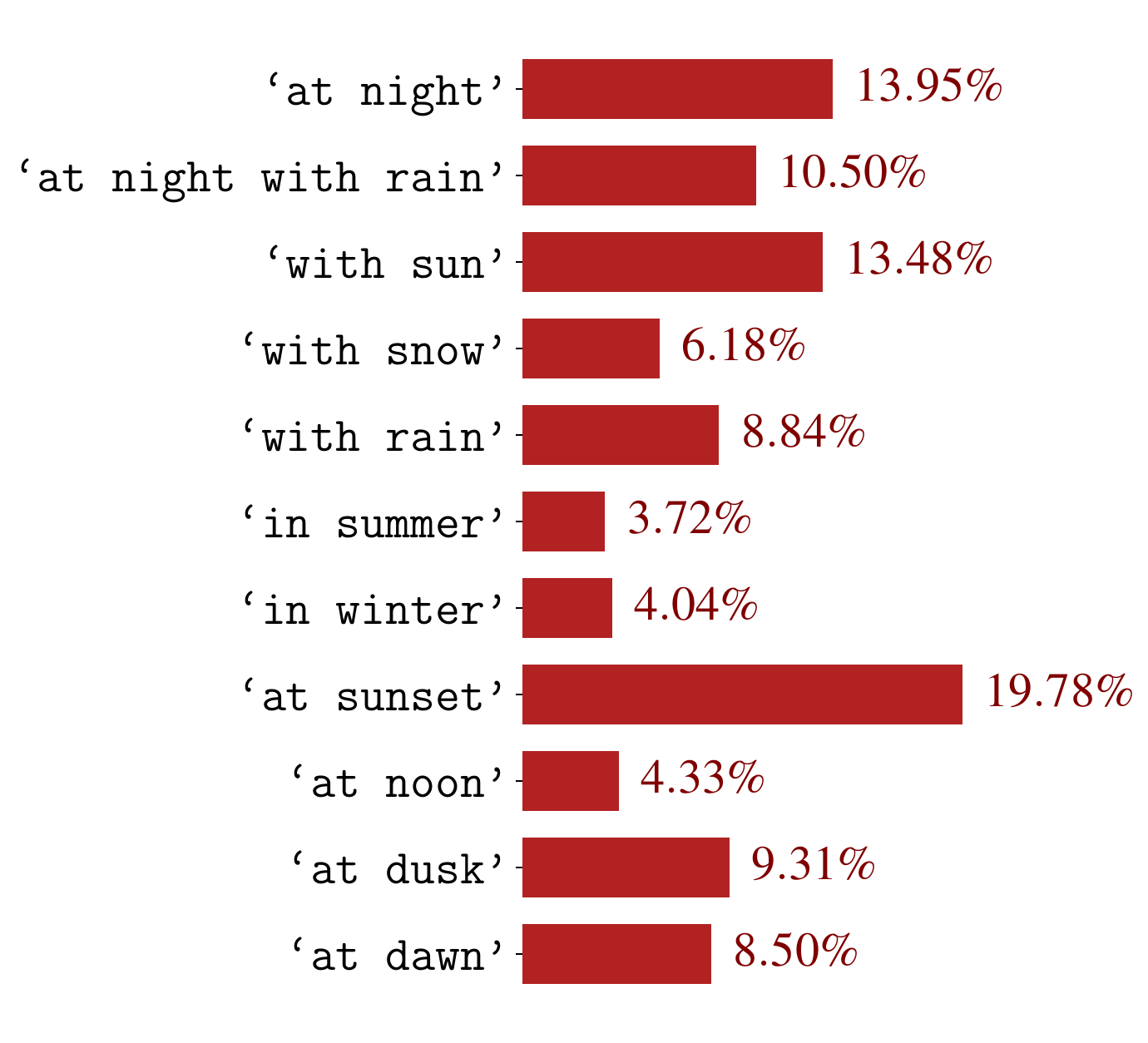}
         \vspace{-4pt}
         \caption{Synthetic pairs dropped per prompt for $\tau=0.5$.}
         \label{fig:drop_per_variant}
     \end{subfigure}
     \caption{(Left) Percentage of synthetic pairs dropped for different thresholds $\tau$ on the geometric consistency score $s$. (Right) Percentage of synthetic pairs dropped per textual prompt for $\tau=0.5$.}
    \label{fig:tau_ablation}
\end{figure}

In~\Cref{fig:tau_ablation} we plot on the left the percentage of synthetic pairs that would be dropped for different thresholds $\tau$ on the geometric consistency score $s$.
On the right we report the percentage of synthetic pairs that are dropped for $\tau=0.5$ for each textual prompt.
We see that \texttt{`at sunset'} has a much higher probability of being dropped; the strong colors of sunset images dominate after alteration in many cases, making geometric consistency fail.

\begin{figure}[h]
     \centering
    \includegraphics[height=4cm]{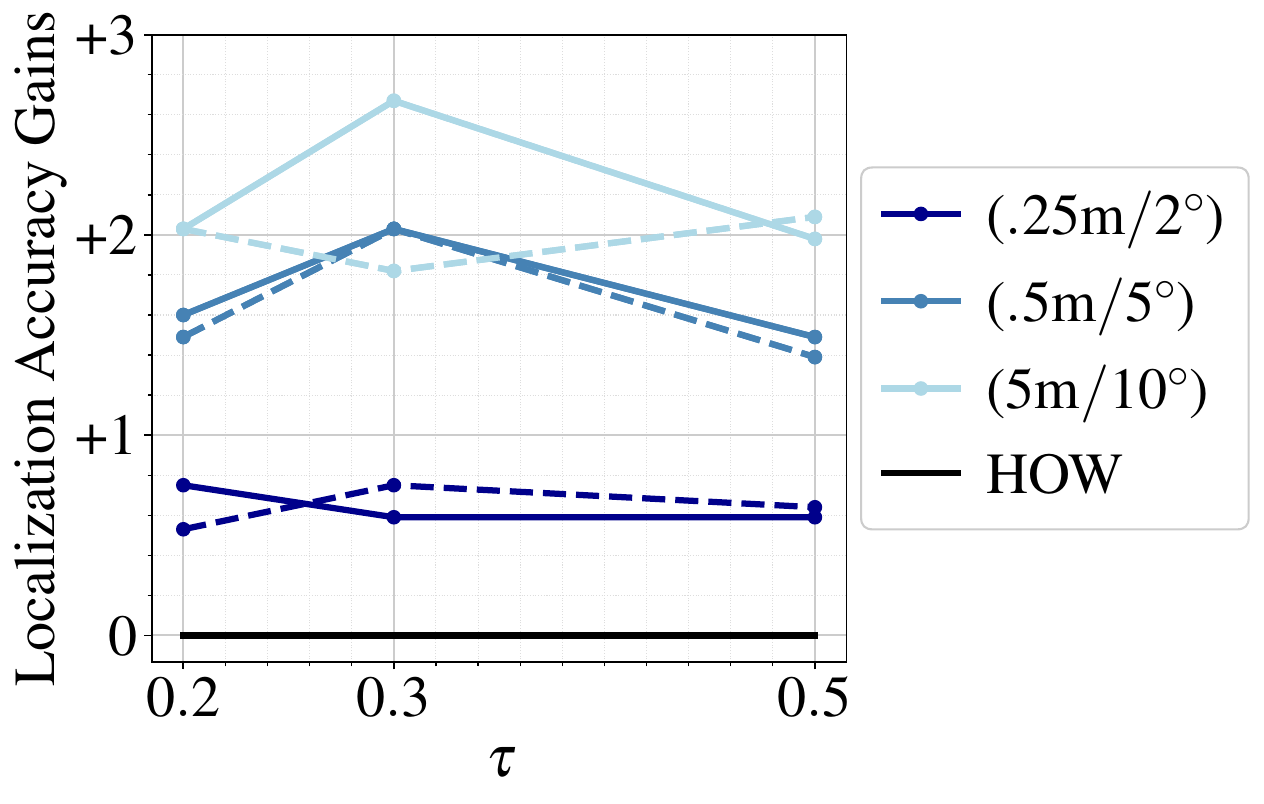}
    \captionof{figure}{\textbf{Relative gains on Robotcar-v2} for different $\tau$ values for \Synthp (solid lines) and \Synthpp (dashed lines) for localization across three error thresholds.}
    \label{fig:thres_ablation}
\end{figure}

In~\Cref{fig:thres_ablation} we study the effect of the selection threshold and report gains on RobotCar-v2 for different values of $\tau$ with and without geometric-aware sampling.
We see that performance is generally consistent for $\tau=[0.2, 0.3]$ and slightly dropping for $\tau=0.5$. We select the best value of $\tau$ among $\{0.2, 0.3\}$ via a validation set. For the two named Ret4Loc models, \Synthp uses $\tau=0.3$ and \Synthpp uses $\tau=0.2$.

\subsection{Complete sets of results for five common datasets}
\label{sec:app_extended_results}

\begin{figure}
    \centering
    \includegraphics[width=\linewidth]{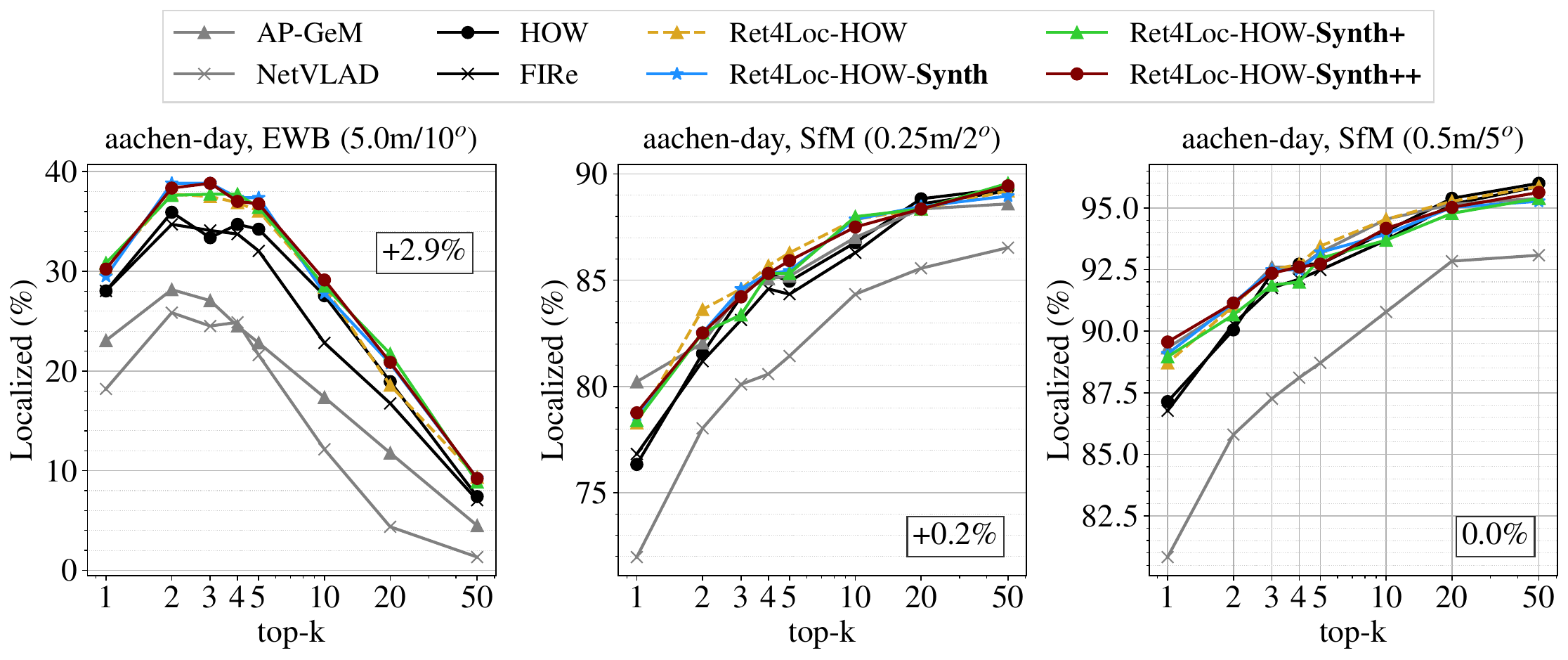} \\
    \includegraphics[width=\linewidth]{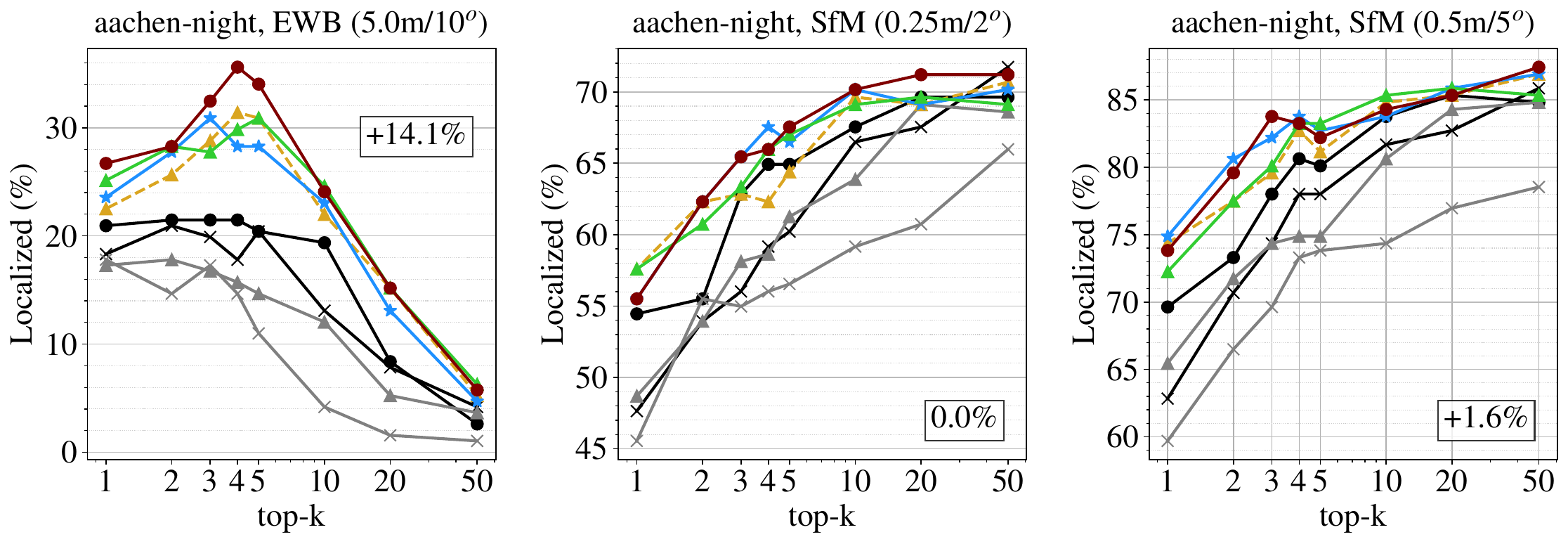} \\
    \includegraphics[width=\linewidth]{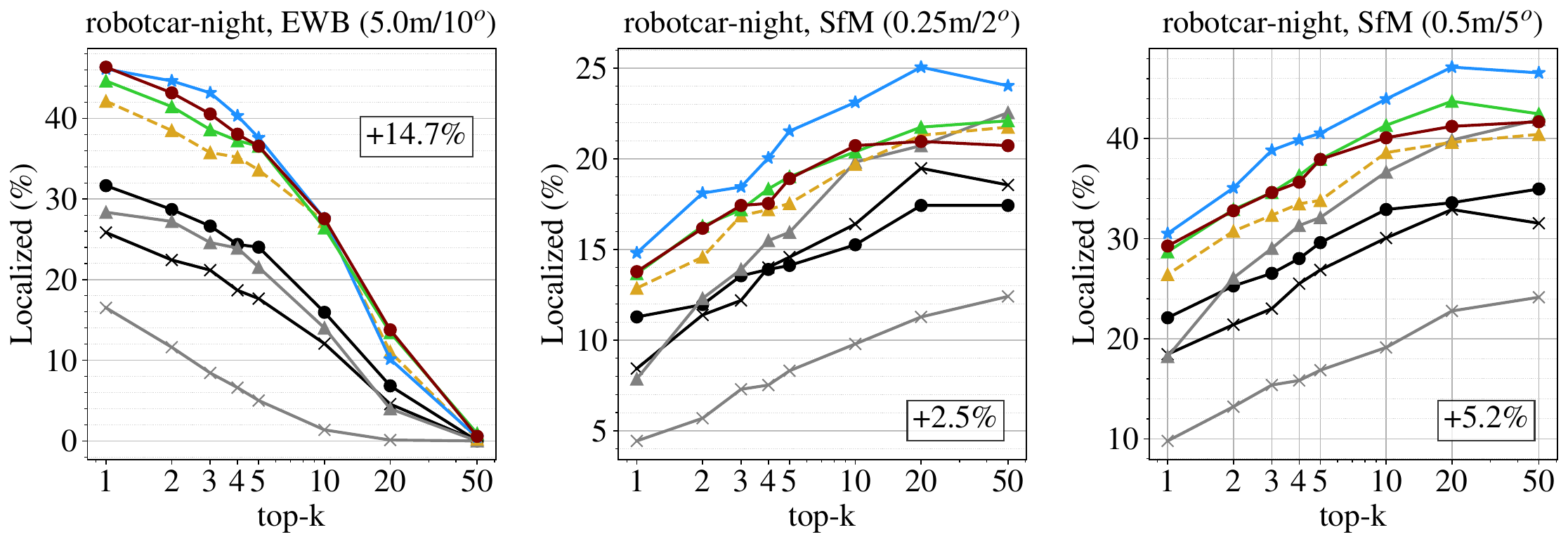} \\
    \includegraphics[width=\linewidth]{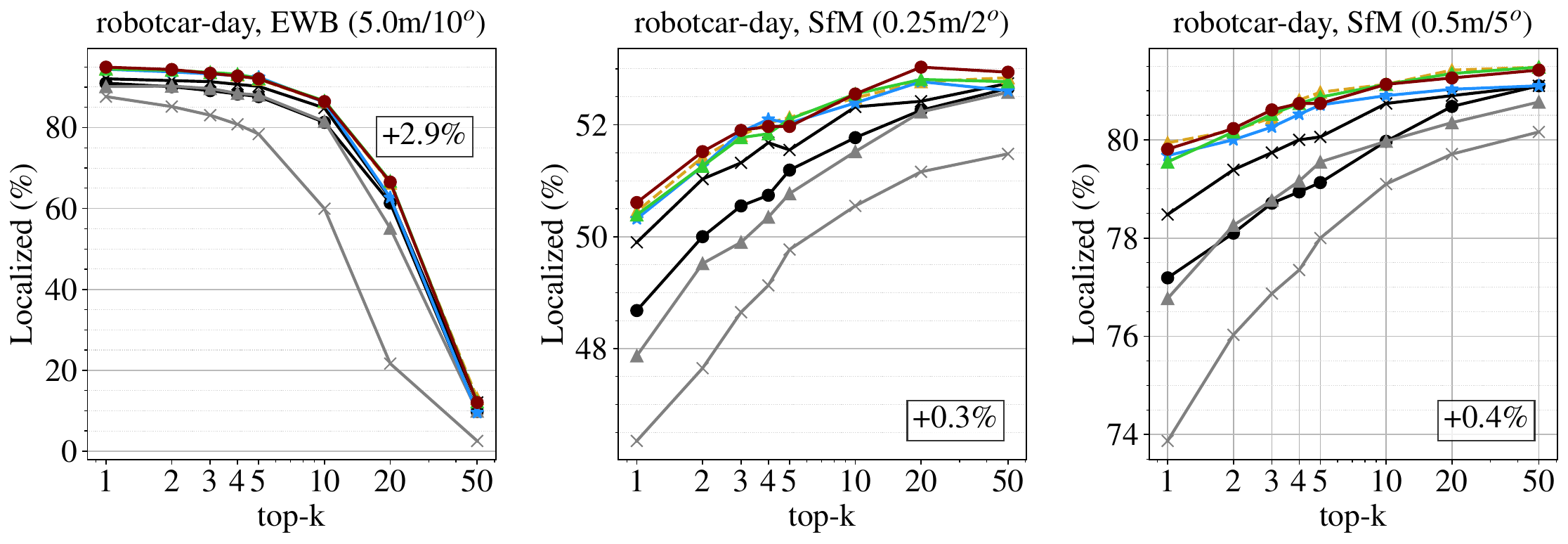} 
    \caption{\textbf{Visual localization results on Aachen Day-Night v1.1 and Robotcar Seasons (v1)}.}
    \label{fig:complete_res_aachen_robotcar}
\end{figure}

\begin{figure}
    \centering
    \includegraphics[width=\linewidth]{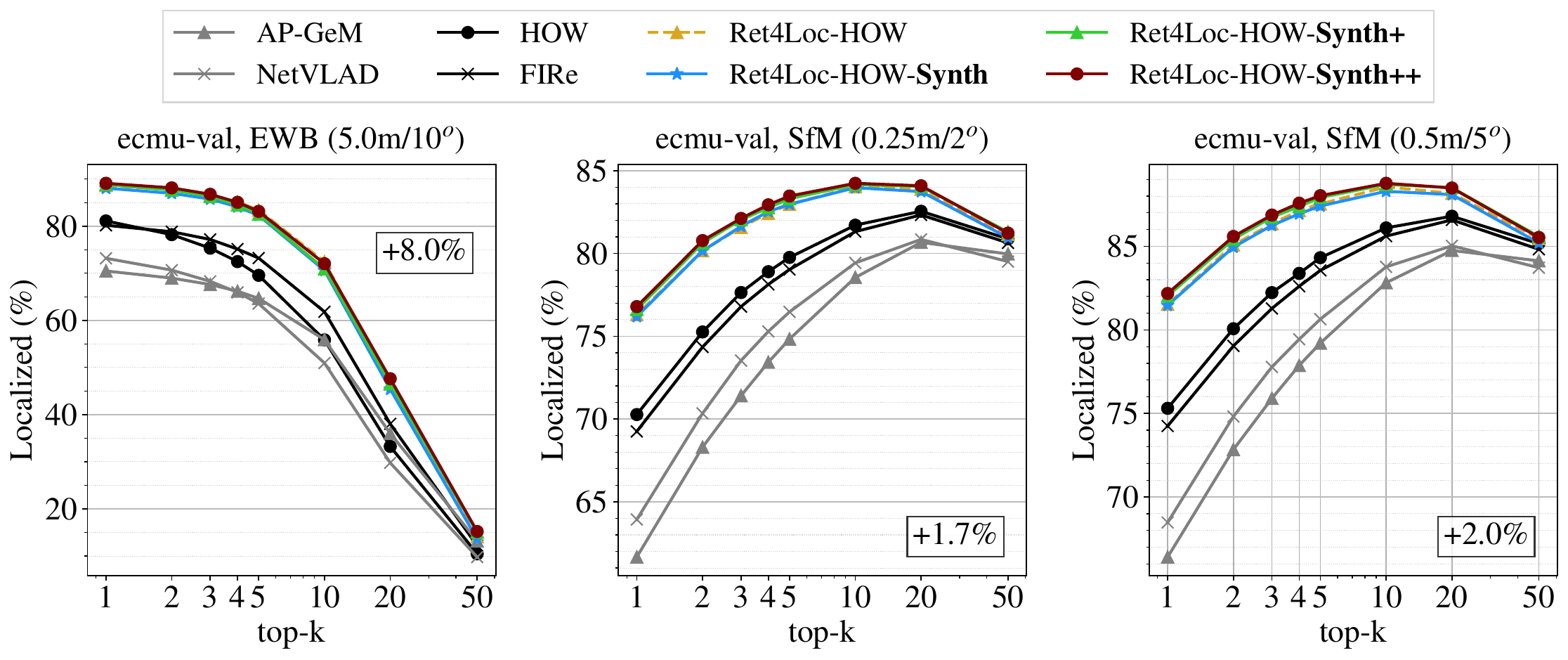} \\
    \includegraphics[width=\linewidth]{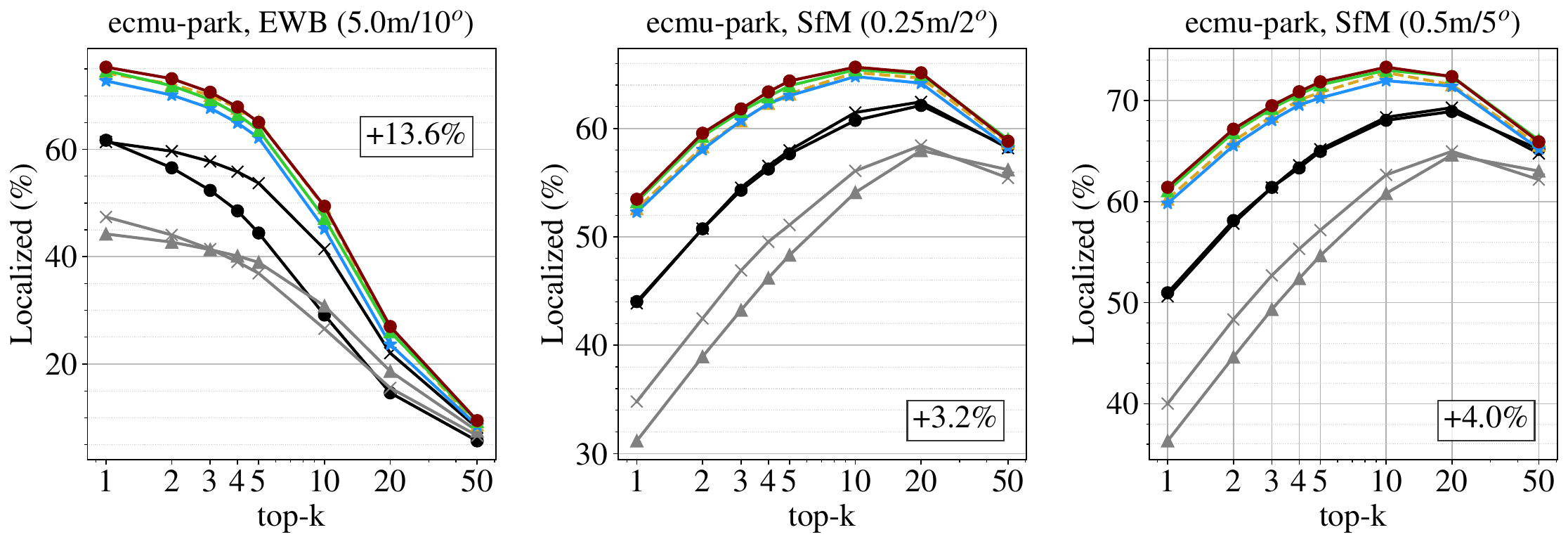} \\
    \includegraphics[width=\linewidth]{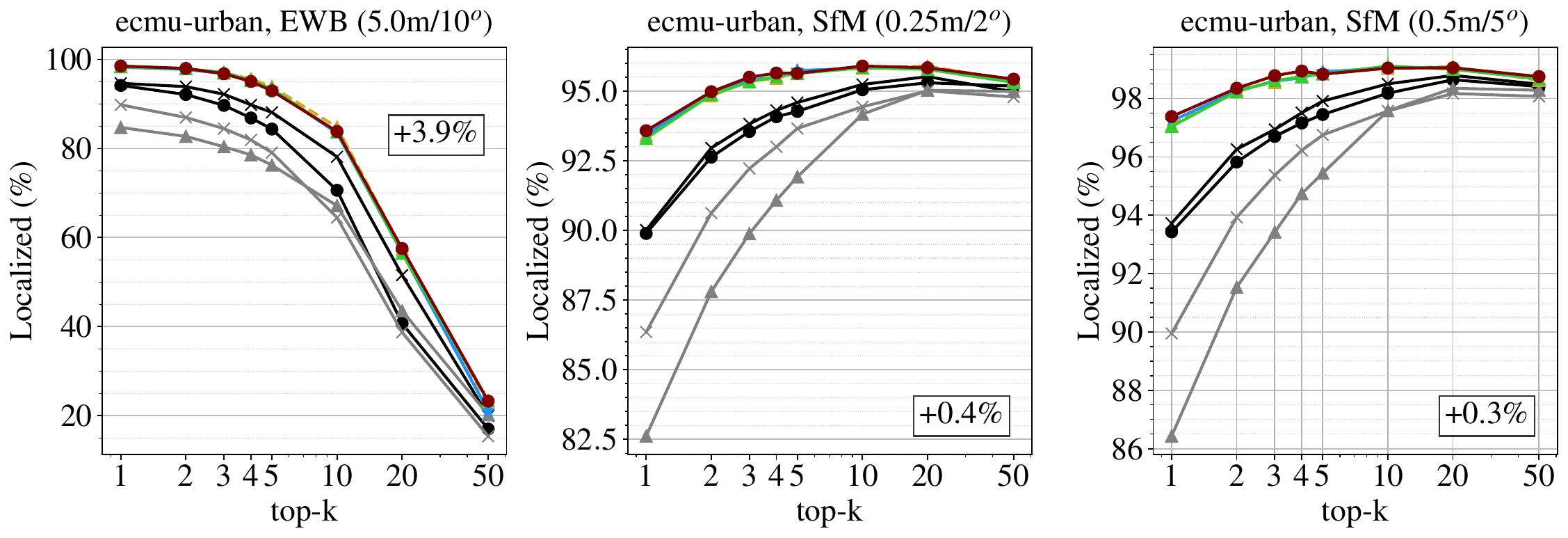} \\
    \includegraphics[width=\linewidth]{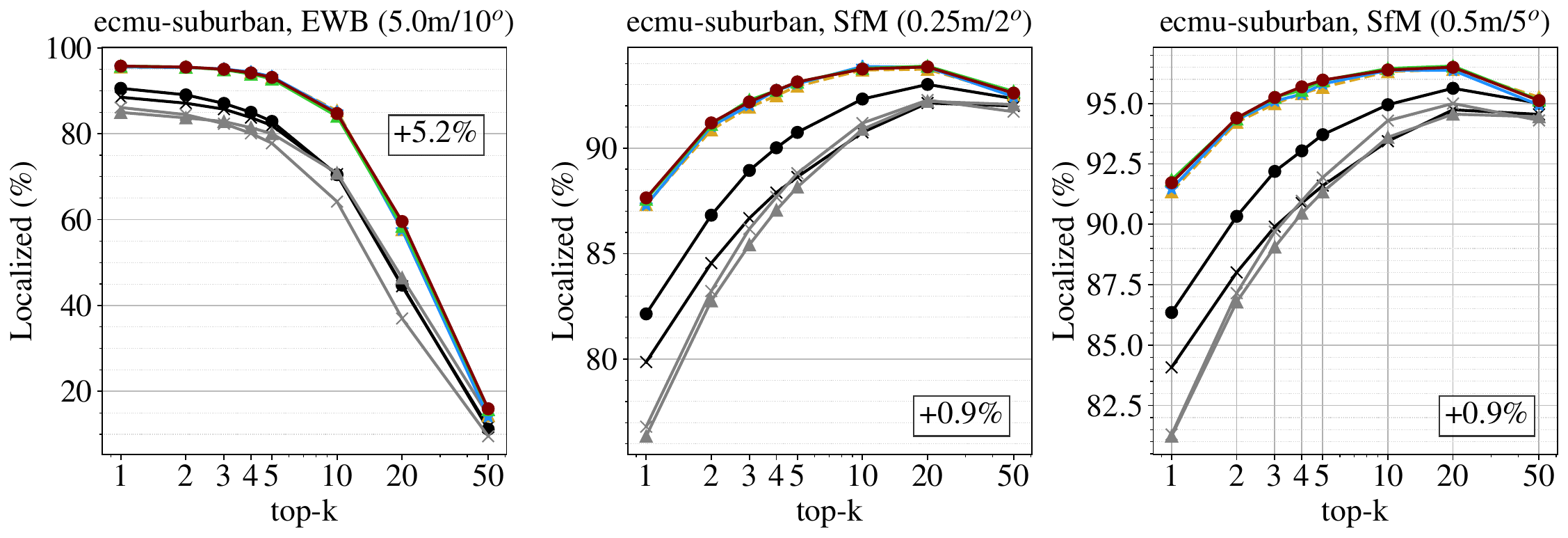} \\
    \caption{\textbf{Visual localization results on four splits of the Extended CMU Seasons dataset}.}
    \label{fig:complete_res_ecmu}
\end{figure}

\begin{figure}
    \centering
    \includegraphics[width=\linewidth]{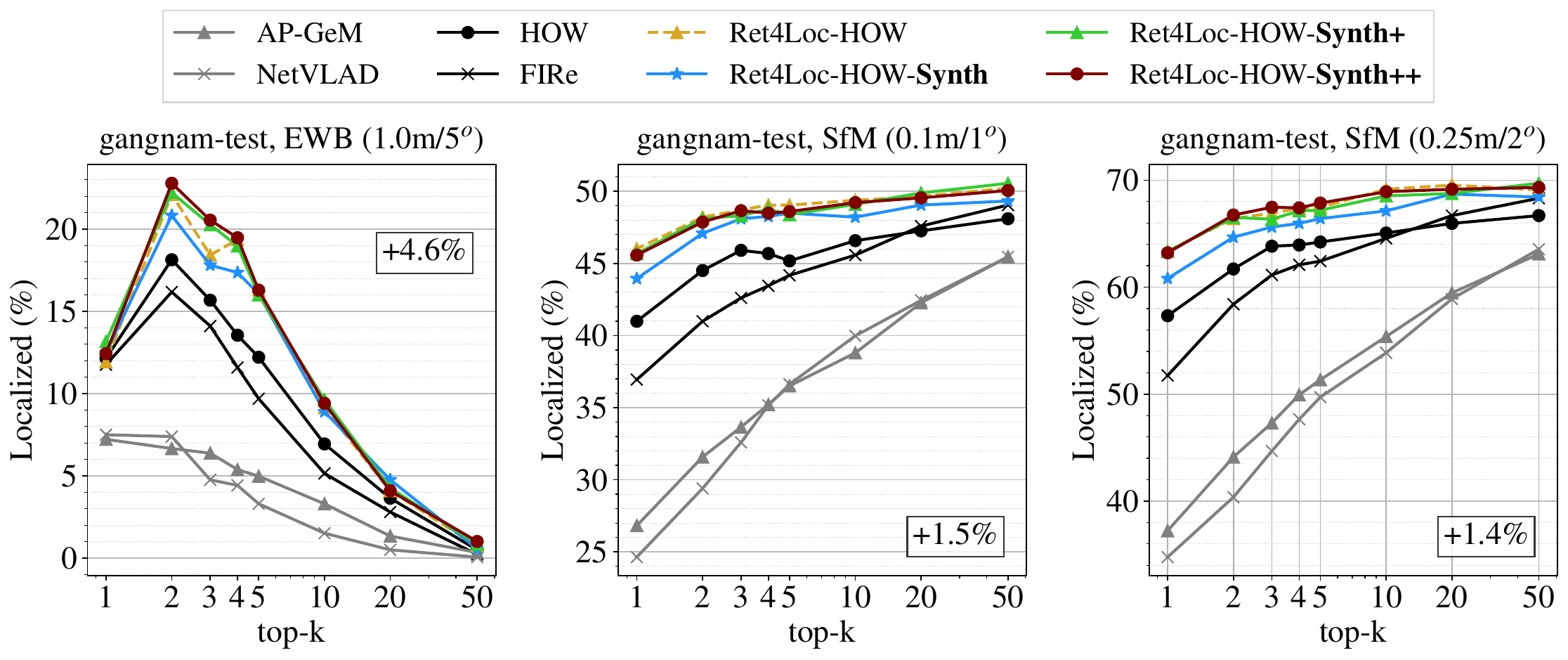} 
    \includegraphics[width=\linewidth]{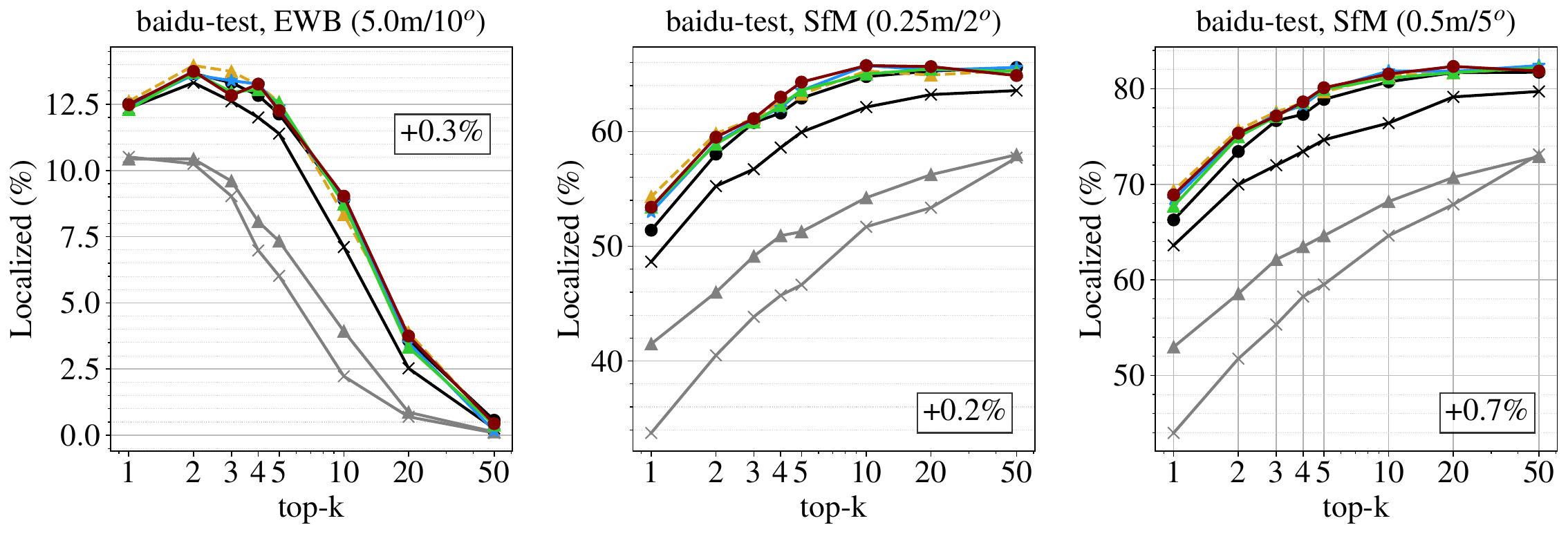} \\
    \caption{\textbf{Visual localization results on Baidu Mall and Gangnam Station B2 datasets}.}
    \label{fig:complete_res_indoor}
\end{figure}

In~\Cref{fig:complete_res_aachen_robotcar,fig:complete_res_ecmu,fig:complete_res_indoor} we show complete results for 5 popular long-term visual localization datasets available on the Visual Localization Benchmark\footnote{\url{https://www.visuallocalization.net/}} and available in the kapture platform~\citep{HumenbergerX20RobustImageRetrievalBasedVisLocKapture}, \ie Aachen Day-Night, Robotcar Seasons, Extended CMU Seasons, Baidu Mall and Gangnam Station B2. We report results for the highest error threshold for EWB and the two lowest for SfM (we explain this choice in~\cref{sec:app_implementation_details}).

\subsection{Results per query condition for ECMU-test}

We present in~\Cref{tab:ecmu-test} the in-depth results for our proposed models and some baseline on the Extended CMU Seasons test split. The results for this split are only available via submission to the Long-Term Visual Localization benchmark\footnote{\url{https://www.visuallocalization.net/}}, and it provides performance broken down according to query environment (urban, suburban and park) as well as seasonal or weather conditions.  
Firstly, we notice that our Ret4Loc-HOW models outperform HOW on all conditions, with improvements superior to 3\% on several query types where HOW struggles (\ie accuracy below 90\%), such as park, sunny, foliage, and mixed foliage. Our base \howam 
present the best results on overcast, sunny, and foliage, which validates the importance of the changes our method brought to HOW, namely the optimization and the necessity of data augmentations.
Secondly, Ret4Loc-HOW+synthetic performance on rows 3 and 4 shows straight improvements to some types of queries with respect to our base \howam, in particular to suburban, no foliage, and snow queries. Such conditions seem to be better addressed with synthetic images instead of pixel-level data augmentations. However, such improvements are not generalized, as overcast, sunny, foliage and cloudy see their results drop with respect to \howam.
These set-backs are finally addressed with the use of 
geometry-aware sampling,
as we notice in rows 5 and 6, showing the importance of the geometric consistency based sampling and filtering.  Ret4Loc-HOW+synthetic models in these lines obtain the best results for all but two conditions, and obtain the best overall results on the Extended CMU Seasons test.
% Table for ECMU-test results separated by conditions (provided automatically by https://www.visuallocalization.net/details )

% \usepackage{multirow}
{
    \newcommand{\tworowssingle}[1]{\multirow{2}{*}{#1}}
    \newcommand{\tworowssplit}[2]{\multirow{2}{1cm}{\centering #1 \\ #2}}
    \setlength{\tabcolsep}{2pt}
    \setcounter{rownumbers}{0}
    \begin{table}[h]
    \centering
    \adjustbox{max width=\textwidth}{
        \begin{tabular}{r lccc|cccccccccccc}
        \toprule
        % & \multirow{3}{*}[-3pt]{Model} & \multirow{3}{*}[-3pt]{VMix} & \multirow{3}{*}[-3pt]{GaS} & \multirow{3}{*}[-3pt]{Filt.} &   
        & & & & &   
            \multicolumn{12}{c}{\bf ECMU-Test, EWB (top-1), 5m/10\degree} \\ % \cmidrule{6-17}
        &  &  &  &  &
        \tworowssingle{Urban} & \tworowssplit{Sub-}{urban} & \tworowssingle{Park} & 
        \tworowssingle{Overcast} & \tworowssingle{Sunny} & \tworowssingle{Foliage} & 
        \tworowssplit{Mixed}{Foliage} & \tworowssplit{No}{Foliage} & \tworowssplit{Low}{Sun} & 
        \tworowssingle{Cloudy} & \tworowssingle{Snow} & \tworowssingle{All} \\ 
        & Model & VMix & GaS & Filt & \\ \midrule % \cmidrule{6-17}
        \rownum & HOW             & -- & -- & -- & 
        94.4 & 88.8 & 79.0 & 
        87.2 & 83.1 & 83.9 & 
        88.7 & 95.1 & 90.0 & 
        90.9 & 91.7 & 87.9  \\
        \rownum & \howam           & -- & -- & -- &
        95.4 & 89.5 & 86.0 & 
        \textbf{90.0} & \textbf{86.7} & \textbf{87.3} &
        91.4 & 95.6 & 92.0 & 
        93.0 & 92.9 & 90.5 \\
        \midrule
        
        % row 12 - Ret4Loc-nv3	none	uniform
        \rownum & \multirow{6}{*}{\shortstack[c]{Ret4Loc-HOW \\ + synthetic}} & -- & -- & -- &
        95.3 & 89.9 & 81.8 & 
        88.9 & 84.2 & 85.2 & 
        90.4 & 96.6 & 91.9 & 
        92.3 & 94.6 & 89.4  \\
        % row 13 - Ret4Loc-nv3-EA	none	uniform
        \rownum && \cmark & -- & -- &
        95.2 & 90.4	& 86.1 & 
        89.7 & 86.3 & 86.9 & 
        92.2 & 96.7 & 93.2 & 
        93.0 & 94.9 & 90.8  \\
        % ----------------------------------------------------------------
        \cmidrule{4-17}
        % ----------------------------------------------------------------
        &&& \multicolumn{12}{l}{\emph{Variations using geometric consistency}} \\
        % row 17 - Ret4Loc-nv3-EA	none	inv-prob
        %\rownum && \cmark & -- & \cmark &
        %95.1 & \textbf{90.9} & 86.5 & 
        %\textbf{90.0} & 86.4 & 87.1 & 
        %92.6 & 97.0 & 93.6 & 
        %\textbf{93.6} & \textbf{95.7} & 91.1 \\
        
        % row 28 - Ret4Loc-nv3-EA	t=0.3	uniform
        \rownum && \cmark & \cmark & -- &
        95.4 & \textbf{90.7} & 86.1 & 
        \textbf{90.0} & 86.3 & 86.9 & 
        92.5 & \textbf{97.2} & 93.7 & 
        93.0 & \textbf{95.7} & 91.0 \\

        % row 39 - Ret4Loc-nv3-EA	t=0.2	inv-prob
        \rownum && \cmark & \cmark & \cmark &
        \textbf{95.5} & \textbf{90.7} & \textbf{86.7} & 
        89.9 & 86.5 & 87.1 & 
        \textbf{92.8} & 97.0 & \textbf{93.9} & 
        \textbf{93.4} & 95.4  & \textbf{91.2} \\

        \bottomrule
        \end{tabular}
    }
    \caption{\textbf{Extended CMU Seasons test results broken down by query conditions.} \textbf{Bold} number represent the best results per column. \textit{VMix} refers to the use of variant mixing with~\cref{eq:emb_avg} instead of simple loss averaging,
    \textit{Filt.} to synthetic tuple filtering, \textit{GaS} to geometry-aware sampling.}
    \label{tab:ecmu-test}
    \end{table}
}

% -------------------------------------------------------------------------
% Retrieval Results
% -------------------------------------------------------------------------

\subsection{Retrieval performance on visual place recognition datasets}
\label{sec:app_retrieval_results}

In~\Cref{tab:vpr} we present results on the Tokyo 24/7 and Pittsburgh30k place recognition datasets. We see that our models consistently outperform compared methods for the case of single-stage retrieval, while, when paired with SuperPoint+LightGlue re-ranking (see \Cref{sec:rerank} for details), Ret4Loc models achieve a new state-of-the-art for Tokyo 24/7 dataset, a commonly used place recognition dataset that focuses on weather and seasonal changes.

{
\setlength{\tabcolsep}{12pt}
\setcounter{rownumbers}{0}

\def\colgroupspace{{\hskip 13pt}}
\def\colgroupspacelarge{{\hskip 40pt}}

\begin{table}[t]
    \centering
    \adjustbox{max width=.99\textwidth}{
    \begin{tabular}{rl  ccc  ccc}
    \toprule
    & \multirow{3}{*}{Model} & 
        \multicolumn{3}{c}{\textbf{Tokyo 24/7}} &
        \multicolumn{3}{c}{\textbf{Pitts30k}} \\
        && 
         \multicolumn{3}{c}{Retrieval (top-$k$ recall)}  &
         \multicolumn{3}{c}{Retrieval (top-$k$ recall)}  \\
        &&
        \small{R@1} & \small{R@5} & \small{R@10} &
        \small{R@1} & \small{R@5} & \small{R@10} \\
    \midrule
    \multicolumn{8}{l}{\emph{Single stage retrieval methods}} \\
    
    \rownum & NetVLAD$^*$  &  37.8 & 53.3  &  61.0 &  70.3 & 84.1  &  89.1 \\
    \rownum & TransVPR$^*$ & - & - & - & 73.8 & 88.1 & 91.9 \\
    \rownum & GCL$^*$  & 69.5 & 81.0 & 85.1 & 80.7 & 91.5 & 93.9 \\
    \rownum & FIRe  & 84.8	& 89.8 & 92.7 & 84.5 & 92.2 & 94.8 \\ 
    \rownum & HOW  & 89.2	&94.6	& 96.5 & 84.4 & 92.0 & 94.6 \\ 
    \cmidrule{2-8}
    \rownum & \howam  & 89.2	& \textbf{95.6}&	\textbf{97.1} & 85.0 & 92.6 & \textbf{94.9} \\  
    \rownum & \Synthpp & \textbf{91.1} & 94.0 & \textbf{97.1} & \textbf{85.9} & \textbf{92.9} & \textbf{94.9} \\
    \midrule
    \multicolumn{8}{l}{\emph{Retrieval methods \textbf{with re-ranking}}} \\
    \rownum & SP-SuperGlue$^*$ & 88.2 & 90.2 & 90.2 & 87.2 & 94.8 & 96.4 \\ 
    \rownum & DELG$^*$ & 95.9 & 96.8 & 97.1 & \textbf{89.9} & \textbf{95.4} & \textbf{96.7} \\ 
    \rownum & Patch NetVLAD$^*$ & 86.0 & 88.6 & 90.5 & 88.7 & 94.5 & 95.9 \\ 
    \rownum & TransVPR$^*$ & - & - & - & 89.0 & 94.9 & 96.2 \\ 
    \cmidrule{2-8}
    \rownum & \Synthpp & \textbf{97.5} & \textbf{98.1} & \textbf{98.4} & 88.5 & 93.8 & 96.2 \\
    
    \bottomrule
    \end{tabular}
    }
    \caption{
        \textbf{Visual place recognition results on place recognition for Tokyo 24/7 and Pittsburgh30k.} We report the usual metrics (top-$k$ recall), \ie if one correct image is retrieved in the top $k$. $^*$ denotes results from GCL~\citep{leyva2023data};
    }
    \label{tab:vpr}
\end{table}
}

\subsection{Feature space analysis}\label{sec:app_feature_analysis}

In order to better understand how individual model components impact representations, we analyze features learned by the baseline HOW model and 4 of our models: \retforloc, \Synth, \Synthp and \Synthpp. 

To this end, we sample 1000 real images from the validation set of SfM-120K and extract features for each of those images along with their 11 synthetic variants (which we generate as described in~\Cref{sec:synthetic}).
While extracting features, we resize images such that their longest side is 1024 pixels, and we do not apply any cropping to images.
In total, we obtain 12000 128-dimensional image features for each model.
Then, we create 1000 ``classes'' by grouping each real image with its synthetic variants and assign a pseudo-label to each such pseudo-class. 
These pseudo-labels are used for ground-truth in the following analyses we perform.

\begin{figure}[t]
    \centering
    \adjustbox{max width=0.5\linewidth}{
    \includegraphics{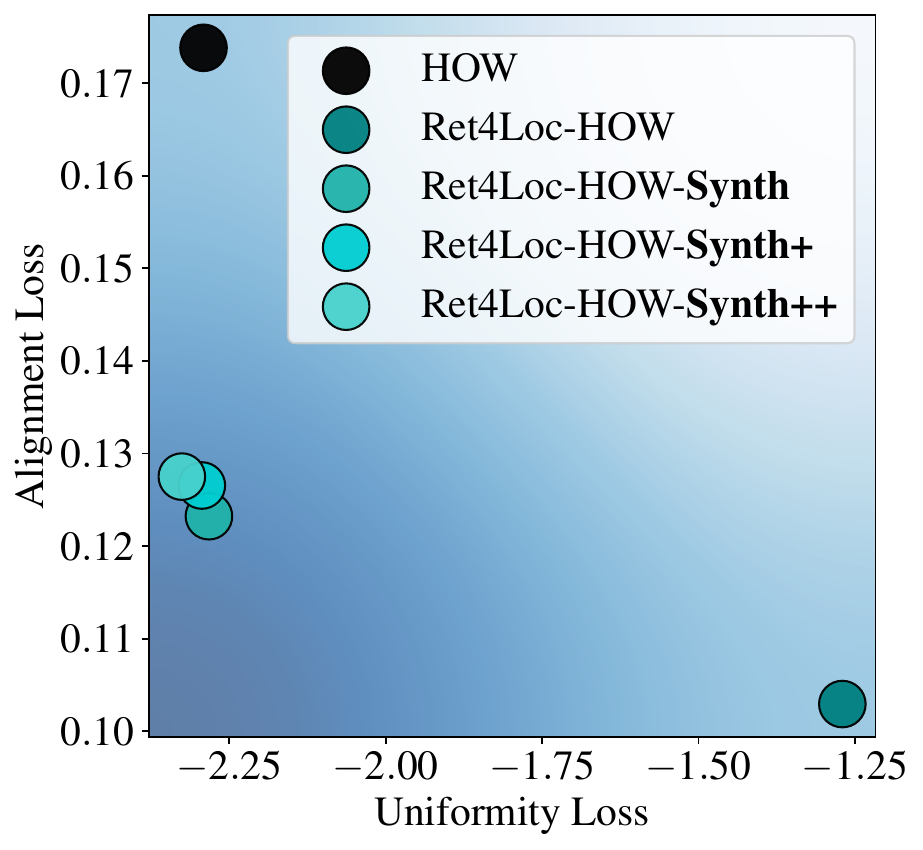}
    }
    \caption{
        \textbf{Alignment \vs uniformity analysis}~\citep{wang2020understanding} using pre-extracted features from the validation set of SfM-120K.
        We sample 1000 real images from the validation set and extract features for each of those images along with their 11 synthetic variants (which we generate as described in~\Cref{sec:synthetic}).
        Each real and its 11 synthetic variants form a ``class'', which we use to measure the alignment loss.
        More concretely, we measure the alignment loss between a real image and each of its 11 synthetic variants.
        The uniformity loss is measured across all features.
        This analysis was originally performed in the context of visual representation learning via self-supervised models, 
        and \cite{wang2020understanding} observed that best performing models tend to position towards the bottom left part of such 2D plots, as indicated by darker blue in the figure.
    }
    \label{fig:alignment_uniformity}
\end{figure}

\myparagraph{Alignment \vs uniformity analysis}
Following~\cite{wang2020understanding},\footnote{\url{https://github.com/SsnL/align_uniform}} we measure the alignment loss between the feature of a real image and each of its 11 synthetic variants, and the uniformity loss across all features.
The results are shown in~\Cref{fig:alignment_uniformity}.
We see that HOW and \retforloc are located on the upper left and lower right corners, minimizing either the uniformity or alignment losses, respectively.
Whereas, our 3 models trained with synthetic data (\Synth, \Synthp and \Synthpp) exhibit a better balance in the trade-off between the two losses, positioning closer to the lower left corner.
We note that this analysis was originally performed in the context of visual representation learning via self-supervised models,
and \cite{wang2020understanding} observed that best performing models tend to position towards the bottom left part of such 2D plots, as indicated by darker blue in~\Cref{fig:alignment_uniformity}.

\myparagraph{Clustering analysis}
We cluster all 12000 features for each model into 1000 clusters via $k$-means.
Then we measure the purity, adjusted random index (ARI) and normalized mutual information (NMI) scores between hard cluster assignments and pseudo-labels.
The results are reported in~\Cref{tab:clustering}.
We observe that our models trained with synthetic data consistently improve the clustering metrics over HOW and \retforloc.

\begin{table}[t]
    \centering
    \begin{tabular}{lcccc}
    \toprule
        Method &  $K$ & training images (total) & training time & max. GPU memory \\ \midrule
       \retforloc  & --  & 91K (real) & $\sim$4 hours &  21.8 GB    \\
       \Synthpp & 3  & 91K (real) $ + $ 1M (synthetic) &  $\sim$8 hours       & 65.4 GB  \\ \bottomrule
    \end{tabular}
    \caption{\textbf{Training time and memory usage for Ret4Loc models.} Note that these timings \textit{do not include the time needed for synthetic data generation}, since we ran generation as a pre-processing step, only once. Timings for image generation are presented in~\Cref{tab:generationtime}.}
    \label{tab:timings}
\end{table}

\begin{table}[t]
    \centering
    \begin{tabular}{lcc}
    \toprule
 Resolution & Steps & Time (sec) \\ \midrule
 512        & 10                  & 0.30                          \\
 512        & 20                  & 0.44                          \\
 512        & 30                  & 0.64                          \\
 768        & 10                  & 0.62                          \\
 \underline{768}        & \underline{20}                 & \underline{0.98}                          \\
 768        & 30                  & 1.39                          \\
 1024       & 10                  & 1.16                          \\
 1024       & 20                  & 1.99                          \\
 1024       & 30                  & 2.78                          \\
 \bottomrule
    \end{tabular}
\caption{\textbf{Generation time for InstructPix2Pix.} Time required to generating a single variant for an image, hence approximately 25 hours for processing the full SfM-120k dataset (91K images) for resolution 768 and 20 steps. Different variants can be processed in parallel.}
    \label{tab:generationtime}
\end{table}

\subsection{Generation and training times}
\label{sec:app_timings}

In this section we separately report and discuss i) the model training time and memory usage, and ii) the synthetic image generation time, as, in practice, we ran synthetic image generation \textit{as an offline process} once and saved all 11 variants for each training image. 

\Cref{tab:timings} reports training time and memory usage for Ret4Loc model obtained using a single A100 GPU. 
Again, training times in this table  do not include time corresponding to the image generation process.

We present synthetic image generation time in~\Cref{tab:generationtime}. The latter is affected by two parameters: the generated image resolution (Resolution) and the number of diffusion steps (Steps). The table shows the generation time per image in seconds (Time), for different resolutions and number of steps. 
We chose to use a resolution of 768 pixels and 20 steps for all results in the paper. For these parameters, the time needed for generating all synthetic images for a single prompt was approximately 25 hours for the complete SfM-120k dataset. Given that the process is independent per prompt, we ran synthetic data extraction in parallel for all prompts and generated images for all variants \textit{in about a day}.  With model training taking approximately 8 hours, we see that data generation time is at least 3 times higher than the actual model training. However, we do not believe this is an issue in practical applications for mainly three reasons:

\begin{enumerate}
    \item As a pre-processing step, data generation needs to run only once. This is unlike model training, a process that one usually needs to run multiple times, \eg to perform hyperparameter validation. 
    \item The overall training time for our model (data generation + model training) is approximately 32 hours. This means that learning state-of-the-art retrieval models for visual localization using our method is manageable in reasonable time, even with modest resources.
    \item We believe that over time the data synthesis overhead will eventually be even less of an issue: generative AI models will inevitably become more efficient and we envision in the near future such generation to happen on-the-fly, during batch construction. 

\end{enumerate}

\section{Qualitative evaluation of synthetic variants}
\label{sec:app_synthetic}

\begin{figure}
    \centering
    \includegraphics[width=\linewidth]{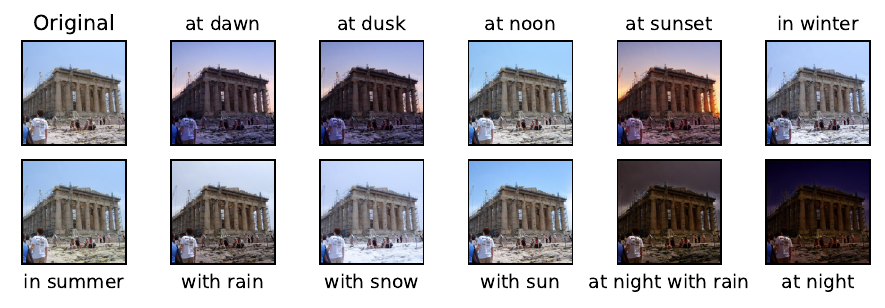} \\
    \includegraphics[width=\linewidth]{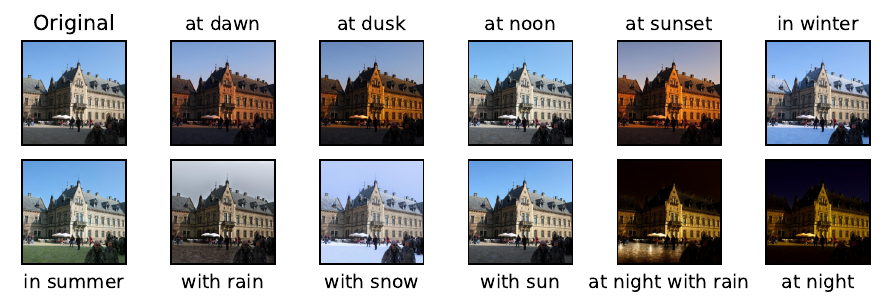} \\
    \includegraphics[width=\linewidth]{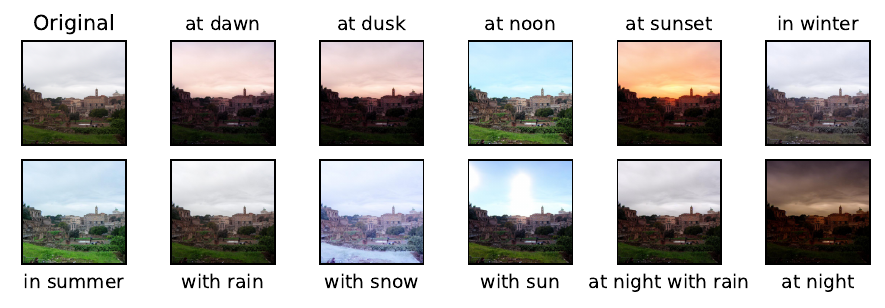} \\
    \caption{\textbf{All synthetic variants} for training images shown in~\Cref{fig:synthetic_data}.}
    \label{fig:all_variants}
\end{figure}

In~\Cref{fig:all_variants}, we show the complete set of 11 variants for the three images depicted in~\Cref{fig:synthetic_data}.

\subsection{Results with ControlNet}

When exploring which generative AI model to use, besides InstructPix2Pix, we also tested ControlNet~\citep{ZhangICCV2AddingConditionalControl2T2IDiffusionModels}. We found the latter to produce images far more stylized and we deemed them unfit for our purpose. 
We compare InstructPix2Pix and ControlNet generations for the images depicted in~\Cref{fig:synthetic_data} of the main paper in~\Cref{fig:ip2p_vs_controlnet_1} and ~\Cref{fig:ip2p_vs_controlnet_2}.

\subsection{Geometric consistency for different steps of the generator}

\begin{figure*}[t!]
    \centering
        \includegraphics[width=.5\linewidth]{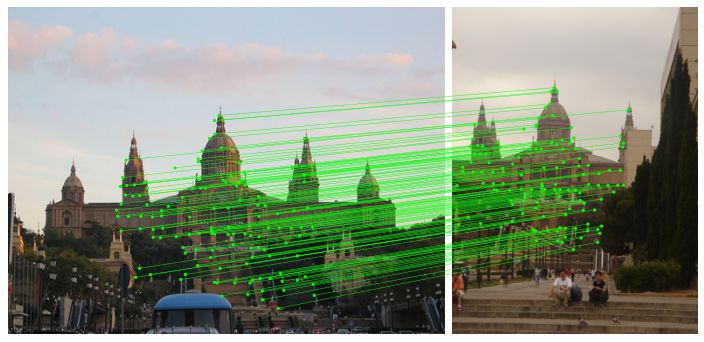}
        \includegraphics[width=.5\linewidth]{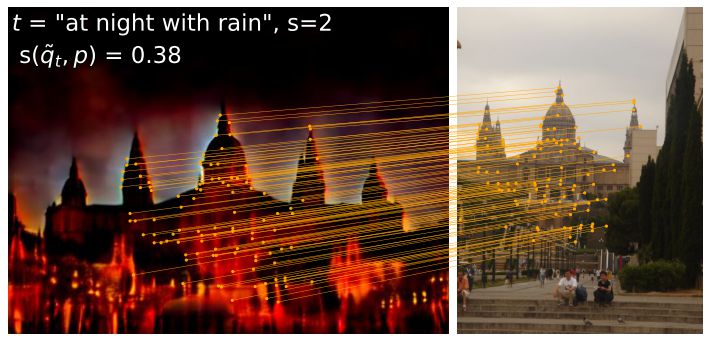}
        \includegraphics[width=.5\linewidth]{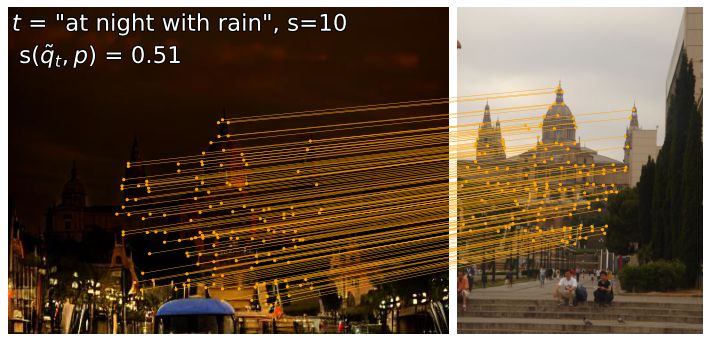}
        \includegraphics[width=.5\linewidth]{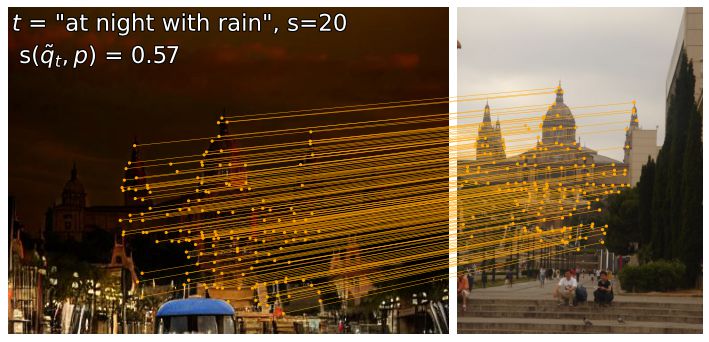}
        \includegraphics[width=.5\linewidth]{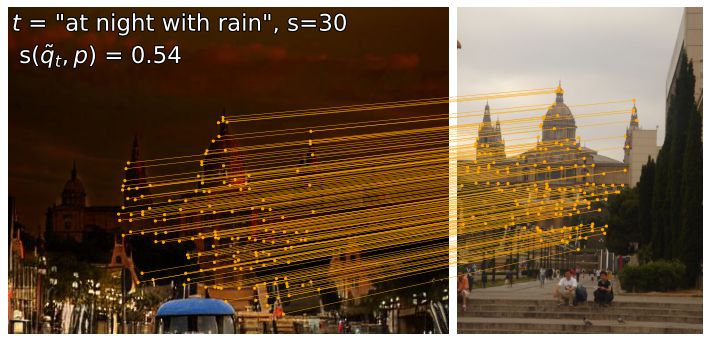}
    \caption{
    \textbf{Geometric matching for image generation after different steps.} We see that the number of matches is generally consistent for steps between 10 and 30. For this visualization we increase the number of features used by LightGlue to 2048.}
    \label{fig:geom_steps}
\end{figure*}

In~\Cref{fig:geom_steps} we show geometric consistency results when varying the number of diffusion steps, \ie using 2, 10, 20 or 30. We see that 10 steps are enough for getting a large number of geometric matches. After preliminary explorations, we set the number of steps for generation to 20 for all results presented in the paper.

\subsection{Synthesis failure cases}

\begin{figure}
    \centering
    \includegraphics[width=\linewidth]{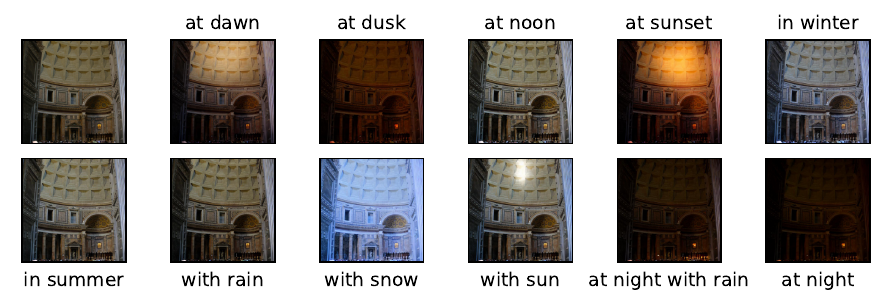} \\
    \includegraphics[width=\linewidth]{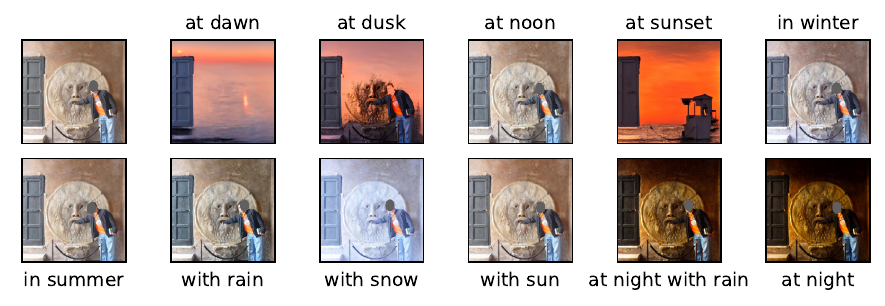} \\
    \includegraphics[width=\linewidth]{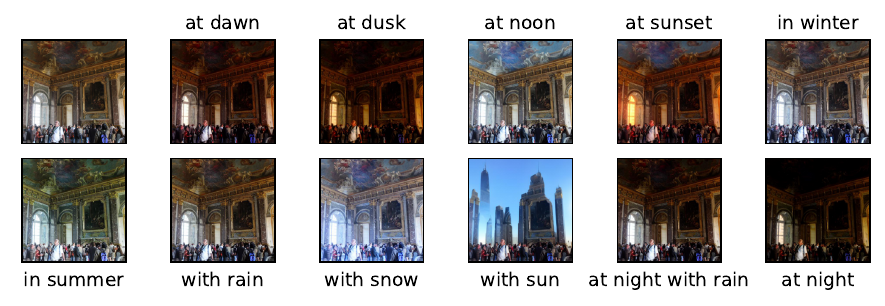} \\
    \caption{\textbf{Peculiar synthesis and failure cases} for training images not suited to the prompts.}
    \label{fig:all_variants_fail}
\end{figure}

In~\Cref{fig:all_variants_fail}, we show what synthetic variants look like for cases where most prompts don't really make sense, \ie for indoor images.
We see that in many cases generations are still plausible and can be interpreted as extreme illumination changes.
We can also see a few failure cases in the middle (``at dawn/dusk/sunset'') and bottom (``with sun'') examples in~\Cref{fig:all_variants_fail}.
As we show in~\Cref{tab:ablations}, keeping such images during training does not really decrease the model's performance significantly.
We believe this to be because, even in such cases, a part of the depicted instance (\eg the door in the middle example or parts of the walls in the bottom one) is still visible even in the most distorted variants.

\subsection{Geometric consistency for indoor and failure cases}

\begin{figure}
    \centering
    \adjustbox{max width=\textwidth}{
    \includegraphics[height=3cm]{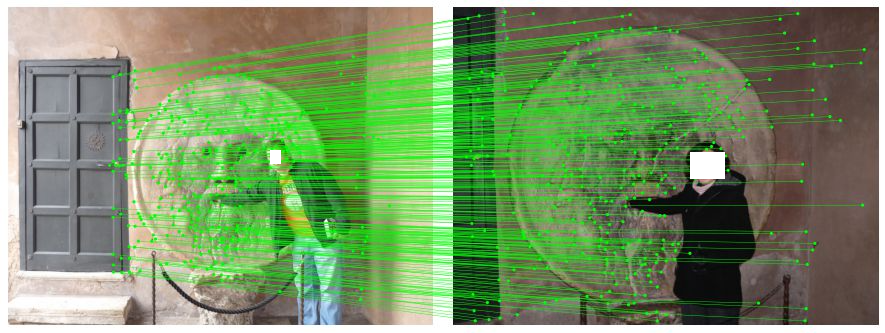} 
    \includegraphics[height=3cm]{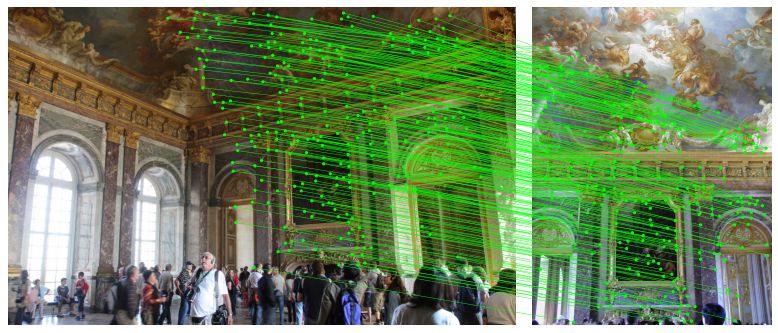} } \\
    \adjustbox{max width=\textwidth}{
    \includegraphics[height=3cm]{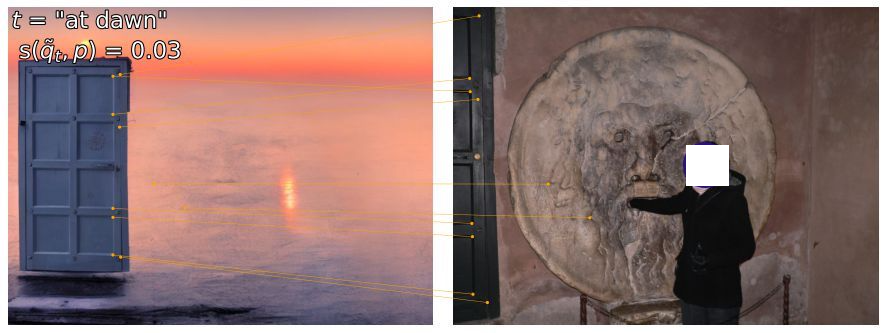} 
    \includegraphics[height=3cm]{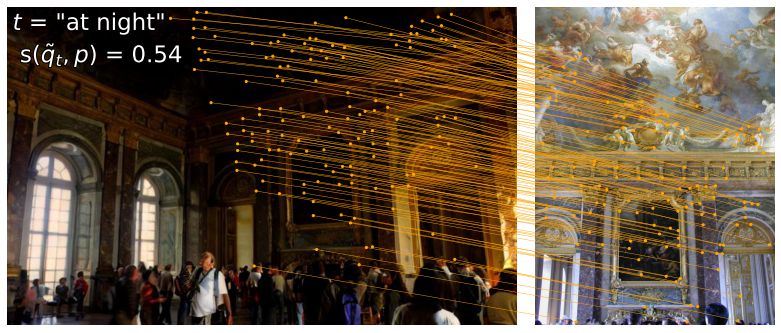} } \\
    \adjustbox{max width=\textwidth}{
    \includegraphics[height=3cm]{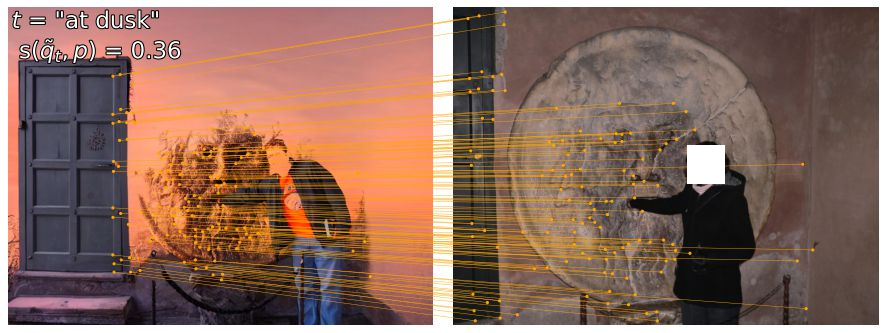} 
    \includegraphics[height=3cm]{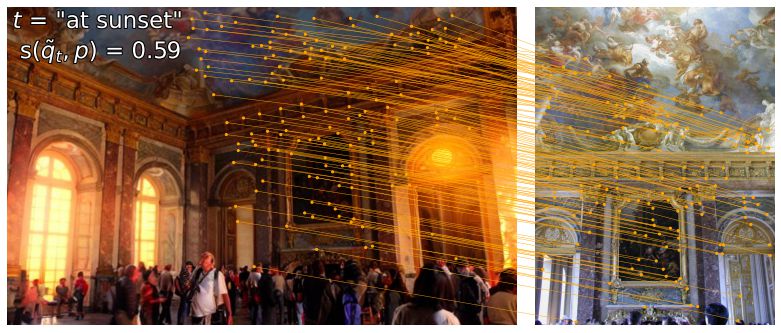} } \\
    \adjustbox{max width=\textwidth}{
    \includegraphics[height=3cm]{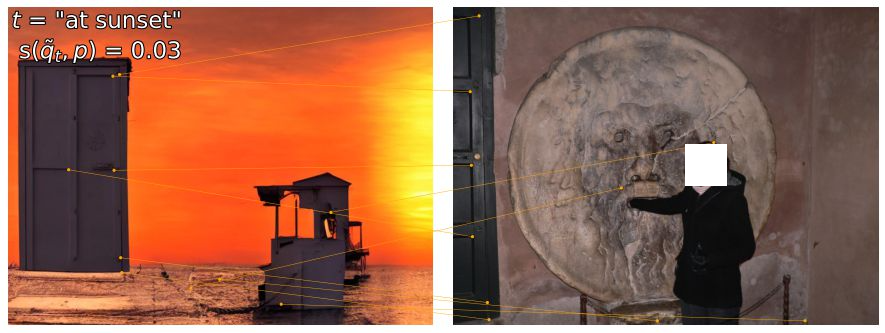} 
    \includegraphics[height=3cm]{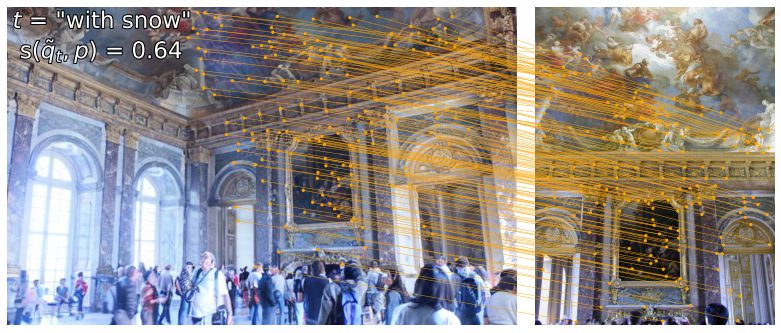} } \\
    
    \caption{\textbf{Geometric verification for indoor image variants}. The corresponding prompt and score $s$ are printed on each pair.    Matches are discovered with the LightGlue~\citep{lindenberger2023lightglue} algorithm.}
    \label{fig:geom_indoor}
\end{figure}

In~\Cref{fig:geom_indoor}, we show some interesting cases of geometric matching for a failure case (left) and for an indoor image.
We see that even some of the extremely distorted images (left side, third row) or images with unrealistic alterations, \ie the indoor image on the right altered with an ``at sunset'' prompt, can still be successfully matched with the LightGlue~\citep{lindenberger2023lightglue} method.

% -------------------------------------------------------------------------
% Extended RW
% -------------------------------------------------------------------------
\section{Extension of the related work}
\label{sec:AppRW}

In this section, we mention additional prior work that we consider related to our submission and that could provide a broader perspective. It complements the discussion of the most related works available in Sec~\ref{sec:related}.

\myparagraph{Retrieval for visual localization and place recognition}
Visual localization consists in estimating the 6-DoF camera pose from a single RGB image within a given area. The reference area can be represented as a 3D point cloud, also referred to as an SfM map~\citep{SeIROS02GlobalLocalizationDistinctiveVisualFeatures,IrscharaCVPR09FromSFMLocationRecognition,LiECCV10LocationRecPriorFeatureMatching,SattlerPAMI17EfficientPrioritizedMatching,SarlinCVPR19FromCoarsetoFineHierarchicalLocalization,TairaPAMI21InLocIndoorVisualLocalization,
Germain3DV19SparseToDenseHypercolumnMatchingVisLoc,HumenbergerX20RobustImageRetrievalBasedVisLocKapture} 
and camera localization relies on using 2D-3D matches between a query image and the 3D representation. 
Alternatively, learning-based localization methods 
are trained to regress 2D-3D matches~\citep{ShottonCVPR13SceneCoordinateRegression,BrachmannICCV19ExpertSampleConsensusReLocalization,CavallariCVPR17OntheFlyCameraRelocalisation,BrachmannPAMI22VisualCameraReLocalizationFromRGBAndRGBDImagesUsingDSAC} 
or to directly predict the camera pose~\citep{KendallICCV15PoseNetCameraRelocalization,SattlerCVPR19UnderstandingLimitationsPoseRegression}.
State-of-the-art visual localization approaches often rely on
image retrieval techniques to either provide an approximate pose estimate that is particularly relevant for place recognition known also as geo-localization \citep{ZamirB16LargeScaleVisualGeoLocalization,KimCVPR17LearnedContextualFeatureReweightingGeoLoc,LowryTROB16VisualPlaceRecognitionSurvey,leyva2023data,ToriiPAMI15VisualPlaceRecognRepetitiveStructures,ToriiPAMI18247PlaceRecognitionViewSynthesis,ArandjelovicCVPR16NetVLADPlaceRecognition}, 
or as an initial step for SfM based methods as it allows  large scale deployment. Indeed, it was shown that a good retrieval model not only reduces  the localization computational cost  by limiting the search to the scene parts potentially visible in a given query image, but can also yield to  improved  localization accuracy~\citep{HumenbergerX20RobustImageRetrievalBasedVisLocKapture,TairaPAMI21InLocIndoorVisualLocalization,SarlinCVPR19FromCoarsetoFineHierarchicalLocalization}.

Typically, visual localization methods that rely on compact image-level descriptors in their first retrieval step use
representations trained for landmark retrieval or place recognition~\citep{ToriiPAMI18247PlaceRecognitionViewSynthesis,ArandjelovicCVPR16NetVLADPlaceRecognition,LiuICCV19StochasticAttractionRepulsionEmbedding,radenovic2018fine,revaud2019learning,KimCVPR17LearnedContextualFeatureReweightingGeoLoc,leyva2023data}. %\gcs{is that clearer?}
In order to further refine the retrieval step used in localization,
several works propose to apply local feature based re-ranking of the top retrieved images \citep{SarlinCVPR19FromCoarsetoFineHierarchicalLocalization,wang2022transvpr,sarlin20superglue,cao2020unifying,hausler2021patch}. Moreover, local features have been employed without aggregation for retrieval for a long time. Among them, the most recent ones rely on Aggregated Selective Match Kernels (ASMK)~\citep{tolias2013aggregate}. This is the case of HOW~\citep{tolias2020learning} and FIRe~\citep{weinzaepfel2022learning}. In a concurrent work, \citet{aiger2023yes} replace the ASMK matching step with Constrained Approximate Nearest Neighbors (CANN) and shows large gains for HOW and FIRe. In our approach, which also relies on HOW, we could
also replace ASMK with CANN to further improve the results, but we consider this to be out of the scope of this paper.

\myparagraph{Building datasets for training models robust to adversarial weather, seasonal and lighting conditions}
The recent progress on computer graphics platforms enables  the generation of photo-realistic virtual worlds with diverse, realistic, and physically plausible conditions including  adversarial weather, seasonal and lighting 
conditions~\citep{RosCVPR16SYNTHIADataset,RichterECCV16PlayingForData,GaidonCVPR16VirtualWorldsAsProxyTracking,CabonX20VirtualKITTI2}. However there are two issues with such platforms. First, the generation of such datasets still requires a large amount of expert knowledge and important manual effort. Second, while they might help reducing the domain gap between these conditions, they might introduce an extra domain shift, known as the sim-to-real gap. 
%\ie synthetic to real. 
Therefore, in parallel, researchers also collected real images 
sometimes recorded from the same streets at different times and under different conditions~\citep{maddern20171,ToftPAMI22LongTermVisualLocalizationRevisited,SakaridisICCV19GuidedCurriculumModelAdaptationSemSegm,SakaridisIJCV17SemanticFoggySceneUnderstandingWithSyntheticData}, to be used either to properly assess the impact of adverse weather/daylight on the models or to train models robust to such conditions. A third option to generate such datasets
is to add weather-related artifacts on top of  existing images~\citep{RenCVPR17VideoDesnowingAndDerainingBasedOnMatrixDecomposition,QianCVPR18AttentiveGAN4RaindropRemoval,YouPAMI16AdherentRaindropModelingDetectionAndRemovalInVideo,CaiTIP17DehazeNetAnEnd2EndSystem4SingleHazeRemovaly,LiICCV17AODNetAllInOneDehazingNetwork,LiCVPR18SingleImageDehazingViaConditionalGenerativeAdversarialNetwork,RenCVPR18GatedFusionNetwork4SingleImageDehazing,ZhangCVPR18DenselyConnectedPyramidDehazingNetwork,JenicekICCV19NoFearOfTheDark,WuCVPR21DANNetOneStageDANetworkUnsupervisedNighttime}.

With the recent success of image generative models~\citep{rombach2022high,ZhangICCV2AddingConditionalControl2T2IDiffusionModels,QinX23UniControlAUnifiedDiffusionModel4ControllableVisualGeneratioInTheWild,brooks2022instructpix2pix,NicholICML22GLIDETowardsPhotorealisticImageGenerationEditingDiffusionModels}, generating photo-realistic images became much simpler.
User without any expertise can easily augment existing datasets or even generate completely synthetic image sets~\citep{dunlap2023alia,TrabuccoX23EffectiveDataAugmentationWithDiffusionModels,he2023synthetic,azizi2023synthetic,sariyildiz2023fake}. However, our augmented SfM-120K dataset~\citep{radenovic2018fine}, 
is the first one that focuses on adversarial weather, seasonal and lighting conditions with the aim of improving long-term visual localization.

\myparagraph{Leveraging semantics to learn features robust to weather and seasonal conditions} To address robustness to weather or seasonal variations, one could alternatively leverage semantics as an auxiliary source of information and learn robust representations that encapsulate some high-level semantic information, so they are by design more robust to appearance changes~\citep{Kobyshev3DV14MatchingFeaturesCorrectlyThroughSemanticUnderstanding,ToftICCVWS17LongTerm3DLocalizationPoseFromSemanticLabellings,SchonbergerCVPR18SemanticVisualLocalization,ToftECCV18SemanticMatchConsistencyLTVL,GargRSSC18LoSTAppearanceInvariantPlaceRecognVisualSemantics, YuIROS18VLASEVehicleLocalizationAggregatingSemanticEdges,ShiICIP19VisLocSparseSemantic3DMap,BenbihiICRA20ImageBasedPlaceRecBucolicEnv,HuTIP21DASGILDA4SemanticAndGeometricAwareImageBasedLocalization,PaolicelliX22LearningSemantics4VisualPlaceRecognitionThroughMultiScaleAttention,XueCVPR23SFD2SemanticGuidedFeatureDetectionDescription}.  
All these works  nevertheless require to train an additional 
full network, or at least an additional segmentation head, to be able predict the semantic information. On top of the extra training cost, these additional components also need to be trained in a way that is robust to domain shifts and dataset variations. This is an non trivial goal which constitutes an full research field in itself~\citep{CsurkaFTCGV22SemanticImageSegmentationTwoDecadesOfResearch}. 

\myparagraph{Visual style transfer}
Visual style transfer allows to combine the content of one image with the style of another one to create a new image. Such techniques have been used to mitigate the domain shift that can exist between training and test images, in particular could also handle visual appearance variations, due to seasonal, weather, or daylight changes.
By exploiting the progress in image-to-image translation and style transfer~\citep{GatysNIPS15TextureSynthesisCNN,HuangICCV17ArbitraryStyleTransfer,LiECCV18ClosedFormImageStylization}, several methods have used them either as a pre-processing step or to align images~\citep{CsurkaTASKCV17DiscrepancyBasedNetworksUDA,ThomasACCV19ArtisticObjectRecognitionUnsupervisedStyleAdaptation}, to learn to synthesize target-like images~\citep{LiuNIPS17UnsupervisedI2ITranslationNetworks,RomeraIVS19BridgingDayNightDomainGapsSIS,XuICCV21CDAdaACurriculumDA4NighttimeSiS}, 
independently  or integrated within domain adaptation techniques~\citep{HoffmanICML18CyCADACycleConsistentAdversarialDA,WuECCV18DCANDualChannelWiseAlignmentNetworksUDA,LiCVPR19BidirectionalLearningDASemSegm,ChoiICCV19SelfEnsemblingGANBasedDataAugmentationDASemSegm,ToldoIVC20UDAMobileSemSegmCycleConsistencyFeatureAlignment,ChengICCV21DualPathLearning4DASS,WangAAAI21ConsistencyRegularizationDASIS,Yang21BiMixBidirectionalMixing4DANighttimeSiS}.
In general, these models learn to cope with new appearances/styles that are learned from a set of images representing the targeted style or appearance. Furthermore, while in theory applicable to our problem, few of them addressed variations due to seasonal, weather, or daylight changes.
In contrast, our model changes the appearance of the training images
where the new appearance, in particular adversarial weather and daylight variation is simply described with natural language.

\begin{figure*}[t!]
    \centering
    \begin{subfigure}[t]{0.5\textwidth}
        \centering
        \adjustbox{max width=0.95\textwidth}{
        \includegraphics[height=30cm]{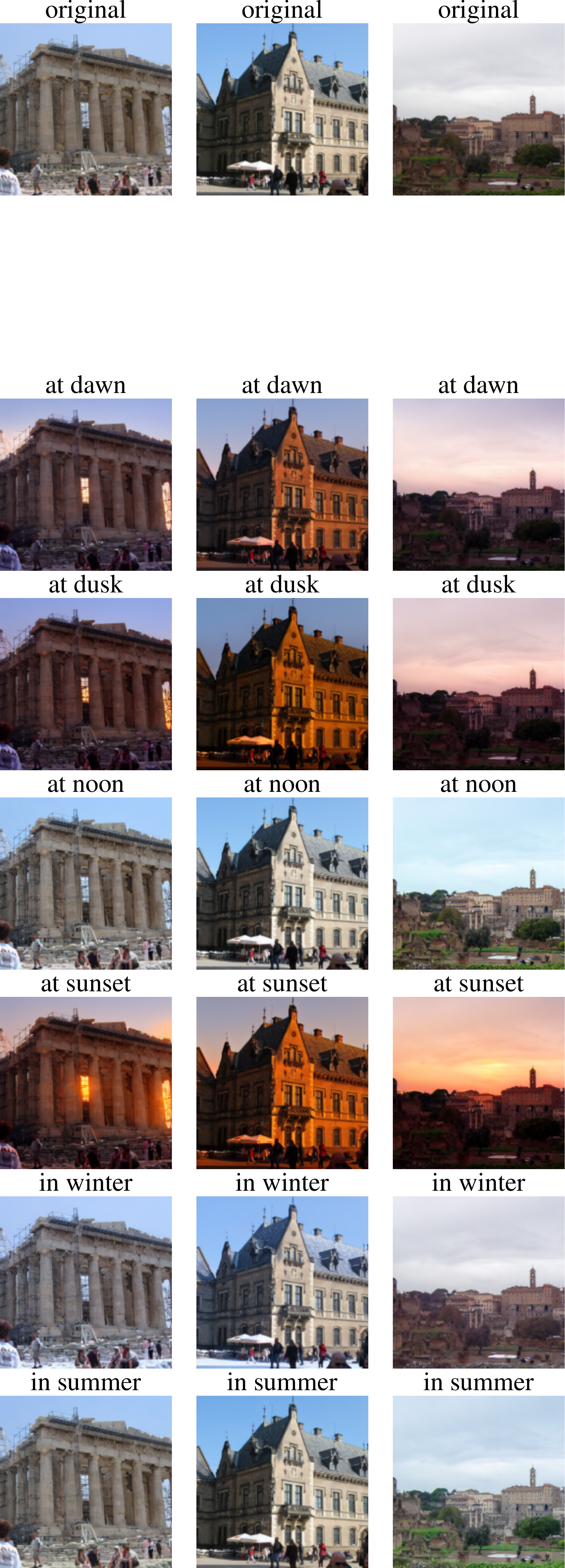}
        }
        \caption{InstructPix2Pix}
    \end{subfigure}%
    ~ 
    \begin{subfigure}[t]{0.5\textwidth}
        \centering
        \adjustbox{max width=0.95\textwidth}{
        \includegraphics[height=30cm]{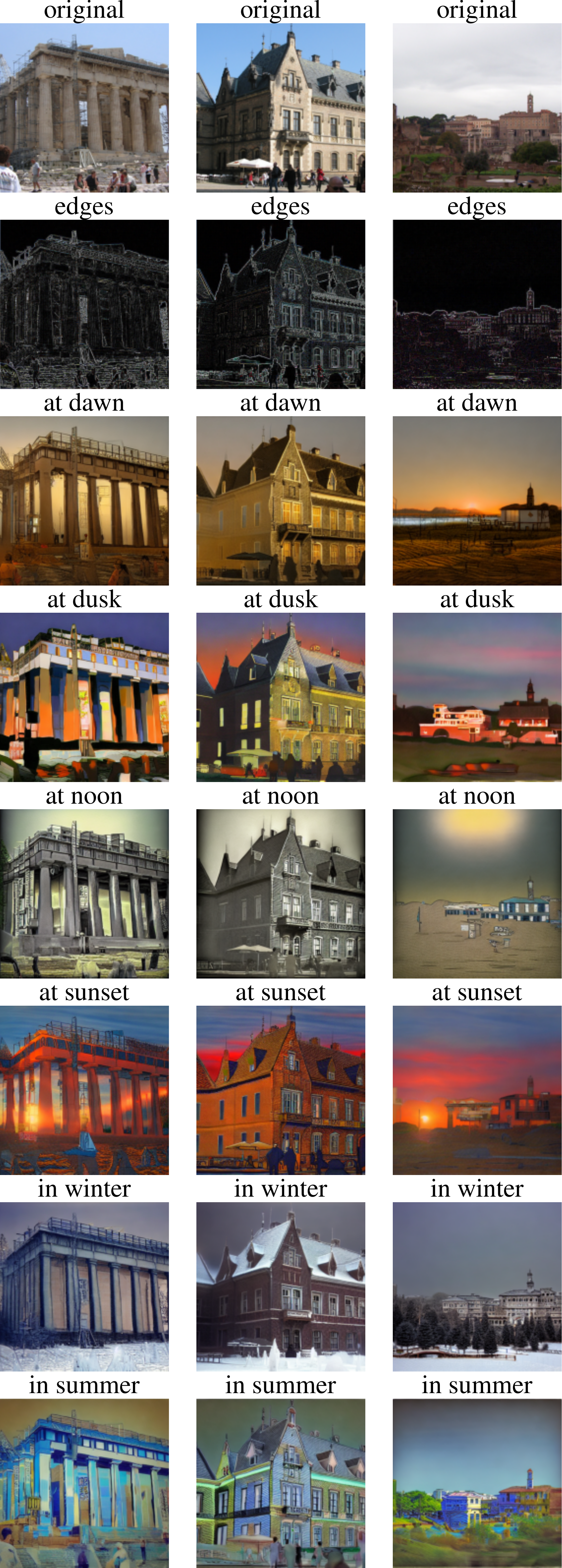}
        }
        \caption{ControlNet}
    \end{subfigure}
    \caption{
        \textbf{Synthetic images from InstructPix2Pix~\citep{brooks2022instructpix2pix} (left) and ControlNet~\citep{ZhangICCV2AddingConditionalControl2T2IDiffusionModels} (right)}. For a given image (``original'') and the 11 prompts we consider in this work,
            for ControlNet, we generate images using edge maps obtained from the original images.
            See~\cref{fig:ip2p_vs_controlnet_2} for the rest of the prompts.
    }
    \label{fig:ip2p_vs_controlnet_1}
\end{figure*}

\begin{figure*}[t!]
    \centering
    \begin{subfigure}[t]{0.5\textwidth}
        \centering
        \adjustbox{max width=0.95\textwidth}{
        \includegraphics[height=30cm]{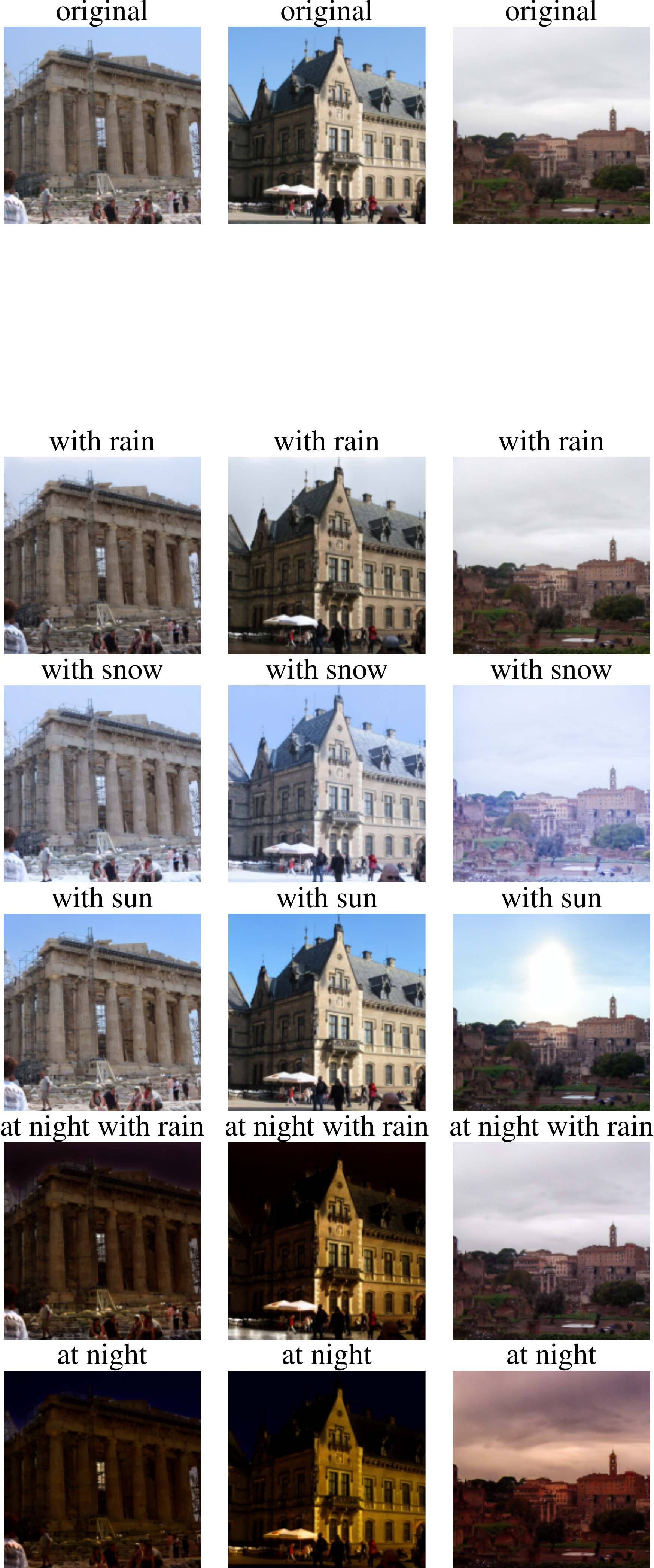}
        }
        \caption{InstructPix2Pix}
    \end{subfigure}%
    ~ 
    \begin{subfigure}[t]{0.5\textwidth}
        \centering
        \adjustbox{max width=0.95\textwidth}{
        \includegraphics[height=30cm]{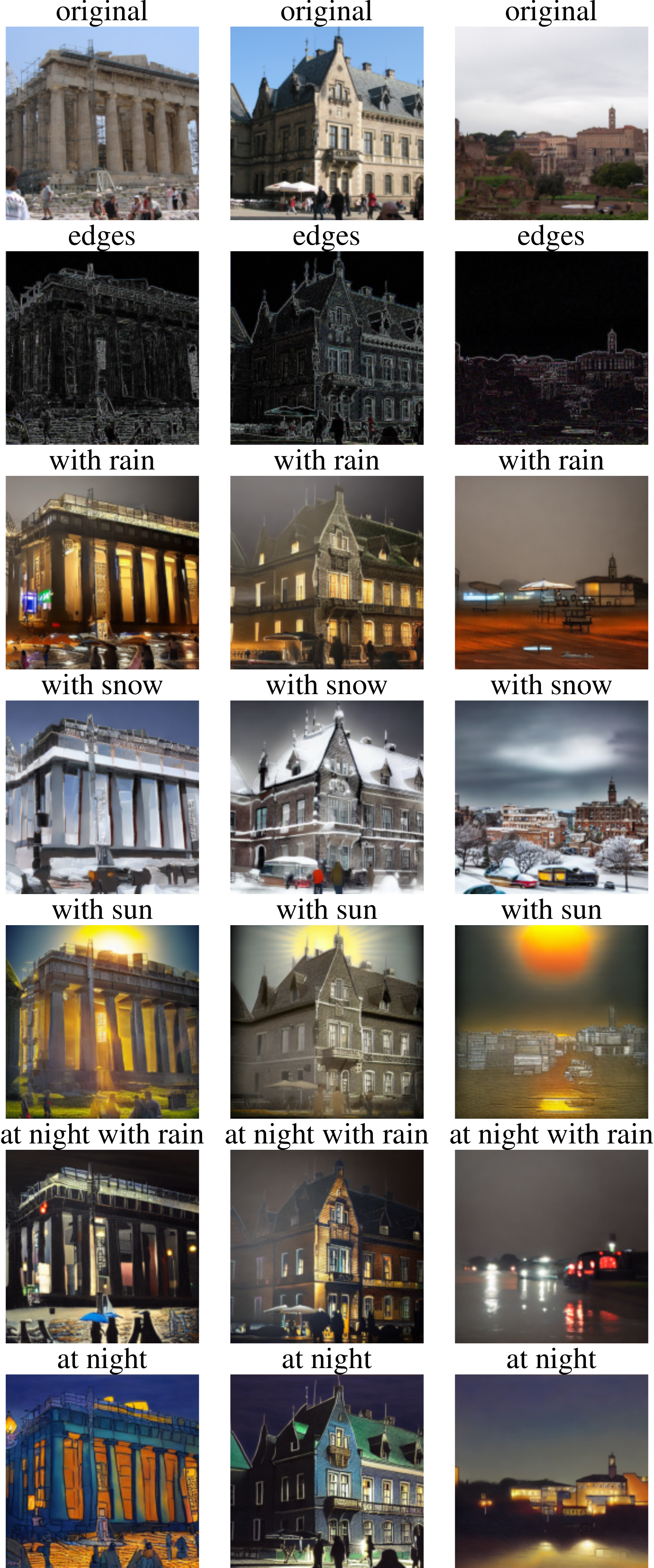}
        }
        \caption{ControlNet}
    \end{subfigure}
    \caption{Continuation of~\cref{fig:ip2p_vs_controlnet_1}.}
    \label{fig:ip2p_vs_controlnet_2}
\end{figure*}

\end{document}